\documentclass[final,5p,times,twocolumn]{elsarticle}

\usepackage{amssymb}
\usepackage{lipsum}

\usepackage{amssymb}
\usepackage{graphicx}
\usepackage{subfig}
\usepackage{xcolor}
\usepackage{amsmath} 
\usepackage{multirow}
\usepackage{multirow}
\usepackage{hyperref}
\usepackage[most]{tcolorbox}
\usepackage{adjustbox}
\usepackage{amsfonts}
\usepackage{wrapfig}
\usepackage{subcaption}
\usepackage{amsmath,amsfonts}
\usepackage{algorithmic}
\usepackage{algorithm}
\usepackage{array}
\usepackage{comment}
\usepackage{xr}
\usepackage{xr-hyper}
\usepackage{hyperref, cleveref}
\usepackage{enumitem}
\usepackage{amsmath}
\journal{Biomedical Signal Processing and Control}

\begin{document}

\begin{frontmatter}



\title{CEM-TUDASR: Computationally Efficient Multi-Modality Transformer based Unsupervised Domain Adaptive Super-Resolution Approach}


\author[first]{Anjali~Sarvaiya}
\affiliation[first]{organization={Sardar Vallabhbhai National Institute of Technology (SVNIT), Surat, India.}}

\author[first]{
        Jay~Kadel
        }
        
\author[first]{
        Kishor~Upla
        }%% Author name

\author[second]{Kiran~Raja}
\affiliation[second]{organization={Norwegian University of Science and Technology (NTNU), Norway.}\\ }
% \email{\{anjali.sarvaiya.as, jay.k.kadel, kishorupla\}@gmail.com}\\ \email{\{kiran.raja\}@ntnu.no}\\}
% \affiliation{organization={},%Department and Organization

%% Abstract
\begin{abstract}
Wireless Capsule Endoscopy (WCE) has emerged as a non-invasive and patient-friendly imaging modality for comprehensive visualization of the Gastrointestinal (GI) tract. However, due to the miniaturized capsule design, limited onboard optics, sensor constraints, and wireless transmission bandwidth restrictions, WCE images are inherently acquired at low spatial resolution, often accompanied by noise, motion blur, and illumination degradation. These limitations significantly reduce the visibility of fine anatomical structures such as mucosal textures, vascular patterns, and lesion boundaries, thereby affecting diagnostic reliability. To address this challenge, this paper proposes a computationally efficient unsupervised Transformer-based super-resolution framework, termed as \emph{CEM-TUDASR}, for enhancing WCE images without relying on paired Low-Resolution (LR) and High-Resolution (HR) training data. The proposed framework integrates a domain-adaptive degradation modeling network that learns to synthesize realistic WCE-like LR images from HR conventional endoscopy images, enabling effective unpaired training and reducing the domain discrepancy between conventional and capsule endoscopy data. Furthermore, a Transformer-based SR generator incorporating Deep Attention Blocks (DABs) and a Fusion Attention Block (FAB) is introduced to jointly capture long-range contextual dependencies and fine-grained local structural details. This architecture facilitates improved reconstruction of diagnostically relevant regions while preserving structural consistency and perceptual fidelity. The proposed model is trained on a newly curated WCE dataset derived from the Kvasir Capsule dataset and extensively evaluated on external benchmark datasets, including KID and GIANA, to validate its effectiveness and generalization capability. Quantitative evaluation using no-reference image quality assessment metrics, including BRISQUE, PIQE, NIQE, and the domain-specific EndoQM, demonstrates that the proposed method consistently outperforms existing unsupervised SR approaches in terms of perceptual quality, structural preservation, and domain-specific visual fidelity. Qualitative analysis further confirms superior restoration of subtle mucosal textures, vascular structures, and clinically significant anatomical details essential for accurate interpretation. In addition, cross-domain evaluation on retinal images demonstrates the robustness and adaptability of the proposed framework across diverse medical imaging modalities. Despite achieving high-quality reconstruction performance, the proposed architecture maintains computational efficiency with only 2.67 million parameters and 169.94 GFLOPs, making it highly suitable for deployment in real-time and resource-constrained clinical environments, including portable and embedded endoscopic systems. GitHub Link: https://github.com/Jay042003/CEM-TUDASR.git
\end{abstract}


% \begin{highlights}
% \end{highlights}

%% Keywords

\end{frontmatter}

\section{Introduction}
\label{sec:introduction}
In the medical field, optical colonoscopy is a gold standard technology for the early detection, diagnosis, and treatment of significant disorders related to both, upper and lower Gastrointestinal (GI) tract of human body. It is useful for Colorectal Cancer (CRC), Crohn’s disease, ulcerative colitis, hemorrhoids, and inflammation. While this has acknowledged significant clinical efficacy in reducing CRC incidence, it remains invasive procedure with suboptimal sensitivity in detecting adenomatous polyps. Thus, despite of approximately $19$ million colonoscopies being performed annually in the United States, routine screenings exhibit a polyp miss rate ranging from $6\%$ to $28\%$, contributing to CRC's status as the second most prevalent cancer and leading cause of cancer-related mortality \cite{risk}. Additionally, the small intestine presents a significant challenge for examination using conventional endoscopes due to its anatomical complexity, and traditional procedures which often induce substantial patient discomfort and pain. To avoid these limitations, Wireless Capsule Endoscopy (WCE) is an effective alternative technology introduced recently which enables comprehensive visualization of the entire GI tract and facilitating the detection of adenomatous polyps and non-polypoid lesions that may contribute to CRC progression \cite{2002}. Unlike conventional endoscopes, WCE device is a capsule-shaped imaging system that is swallowed by the patient and, captures real-time video of the GI tract without requiring sedation or extensive bowel preparation. Thus, it is a non-invasive, patient-friendly and painless technology which provides easy diagnosis of irregularities associated with the entire GI tract including small intestine \cite{WCE2000}. However, this forthcoming technology is constrained by hardware limitations, resulting to sense Low-Resolution (LR) images with significant noise and degradation, making the identification of colorectal cancer polyps challenging for endoscopists \cite{Swainiv48}. Hence, it limits the detection of likelihood of missed diagnoses which also reduces diagnostic accuracy. 

To address above issue, Super-Resolution (SR) techniques are often incorporated as an effective solution to reconstruct High-Resolution (HR) images from their LR counterparts without additional imaging time \cite{Application}. By enhancing image fidelity and structural details, SR techniques provide higher-quality visualizations that improve both manual assessment by endoscopists and the performance of automated polyp detection algorithms, ultimately enhancing diagnostic require in clinical settings. Additionally, to analyze the domain discrepancy between natural images and WCE images, we computed the Gradient Magnitude Distribution (GMD) for both data in Fig. 1 in \emph{Supplementary material} which shows the necessity for specialized SR models that incorporate domain-specific characteristics and robust adaptation strategies to bridge the gap between natural and clinical imaging domains. %(i.e., WCE data).

Previously, numerous SR models based on deep learning have been actively explored due to the availability of computing resources and datasets, from early methods using Convolutional Neural Networks (CNNs) (i.e., SRCNN \cite{SRCNN, VDSRN, srdensenet}) to the promising SR methods using transformers (i.e., TTSR \cite{TTSR}) for natural images. Due to the exceptional performance of deep learning based SR methods for RGB data, they are also experimented to the traditional endoscopic images \cite{endodeep1, endodeep2, EndoL2h, dcan}. However, improving the reconstruction accuracy to the WCE data is still relatively underexplored mainly due to reasons listed in \emph{Supplementary material}. Moreover, these SR approaches for natural and medical images usually employ supervised training, where models learn to upscale LR images by leveraging paired LR and HR images. The availability of HR endoscopy images is limited, and acquiring well-aligned LR-HR image pairs in the medical domain is particularly challenging \cite{MDASR}. In some cases, synthetic generation of these LR-HR pairs is considered; however, this approach is only feasible when the image acquisition process is meticulously controlled and well-defined. Typically, only an approximate understanding of the acquisition process is available, and LR images are often simulated using known downsampling techniques, such as bicubic downsampling. Consequently, supervised SR models are trained on these synthetic LR-HR pairs, where the LR images do not correspond to true camera-captured observations. As a result, such models tend to learn the degradation patterns of simulated LR samples, limiting their ability to generalize to real-world, clinically obtained LR images \cite{EndoL2h,RCAN,VDSRN}. The challenge is further exacerbated in medical imaging, where generating true LR-HR pairs by manipulating camera settings is impractical and often infeasible.

%Previously, numerous works on SR have been addressed for traditional RGB images. These techniques can be classified broadly into three categories: (i) interpolation-based methods (i.e., nearest neighbor interpolation \cite{NN}, bilinear interpolation \cite{bilinear} and bicubic interpolation \cite{bicubic}), (ii) reconstruction-based methods (i.e., frequency domain methods \cite{frequency} and spatial methods \cite{spatial}), and (iii) learning-based methods (i.e., sparse representation \cite{sparse}, dictionary learning methods \cite{dictionaries} and neural network methods \cite{NN1,brief}. Additionally, the SR models based on deep learning have been actively explored over the recent years due to the availability of computing resources and datasets, from early methods using Convolutional Neural Networks (CNNs) (i.e., SRCNN \cite{SRCNN, VDSRN, srdensenet}) to the promising SR methods using transformers (i.e., TTSR \cite{TTSR}) for natural images. Due to the exceptional performance of deep learning based SR methods for RGB data, they are also experimented to the traditional endoscopic images \cite{endodeep1, endodeep2, EndoL2h, dcan}. However, improving the reconstruction accuracy to the WCE data is still relatively underexplored mainly due to above above-listed reasons.

To circumvent the above-listed limitations, we propose \emph{CEM-TUDASR}, a deep learning SR architecture that is trained in an unsupervised manner, eliminating the need for precisely aligned LR-HR WCE image pairs. The unsupervised training approach allows the model to learn directly from the available data without relying on synthetic pairings, thus improving its generalization performance when applied to real-world WCE images. Here, we introduce a Generative Adversarial Network (GAN)-based transformer architecture that leverages adversarial learning to reconstruct visually realistic SR images with rich high-frequency details. It is important to note that since the vast majority of CNNs for medical imaging makes use of rather small convolution kernels (i.e., $3 \times 3$ or $5 \times 5$), such local convolution operations results in the CNNs being biased towards local spatial structures, which makes them less effective at modeling the long-range dependencies required to better comprehend the contextual information presented in the image \cite{9607461}. 

Transformers, as an alternative network architecture to CNNs, has recently demonstrated superior performance in many computer vision tasks \cite{trans1,trans2}. The core element of a transformer is the self-attention mechanism, which is not subject to the same limitations as convolution operations, making it better at capturing explicit long-range dependencies. Additionally, Transformers have other appealing features, such as they scale up more easily and also more robust to corruption. Further, their weak inductive bias enables them to achieve better performance than CNNs with the aid of large-scale model sizes and datasets. The existing Transformer-based models have shown encouraging results in several medical imaging applications, prompting a surge of interest in further developing such models \cite{transformer}. Thus, the proposed design (i.e., \emph{CEM-TUDASR}) addresses the inherent limitations of conventional architectures in recovering fine anatomical structures, enabling the generation of perceptually consistent and diagnostically meaningful outputs. Importantly, the entire framework is trained in an unpaired setting, eliminating the need for aligned LR-HR image pairs. This enables the model to jointly learn domain-aware degradation and SR reconstruction processes in a fully unsupervised manner, enhancing its adaptability and clinical relevance across real-world WCE scenarios. Thus, it improves the practical applicability of SR in real-world clinical settings where paired data are unavailable.

From a broader clinical and societal perspective, improved WCE image quality can significantly support gastroenterologists in the early detection and diagnosis of gastrointestinal abnormalities such as bleeding  \cite{bleeddetection}, ulcers \cite{ulcerdetection}, inflammation, and colorectal lesions. Better visualization of subtle pathological features can reduce the risk of missed diagnoses, improve diagnostic confidence, and enhance patient outcomes. In addition, the computational efficiency of the proposed framework facilitates its deployment in portable and resource-constrained healthcare environments, contributing to more accessible and reliable diagnostic support systems for large-scale clinical use. The overall motivation, aim, novelty, and objectives are highlighted in the following form.\\
\begin{tcolorbox}[
colback=gray!8,
colframe=black,
boxrule=0.5pt,
arc=2mm,
left=2mm,
right=2mm,
top=1mm,
bottom=1mm,
title={CEM-TUDASR, Aim, Novelty, and Objectives},
fonttitle=\bfseries
]
\textbf{Motivation:} The existing supervised image super-resolution methods for natural images exhibit poor performance on WCE data due to significant domain discrepancies, low-contrast imaging conditions, and the absence of realistic paired LR-HR training data. Additionally, they fail to preserve diagnostically important mucosal textures and fine anatomical structures. This motivates us to develop an unsupervised transformer-based SR framework for WCE medical data that provides perceptually consistent output.  
\vspace{1mm}\\
\textbf{Aim:} The primary aim of this work is to develop a computationally efficient unsupervised transformer-based super-resolution framework with upscale factor of $\times4$ for low-resolution WCE images with preservation of clinically relevant structural and perceptual information.
\vspace{1mm}\\
\textbf{Novelty:} The proposed unsupervised \emph{CEM-TUDASR} framework integrates degradation-aware unsupervised domain adaptation with transformer-based contextual feature representation, Deep Attention Blocks (DABs), Efficient Attention (EA), Efficient Spatial Attention (ESA), and Fusion Attention Blocks (FABs) for robust and perceptually consistent WCE image reconstruction.
\vspace{1mm}\\
\textbf{Objectives:}
\begin{itemize}
    \item To develop an unsupervised transformer-based super-resolution framework for upscale factor of $\times4$ for WCE image enhancement without requiring paired LR-HR training data.
    \item To preserve fine mucosal textures, vascular patterns, and structural consistency during super-resolution reconstruction.
    \item To enhance cross-dataset and cross-domain generalization capability on KID \cite{KID}, GIANA \cite{GIANA}, and retinal \cite{retinal} datasets.
    \item To achieve computationally efficient SR reconstruction suitable for practical clinical deployment.
\end{itemize}
\end{tcolorbox}
% The illustration of the proposed method: \emph{CED-TUDASR} is depicted in Fig.~\ref{fig:kvasir4} on a LR WCE image from the Kvasir dataset, alongside a comparison with other unsupervised SR methods. 
Therefore, the key contributions of this study are as follows: \vspace{-0.2cm}
\begin{itemize}
    \item \textbf{Extension of TUDASR \cite{tudasr} conference work:} The proposed model (i.e., \emph{CEM-TUDASR}) is an improved version of our previously published TUDASR framework \cite{tudasr}. Here, we introduce a new Deep Attention Block (DAB) designed to effectively extract and enhance rich contextual and structural features from HR conventional endoscopy images. Additionally, we incorporate a Fusion Attention Block (FAB) that adaptively integrates the bicubically upsampled LR input with the high-frequency features from the SR generator, enabling the model to focus on diagnostically salient regions and preserve fine anatomical details. 
    \item \textbf{GAN-based SR architecture:} Unlike TUDASR \cite{tudasr}, which utilizes a deterministic encoder-decoder-based SR architecture, the \emph{CEM-TUDASR} adopts a GAN-based formulation. Thus, the adversarial learning setup encourages the generator to produce perceptually realistic outputs with sharper textures and also improves structural fidelity, addressing the limitations of conventional encoder-decoder frameworks in modeling high-frequency information.
    % \item Retention of degradation-aware domain adaptation: The framework preserves the degradation module from TUDASR, allowing unpaired HR conventional endoscopy images to be degraded in a manner consistent with LR WCE images, thus maintaining domain alignment.
    \item \textbf{Unsupervised training:} Notably, to overcome the limitations associated to supervised learning which often struggle to generalize effectively to real-world WCE data, the proposed model (i.e., \emph{CEM-TUDASR}) adopts unsupervised training plugged in with transformer-based GAN architecture for performing SR of WCE images for a factor of $\times4$. Such training effectively addresses the challenge posed by the lack of paired low- and high-resolution clinical datasets.
    % \item Gradient Magnitude Distribution (GMD) analysis: We conduct a detailed GMD analysis that quantifies the domain shift between natural images and WCE images, highlighting the necessity for specialized SR models tailored to the unique characteristics of WCE data.
    \item \textbf{Experimental Validation:} The proposed model i.e., \emph{CEM-TUDASR} is trained using a newly curated version of the Kvasir dataset specifically tailored for the SR task of WCE data. To ensure a fair comparison, all baseline state-of-the-art SR methods were also re-trained on the same dataset. Further, to evaluate the generalization capability of the proposed approach, extensive testing is also conducted on two external datasets—KID \cite{KID} and GIANA \cite{GIANA}—which are not included in the training process. The experimental results across these datasets demonstrate that the proposed method consistently outperforms to the existing SR techniques, both in perceptual quality and quantitative performance. This superiority is reflected in several widely used no-reference image quality metrics, including BRISQUE, NIQE, PIQE, and the domain-specific EndoQM, confirming the robustness of the proposed model on real LR clinical samples.
    \item \textbf{Cross-domain adaptability:} To evaluate the cross-domain adaptability of the proposed model (i.e., \emph{CEM-TUDASR}), we tested it on retinal images \cite{retinal}, which differ notably from endoscopic data in structure, color, and texture. Without being trained on retinal images, the model effectively reconstructs fine details such as vascular bifurcations and optic disc boundaries. It achieves competitive performance on no-reference metrics (i.e., BRISQUE, NIQE, PIQE), demonstrating its ability to generalize across unseen medical imaging modalities and highlighting its potential for broader 2D medical image enhancement tasks without requiring paired training data.
    \item \textbf{Statistical Validation:} Additionally, we conduct a statistical analysis of the quantitative results using a reliability assessment based on the Analysis of Variance (ANOVA) test, which provides evidence of the proposed model’s performance superiority over the other competing methods in a statistical sense. Furthermore, a comprehensive ablation study is also presented to evaluate the contribution and stability of various network components and hyperparameters, highlighting their individual impact on the overall performance of the model.
    \item \textbf{Computational Efficiency}: In particular, \emph{CEM-TUDASR} achieves an effective trade-off between reconstruction performance and computational complexity, producing high-quality SR outputs while maintaining a significantly reduced number of parameters and lower FLOPs compared to other existing unsupervised SR models. Thus, the proposed computationally efficient design facilitates practical applicability in real-time and resource-limited clinical environments, including portable or embedded WCE systems.

\end{itemize}

% The rest of the paper is organized in following manner. In the next section, the extensive literature survey for different SR methods specifically for medical images is discussed. Section~\ref{sec:proposed} provides a complete details of the proposed architecture, encompassing a comprehensive explanation of the incorporated loss functions. The exhaustive experimental validation of the proposed method with comparison to other competing SR method is depicted in Section~\ref{sec:experimental}, with a thorough justification of different modules of the proposed architecture. Finally, Section~\ref{sec:conclusion} concludes the study, summarizing the key findings of the work.

\subsection{Formulation of Single Image Super-Resolution (SISR)}
The image Super-Resolution (SR) is an off-line algorithm to improve the spatial resolution of a given LR observation. 
The most used models for SR reconstruction include blur, down-sampling and noise in the image formulation \cite{brief}. Here, LR image $(LR_{gen}) $ is degraded due to blur, noise and down-sampling effects and can be represented as,
\begin{equation}
	LR_{gen} = (HR_{conv}\circledast k)\downarrow s + \eta,
\end{equation}
where, $HR_{conv}\circledast k$ is the convolution between kernel $k$ and the HR image $HR_{conv}$, $\downarrow s$ is the down-sampling with the scale factor $s$ and $\eta$ is the independent and identical distributed noise. Thus, the task of Single Image SR (SISR) is to learn inverse estimating $HR_{conv}$ from $LR_{gen}$ which is severely ill-posed in nature. Thus, the down-sampling removes the high-frequency information from the input image, it is very challenging to reconstruct all fine details with pixel-wise accuracy. Therefore, SR algorithms learn to produce the mean of all possible texture to minimize the loss for output image.

\section{Related Work}
\label{sec:related work}

It is evident that deep learning-based SR techniques have achieved significant success in comprehending the complex nonlinear relationships between LR-HR images, especially when compared to conventional SR methods. These methods are largely based on deep Convolutional Neural Network (CNN) and they are frequently reported for natural \cite{SRCNN, VDSRN, EDSR, RCAN, SRGAN} and medical images \cite{MRI1, MRI2, CT1, CT2} in literature. The foundational deep learning-based SR model, Super-Resolution Convolutional Neural Network (SRCNN) \cite{SRCNN}, enabled end-to-end learning for SR tasks on natural RGB images. Successive models, such as VDSR \cite{VDSRN} and DRCN \cite{recursive}, improved upon this by deepening the network architecture and employing recursive layers to enhance feature extraction while maintaining parameter efficiency. EDSR \cite{EDSR} further optimized residual learning by eliminating batch normalization layers. A major advancement came with SRGAN \cite{SRGAN}, which introduced adversarial training to generate perceptually rich SR outputs, effectively capturing fine textures and high-frequency details beyond traditional loss formulations.
% For instance, the pioneering deep learning-based SR model, Super-Resolution Convolutional Neural Network (SRCNN), was introduced by Dong et al. \cite{SRCNN}, establishing a foundation for end-to-end learning of SR tasks on natural RGB images. Building upon SRCNN, subsequent advancements have significantly do to improve reconstruction fidelity, computational efficiency, and model depth. The Very Deep Super-Resolution (VDSR) \cite{VDSRN} leveraged a deeper network architecture compared to SRCNN, enabling more expressive feature extraction. The Deeply-Recursive Convolutional Network (DRCN) \cite{recursive} introduced recursive convolutional layers to increase model depth without excessive parameter overhead, thereby enhancing learning capacity while maintaining parameter efficiency. Further improvements were made by the Enhanced Deep Super-Resolution Network (EDSR) \cite{EDSR}, which optimized residual learning techniques and removed unnecessary batch normalization layers to improve performance. A significant paradigm shift in SR occurred with the introduction of Super-Resolution Generative Adversarial Network (SRGAN) \cite{SRGAN}, which was the first to employ adversarial learning to generate perceptually high-quality SR images, capturing finer textures and high-frequency details beyond traditional pixel-wise losses. 
Building upon the aforementioned groundbreaking SR techniques for natural images, the subsequent sub-sections review the various SR methods applied to medical imaging modalities adopting supervised and unsupervised strategies. %In the proposed study, we introduce an unsupervised approach of specific medical modality i.e., WCE, hence we review the literature of different SR techniques leveraged for various medical modalities adopting supervised and unsupervised strategies. The summary of these SR techniques is depicted in Table~\ref{Tab:related work} for easy access in \emph{supplementary material}.
\vspace{-0.5cm}
\subsection{Supervised SR approaches}
The effectiveness of deep learning Single Image SR (SISR) methods that utilize supervised training for natural images has been extended in the field of medical imaging across multiple modalities, including Computed Tomography (CT), Magnetic Resonance Imaging (MRI), electron microscopy, colonoscopy, and endoscopy. A comprehensive survey of supervised SR approaches is presented in the Section II-A in \emph{Supplementary material}. 
% \vspace{-0.5cm}
\subsection{Unsupervised SR approaches}
A common limitation observed in the supervised SR approaches is the reliance on synthetically generated LR images, typically created by applying predefined degradation operators—such as bicubic downsampling—to HR images. This strategy restricts the SR models to learn only the characteristics of the artificial degradation process, making them less effective when confronted with unknown or complex degradations encountered in real-world clinical settings. While obtaining genuine LR-HR image pairs could address this limitation, it is largely infeasible in the medical imaging domain due to hardware constraints and the substantial cost and effort required to acquire accurately aligned, high-quality clinical datasets. To address this challenge, the concept of unsupervised SR was introduced by Lugamayr et al. \cite{lugmayr2020ntire} for natural RGB images. This approach enables SR without relying on paired training data by simultaneously learning the degradation and upsampling processes. In SRResCGAN \cite{SRResCGAN}, the authors trained a degradation network alongside an upsampling model in an adversarial setting. Similarly, USISResNet \cite{USISResNET} leverages a GAN-based framework for unsupervised domain adaptation in SR tasks, while Prajapati et al. \cite{dusgan} proposed an end-to-end GAN-based architecture for unsupervised SR. 

The concept of unsupervised SR has gained increasing attention in the medical imaging domain too, offering a practical solution where paired low- and high-resolution data are unavailable. For instance, Liu et al. \cite{MRI5} proposed an unsupervised degradation adaptation network for brain MRI, enabling SR enhancement without the need for paired data. Expanding on this, the same authors introduced uSRGR \cite{MRI6}, which jointly performs SR and Gibbs artifact removal for brain MRI images. Iwamoto et al. \cite{MRI7} explored a cross-modality strategy by leveraging complementary information from different imaging modalities to enhance brain MRI resolution. In the context of CT imaging, Zhang et al. \cite{CT6} presented a zero-shot unsupervised SR method; thus, authors introduced a SR method for reconstructing high-resolution CT images directly from low-resolution sinograms without requiring paired datasets. Additionally, Li et al. \cite{CT7} proposed KerSRGAN, a GAN-based model capable of achieving $\times4$ SR with improved perceptual quality. In other modalities, Cui et al. \cite{other2} developed an unsupervised multiscale GAN framework for arterial spin labeling, which not only enhanced spatial resolution but also effectively suppressed noise. Within the WCE domain, MDA-SR \cite{MDASR} employed a multi-level domain adaptation strategy to perform unsupervised SR using unpaired LR-HR image sets. Recently, the TUDASR network \cite{tudasr}, an encoder-decoder architecture, employs transformer-based blocks to enhance the quality of WCE images in an unsupervised manner, efficiently capturing long-range dependencies and high-frequency details. As mentioned earlier, the summary of different supervised and unsupervised SR methods for medical modalities is highlighted in Table~I in \emph{Supplementary material} and their key findings are depicted below.

\begin{itemize}
\item \textbf{Dependence on paired LR–HR training data:} Most existing SR methods \cite{zhang2020stereo, endodeep1, endodeep2} rely on supervised learning frameworks that require large-scale paired Low-Resolution (LR) and High-Resolution (HR) datasets. However, acquiring accurately aligned LR–HR pairs in Wireless Capsule Endoscopy (WCE) is highly challenging due to hardware constraints and varying acquisition conditions.
To address this limitation, the proposed method adopts an unsupervised learning paradigm, where LR WCE images and HR conventional endoscopy images are utilized in an unpaired setting. This eliminates the dependency on paired data while enabling effective learning of SR reconstruction.
\item \textbf{Degradation learning:} Many SR methods \cite{EndoL2h, dcan, conv1, conv2, SISRWCE} assume predefined degradation processes, such as bicubic downsampling, which fail to capture the complex, device-specific degradations present in real-world endoscopic images. This results in poor reconstruction quality and limited applicability in clinical scenarios.
To overcome this, the proposed framework incorporates a domain-adaptive degradation model that learns to generate WCE-like LR images from HR conventional endoscopy data using adversarial training. This enables the SR network to learn from realistic degradation patterns, improving robustness and reconstruction fidelity.
\item \textbf{Limited cross-domain generalization capability:} Existing SR \cite{EndoL2h, MDASR, tudasr} models are typically trained and evaluated on a single dataset or imaging modality, limiting their ability to generalize across different domains and imaging conditions.
In contrast, the proposed approach performs feature-level domain alignment through adversarial learning, reducing the distribution gap between conventional endoscopy and WCE images. This enhances the generalization capability of the model across diverse datasets and imaging modalities.
\item \textbf{Inadequate modeling of global contextual dependencies:} Conventional CNN-based SR methods are inherently limited by local receptive fields, making them less effective in capturing long-range spatial dependencies required for reconstructing complex anatomical structures.
To address this limitation, the proposed framework integrates transformer-based modules within the Deep Attention Blocks (DABs), enabling the network to capture both global contextual information and local spatial details, leading to improved structural consistency in the reconstructed images.
\item \textbf{Suboptimal recovery of fine anatomical details and textures:} Existing methods often struggle to accurately reconstruct fine-grained structures such as mucosal patterns and vascular details, which are critical for clinical interpretation.
The proposed method employs a Fusion Attention Block (FAB) that combines bicubic-upsampled inputs with learned high-frequency features using channel and spatial attention mechanisms. This allows the network to focus on diagnostically relevant regions and enhances the reconstruction of fine anatomical details.
    % \item The most of the unsupervised SR methods are applied to traditional endsoscopy images with supervised learning \cite{zhang2020stereo, endodeep1, endodeep2}. They rely on large-scale paired datasets for training, which are scarce in the medical imaging domain, particularly for endoscopic images.
    % %Most of the unsuervhese methods directly utilize existing deep learning frameworks as the black box to learn the mapping from LR endoscopic image to HR endoscopic image, ignoring the spatial interdependency among the pixels. The relationship among the pixels is essential, reflecting the location and size of the potential cancer polyps for clinicians. 

    % \item Many existing SR methods \cite{EndoL2h, dcan, conv1, conv2, SISRWCE} assume known or simplified degradation models (i.e., bicubic downsampling), which do not accurately reflect the complex and device-specific degradations present in real-world endoscopic images. This results in poor generalizability when applied to clinical data acquired under uncontrolled imaging conditions.
    % \item The cross-domain adaptability remains underexplored. Almost all SR models are developed and validated within a single imaging modality or dataset, lacking the ability to generalize to other types of medical images (i.e., retinal or MRI data), which limits their practical deployment in diverse clinical environments.
\end{itemize}

\section{Proposed Framework:\emph{CEM-TUDASR}}
\label{sec:proposed framework}

The proposed unsupervised SR framework (i.e., \emph{CEM-TUDASR}) addresses the above-listed inherent challenges associated with enhancing WCE images by performing $\times4$ upscaling without relying on paired training data. As mentioned earlier, this is an extension of our earlier work named TUDASR \cite{tudasr}. The major effective modifications included in the current version in comparison to TUDASR \cite{tudasr} are highlighted in Section~III in the \emph{Supplementary material} due to space constraints.

\begin{figure}[!t]
    \centering
    \includegraphics[width=0.49\textwidth, height=5cm]{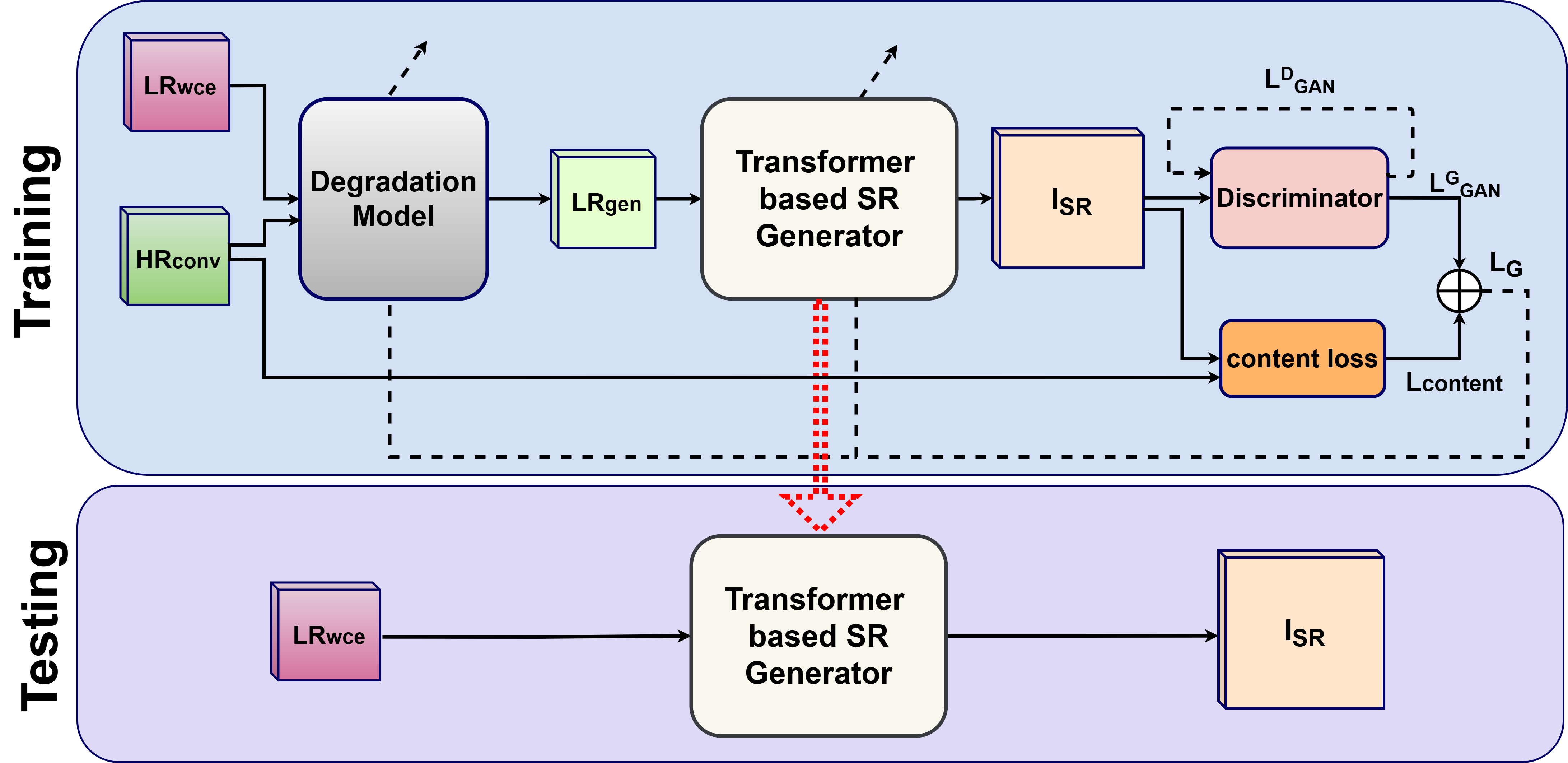}
    %\vspace{-0.4cm}
    \caption{The block schematic of the proposed unsupervised SR method:\emph{CEM-TUDASR}.}   \label{proposed}
    \vspace{-0.7cm}
\end{figure}

In Fig.~\ref{proposed}, the overall architecture of the proposed work,
\emph{CEM-TUDASR} is depicted that comprises of two prominent components: (i) a Degradation model and (ii) a Transformer based SR network. The degradation model is adopted from the TUDASR pipeline \cite{tudasr}. It is employed to learn a degradation prior that maps HR conventional endoscopy images (i.e., $HR_{conv}$) to synthetic LR images (i.e., $LR_{gen}$) during training process, mimicking the degradation characteristics of real WCE images (i.e., $LR_{wce}$). Unlike conventional endoscopic data, WCE images suffer from modality-specific degradations arising from factors such as non-uniform illumination, limited optics, and motion artifacts. To effectively model this domain-specific degradation, the degradation network is trained in an adversarial manner using a GAN framework, wherein a discriminator enforces that the generated $LR_{gen}$ images are perceptually aligned with the true $LR_{wce}$ samples through carefully designed loss constraints. Subsequently, the generated $LR_{gen}$ images serve as input to the SR network, which aims to reconstruct SR outputs at an upscaling factor of $\times4$. To validate the effectiveness of the proposed degradation model in bridging the domain gap between conventional endoscopy and WCE images, we conduct a t-SNE-based feature space analysis and same is illustrated in Fig.~2 in the \emph{Supplementary material}. 

Further, the SR network incorporates a fusion attention-based generator that integrates bicubic-upsampled inputs with learned features, enabling the recovery of fine-grained textures and structural details. Additionally, a discriminator is employed at the feature level to ensure semantic consistency between WCE and conventional domains, thereby facilitating effective domain adaptation. This feature-level adversarial alignment plays a critical role in bridging the distribution gap between the two modalities, leading to improved reconstruction fidelity. Overall, the combination of domain-aware degradation modeling and attention-guided feature fusion allows the proposed framework to generate high-quality SR outputs that preserve diagnostically relevant information, enhancing the clinical utility of WCE imagery. During the inference phase, the proposed framework (i.e., \emph{CEM-TUDASR}) requires only the real LR WCE images (i.e., $LR_{wce}$) as input. These are directly fed into the pre-trained SR generator to produce the final SR outputs. Importantly, although the generator is trained using synthetically degraded images derived from HR conventional endoscopy data, it demonstrates strong generalization to real-world WCE inputs. This effectiveness is attributed to the domain alignment facilitated by the adversarially trained degradation model, which bridges the distribution gap between synthetic and actual WCE degradations. Consequently, the proposed architecture eliminates the dependence on paired LR-HR training data and delivers super-resolved outputs with improved structural fidelity and perceptual realism, thereby enhancing its applicability in real clinical workflows. The detailed procedural steps of the proposed unsupervised SR framework are summarized in Algorithm 1 for clarity and reproducibility in \emph{Supplementary material}.
\vspace{-0.38cm}
\subsection{Transformer-based SR Generator Network}

\begin{figure}[!t]
    \centering
  \includegraphics[width=0.49\textwidth, height=4.5cm]{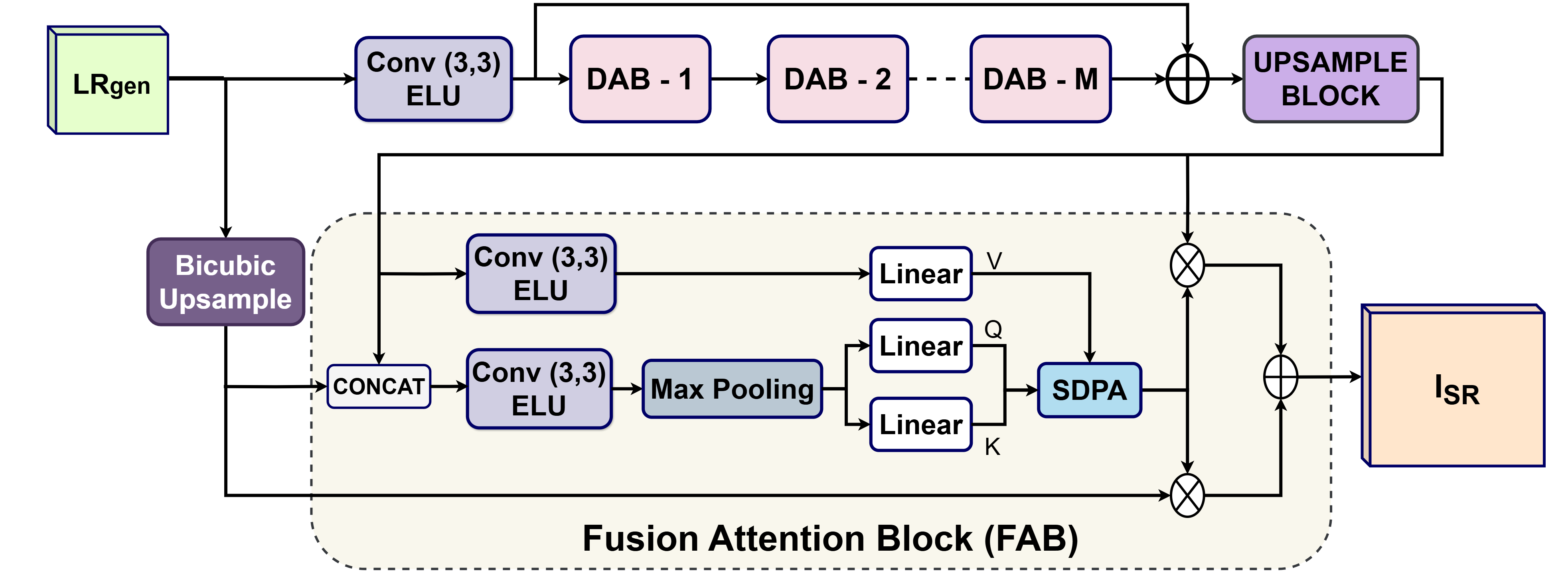}
  \vspace{-0.6cm}
    \caption{ The network architecture of Transformer-based SR Generator in the proposed method: \emph{CEM-TUDASR}.}
    \label{SR_generator}
      \vspace{-0.4cm}
\end{figure}

\begin{figure}[!t]
    \centering
  \includegraphics[width=0.49\textwidth, height=10cm]{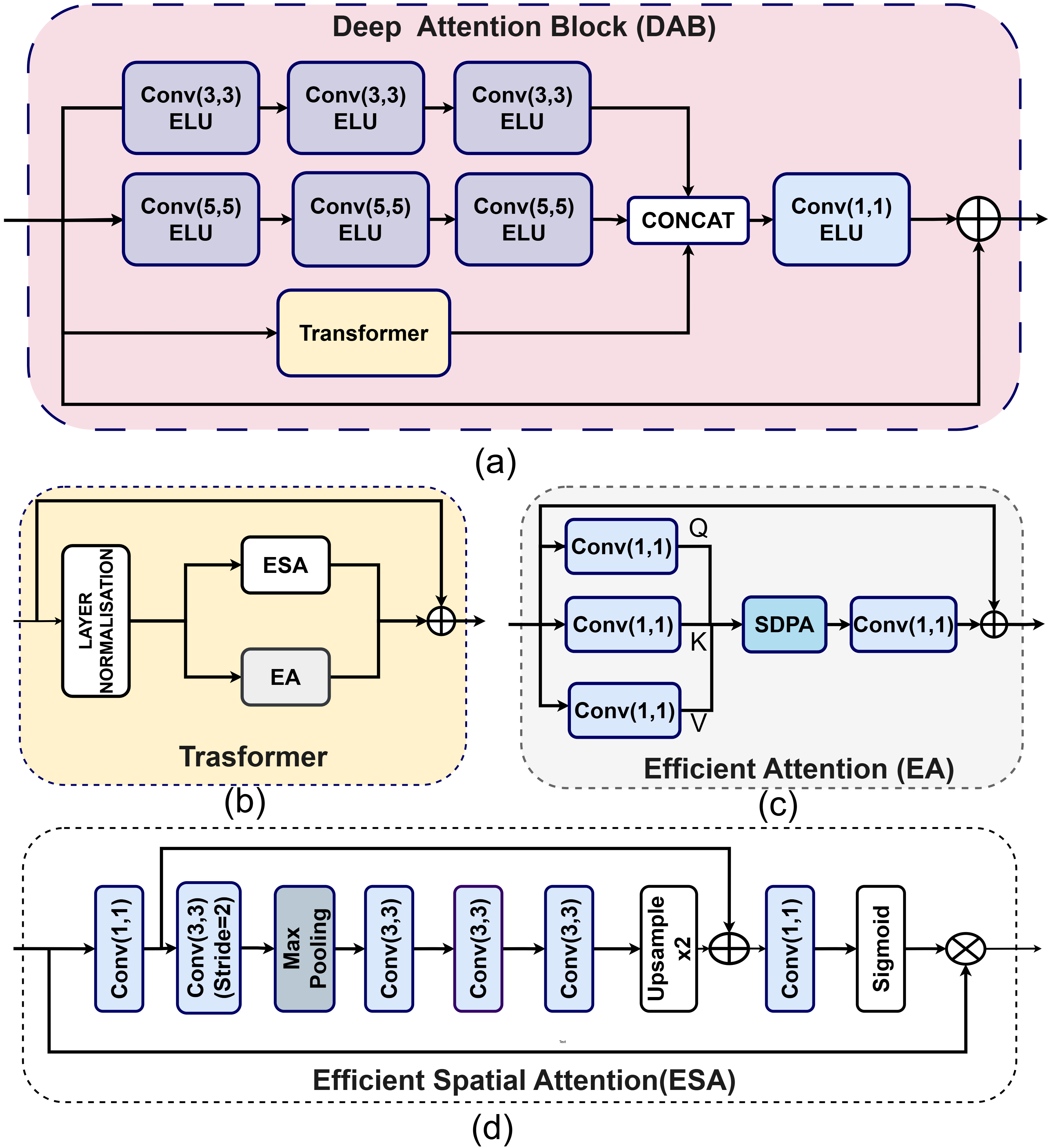}
  \vspace{-0.6cm}
    \caption{The network architecture of different blocks used in Transformer-based SR Generator in the proposed method: \emph{CEM-TUDASR}. (a) Deep Attention Block (DAB), (b) Transformer, (c) Efficient Attention (EA) and (d) Efficient Spatial Attention (ESA).}
    \label{DAB}
      %\vspace{-0.7cm}
\end{figure}

The proposed SR generator network is attributed to enhance LR WCE images by an upscaling factor of $\times4$, with a focus on preserving structural integrity and improving perceptual quality. As depicted in Fig.~\ref{SR_generator}, the generator comprises four pivotal components: an initial feature extraction module, a sequence of Deep Attention Blocks (DABs) for hierarchical feature refinement, a Fusion Attention Block (FAB) for integrating complementary feature representations, and an upsampling module to reconstruct HR outputs.

\subsubsection{Initial Feature Extraction}

The LR image generated by the degradation network is passing through an initial 2D depthwise convolutional layer with a kernel size of $3\times3$ followed by an ELU activation to extract shallow features from the input LR image. This operation can be formally defined as: \vspace{-0.3cm}
\begin{equation}
    F_{0}= ELU(CONV_{3\times3}(LR_{gen})),
\end{equation}
where, $LR_{gen}$ is the input LR image generated by the degradation model and $F_{0}$ denotes the initial feature representation.

\subsubsection{Deep Attention Block (DAB)}
The output from the initial feature extraction stage (i.e., $F_{0}$) is subsequently processed through a series of Deep Attention Blocks (DABs), which are specifically designed to enhance the feature representation of WCE images by modeling both spatial locality and contextual dependencies across channels. Each DAB serves as a crucial architectural component that hierarchically captures both fine-grained mucosal textures and broader anatomical structures while preserving spatial coherence essential for accurate clinical interpretation in WCE imagery. 
%\vspace{0cm}
As illustrated in Fig.~\ref{DAB}(a), each DAB integrates multiple convolutional pathways with varying receptive fields along with an embedded Transformer module to jointly address local detail recovery and global context modeling. Here, the first convolutional branch comprises three successive convolutional layers with a kernel size of $3 \times 3$, each followed by an Exponential Linear Unit (ELU) activation. This branch focuses on extracting localized features such as vascular textures, polyp edges, and fine mucosal structures by maintaining a small receptive field. In contrast, the second branch utilizes three convolutional layers with a $5 \times 5$ kernel size and ELU activations to capture broader contextual information, such as illumination gradients and organ-level shape variations—features critical for understanding spatial continuity in WCE sequences. These dual branches transform the input into complementary feature subspaces, facilitating diverse and robust representation learning tailored to the complex texture distribution in WCE images. To further enrich the contextual understanding, a Transformer module is embedded within each DAB, enabling the network to model long-range spatial dependencies that are often present across adjacent regions in WCE frames. The outputs from the two convolutional branches and the Transformer unit are concatenated along the channel dimension to produce a rich multi-scale representation. Finally, a $1 \times 1$ convolution layer with ELU activation is then applied to fuse and project this high-dimensional feature tensor, acting as both a channel mixer and dimensionality reducer. In last, a residual connection is added between the input and output of the DAB to form a residual attention unit, which enhances gradient flow and encourages effective reuse of learned features during training.

Further, the architecture of the transformer block used in DAB is depicted in Fig.~\ref{DAB}(b). Here, the attention mechanism helps the model focus on critical regions such as mucosal folds, vascular structures, or lesions, which are essential for accurate diagnosis.  It comprises the layer normalization, the parallel branch of Efficient Attention (EA) module and Efficient Spatial Attention (ESA) module. The layer normalization is used for stabilizing training and ensuring scale-invariant feature representations. As depicted in Fig.~\ref{DAB}(c), Efficient Attention (EA) module employs lightweight $1\times1$ convolutions to compute the Query ($Q$), Key ($K$), and Value ($V$) matrices. These are passed through a Scaled Dot-Product Attention (SDPA) unit to compute contextual attention maps, followed by projection and residual addition. The attention score is then calculated using the Scaled Dot-Product Attention (SDPA) \cite{transformer}:
\begin{equation}
    \text{Attention}(Q, K, V) = \text{softmax}\left( \frac{QK^T}{\sqrt{d_k}} \right) V,
\end{equation}
where ${d_k}$ is the dimensionality of the key vector. This allows the model to weigh spatially relevant features while mitigating irrelevant information. In addition to the EA, the Efficient Spatial Attention (ESA) module is introduced to refine the spatial resolution of critical regions within the feature map (See Fig.~\ref{DAB}(d)). The ESA module captures spatial saliency through downsampling (via max pooling), followed by a series of convolutions and upsampling via pixel shuffle, concluding with a sigmoid gate for spatial importance modulation.  The outputs from ESA and EA are fused via residual summation to produce a context-aware global feature map, which is then added to the original input.

\subsubsection{Upsample Block}
The output of the final DAB block is then fed into the upsample block to achieve the required $\times 4$ spatial enlargement while preserving the rich texture details learned by the DAB. It contains two‐stage sub-pixel convolution. In each stage, feature channels are first expanded via a small convolution and then reorganized through a pixel‐shuffle operation, effectively learning to redistribute high-frequency details into a larger spatial grid without introducing checkerboard artifacts. This enhanced representation is subsequently delivered to the Fusion Attention Block (FAB) for final refinement and integration with the bicubic baseline, ensuring both global structure and fine textures are optimally preserved in the SR output.

\subsubsection{Fusion Attention Block (FAB)}
The Fusion Attention Block (FAB) plays a pivotal role in enhancing the performance of the SR network for WCE images. As displayed in Fig.~\ref{SR_generator}, the FAB takes two inputs: (i) the feature map output from the upsampling module, which contains high-frequency details extracted from the network, and (ii) the bicubic-upsampled version of the original LR image. By integrating this, the FAB ensures both global context and fine details are preserved. This module employs channel-wise and spatial attention mechanisms to selectively focus on diagnostically relevant areas, such as blood vessels, lesions, and mucosal folds, while suppressing irrelevant background regions. These are concatenated along the channel dimension and passed through a $3 \times 3$ convolution layer followed by an ELU activation to generate a unified feature map. This feature maps are passed through the max-pooling and then independent linear layers to generate the Query ($Q$) and Key ($K$) matrices for attention computation. The output of upsample block is applied to $3 \times 3$ convolution followed by ELU activation to produce a linear projection for the Value ($V$) component of attention. These attention components are then fed into a Scaled Dot-Product Attention (SDPA) module that models dependencies between spatial locations. The output of the SDPA module, representing context-aware weighted features, is element-wise multiplied with the initial value features. The result is then added to the original fused features through a residual connection, ensuring both stability \& feature reuse and produce SR output.  This allows for better feature refinement, improving both structural integrity and fine-texture details which are crucial for accurate diagnosis. The attention-guided fusion within the FAB effectively enhances the quality of the SR image, ensuring that critical regions are enhanced for clinical relevance while maintaining computational efficiency. This makes the FAB a key component in producing high-quality, clinically useful WCE images, essential for detecting anomalies such as bleeding and polyps.

\subsection{Loss Functions}
This section details the individual loss functions employed within the proposed unsupervised SR framework (i.e., \emph{CEM-TUDASR}).

\subsubsection{Generator ($G$)}
To optimize the effectiveness of the SR generator network, \emph{CEM-TUDASR} incorporates a weighted fusion of loss functions. Mathematically, the total loss (i.e., $L_G$ ) can be written as: \vspace{-0.2cm}
\begin{equation}
    \mathcal{L}_{\text{G}} = \lambda_{1} \mathcal{L}_{\text{content}} + \lambda_{2} \mathcal{L}^{G}_{\text{GAN}},
    \label{srloss}
\end{equation}
where $\lambda_{1}$ and $\lambda_{2}$ are the weight of each loss.  $\mathcal{L}_{content}$ and $\mathcal{L}^{G}_{\text{GAN}}$ are the content loss and generator adversarial loss with discriminator, respectively. To preserve critical details such as texture, sharpness, and high-frequency features in the SR image the content loss is applied between $HR_{conv}$ and $SR$ images which is defined as, \vspace{-0.2cm}
\begin{equation}
\mathcal{L}_{\text{content}} = \| {I_{\text{SR}}} -{HR_{\text{conv}}} \|_1.
\label{content}
\end{equation}
Here, ${I_{\text{SR}}}$ is the SR image. In addition to the content loss, the SR network also incorporates a standard generator adversarial loss which is defined as, \vspace{-0.2cm}
\begin{equation}
    \label{advg}
    \mathcal{L}_{GAN}^{G}=\frac{1}{N}\sum_{}^{N}\left| 1-\mathrm{D}^{}\left( \mathrm{I_{\text{SR}}} \right) \right|,
\end{equation}
where $\mathrm{D}^{}  (\cdot)$ indicates the function of discriminator. $N$  stands for the size of the training batch.

\subsubsection{Discriminator ($D$)}
The discriminator is trained in an adversarial manner to distinguish the generated SR image from the HR image. To stabilize the adversarial training process, the Least Squares GAN (LSGAN) loss is utilized for Discriminator, which is formulated as:
\begin{equation}
    \label{advd}
    \mathcal{L}_{GAN}^{D}=\frac{1}{N}\sum_{}^{N}\left( \frac{| \mathrm{D}{}({I_{\text{SR}}})| +\left| 1-\mathrm{D}{}({HR}_{conv}^{}) \right|}{2} \right),
\end{equation}
where, $D(\cdot)$ denotes the function of Discriminator. 
\vspace{-0.3cm}
\section{Experimental Analysis}
\label{sec:experimental analysis}

In this section, we present a comprehensive evaluation of the proposed unsupervised SR framework (i.e., \emph{CEM-TUDASR}), comparing its performance against several state-of-the-art unsupervised SR techniques for an upscaling factor of $\times 4$. To rigorously assess reconstruction quality, both qualitative and quantitative analyses are performed. For qualitative evaluation, representative image patches from the outputs of the competing models are visually compared to demonstrate the capability of the proposed method in preserving structural integrity and perceptual realism. Additionally, the quantitative evaluation is carried out using widely recognized no-reference image quality metrics, namely the Blind/Referenceless Image Spatial Quality Evaluator (BRISQUE) \cite{brisque}, Natural Image Quality Evaluator (NIQE) \cite{niqe}, and Perception-based Image Quality Evaluator (PIQE) \cite{piqe}. Notably, we utilize EndoQM, a domain-specific perceptual quality metric tailored explicitly for assessing diagnostic quality in super-resolved capsule endoscopy images.

Further, to evaluate the generalization ability of the proposed approach, we validate its performance on well-established publicly available clinical datasets such as KID \cite{KID} and GIANA \cite{GIANA}, thereby confirming its robustness and suitability across diverse endoscopic imaging scenarios. Importantly, we also perform cross-domain evaluations, extending our assessments to retinal imaging dataset \cite{retinal} to demonstrate the adaptability and effectiveness of our method beyond capsule images. In addition, an Analysis of Variance (ANOVA) test is performed to statistically verify the significance of observed performance improvements over existing methodologies. An extensive ablation study is also conducted to systematically analyze the contributions of critical architectural components, including different loss functions and network configurations.   Lastly, to demonstrate the computational efficiency of the proposed method, we perform a comparative analysis of the number of trainable parameters and Multiply-Add operations (Multi-Adds) against those of the baseline and state-of-the-art models discussed above. To further demonstrate the practical feasibility and deployment efficiency of the proposed CEM-TUDASR framework, a detailed layerwise computational complexity analysis is performed in terms of trainable parameters, Floating Point Operations (FLOPs), Multiply–Accumulate Operations (MACs), and execution time. The detailed experimental setup, evaluation protocols, and comprehensive results are elaborated upon in the subsequent sections.
\vspace{-0.4cm}
\subsection{Dataset and training details}
As previously discussed, \emph{CEM-TUDASR} is trained in an unsupervised manner utilizing unpaired LR and HR images. A key novel contribution of this study is the creation of a refined derivative dataset specifically tailored for SR tasks, derived from the publicly available Kvasir Capsule Endoscopy dataset \cite{kvasir}, which consists exclusively of Wireless Capsule Endoscopy (WCE) images. Table~\ref{tab:dataset} provides detailed descriptions of the datasets used, including the number of images, their resolution, and the corresponding splits for training, validation, and testing. Originally, the Kvasir dataset comprised a total of $47,236$ RGB images, each with a resolution of $336 \times 336$ pixels, categorized according to various medical anomalies. Due to redundancy and the presence of unwanted border regions in the original dataset, extensive manual pre-processing was performed to curate a high-quality subset suitable for SR tasks. The resulting curated dataset contains $10,000$ images for training, $550$ images for validation, and $1,000$ images for testing, each resized to $280 \times 280$ pixels after the removal of redundant boundary pixels. To facilitate unsupervised training, an additional dataset comprising $10,000$ conventional endoscopy images \cite{conventional}, initially with a resolution of $1024 \times 1024$ pixels, was also curated by removing extraneous border pixels, reducing their size to $838 \times 838$ pixels. It is important to emphasize that the training dataset does not contain true LR-HR image pairs; rather, the LR images originate from WCE while the HR images are sourced from conventional endoscopy, ensuring genuinely unpaired image data for effective unsupervised training. The testing dataset comprises of $1,000$ WCE LR samples with an image resolution of $280 \times 280$ pixels. Furthermore, to comprehensively evaluate the generalization capability and robustness of the proposed CEM-TUDASR framework, additional external testing datasets, namely the KID dataset \cite{KID}, GIANA dataset \cite{GIANA}, and a retinal image dataset \cite{retinal}, were also utilized during evaluation. It is important to note that these datasets were not used during training and were exclusively employed for testing purposes. The KID dataset contains 500 WCE images with a resolution of 360×360 pixels, while the GIANA dataset consists of 50 WCE images with a resolution of 576×576 pixels. In addition, 50 retinal images with a resolution of 700×605 pixels were utilized to assess the cross-domain adaptability of the proposed framework across different medical imaging modalities.

\begin{table*}[]
\centering
\caption{Dataset description and data splits used for training and evaluation of the proposed CEM-TUDASR framework.}
\label{tab:dataset}
\begin{tabular}{|l|c|c|c|c|c|}
\hline
\textbf{Dataset} & \textbf{Image Type} & \textbf{Resolution} & \textbf{Train} & \textbf{Validation} & \textbf{Test} \\ \hline
Kvasir Capsule \cite{kvasir} & LR (WCE) & $280 \times 280$ & 10,000 & 550 & 1,000 \\ \hline
Conventional Endoscopy \cite{conventional}& HR & $838 \times 838$ & 10,000 & Not used* & Not used* \\ \hline
KID Dataset \cite{KID}& LR & $360 \times 360$ & Not used* & Not used* & 500 \\ \hline
GIANA Dataset \cite{GIANA} & LR & $576 \times 576$ & Not used* & Not used* & 50 \\ \hline
Retinal Dataset \cite{retinal} & LR & $700 \times 605$ & Not used* & Not used* & 50 \\ \hline
\end{tabular}\\
\footnotesize{
*The Conventional Endoscopy dataset is used only for HR-domain training, whereas the KID, GIANA, and Retinal datasets are not utilized during training and exclusively employed for external testing and cross-domain evaluation.}
\end{table*}

In the training process, a patch-based discriminator with normalization is employed, and the degradation model structure is inspired by the degradation network proposed in TUDASR \cite{tudasr}. The parameters $\lambda_{1}$ and $\lambda_{2}$ in Eq.(\ref{srloss}) are set to be $1$ and $0.01$, respectively. Further, an Adam optimizer was used during training with a batch size of $4$, a learning rate of $0.5 \times 10^{-5}$ for SR model and $1 \times 10^{-5}$ for down-sampling model. The number of DABs used in proposed method is $M=5$. To maintain simplicity, enhance training stability, and reduce computational complexity, the degradation model is independently trained for the first $50$ epochs. Subsequently, the SR network and degradation model are jointly trained for an additional 50 epochs.

\vspace{-0.4cm}
\subsection{Endoscopy Quality Metric (EndoQM)}
%\textcolor{red}{lets put our first paper on arxiv and then we will put ref from that paper. This is my thought, If you know any other option where we can write it directly without puting our paper on arxiv then we can discuss}
To quantitatively evaluate the perceptual quality of the reconstructed SR images in the absence of ground truth high-resolution references, we employ a set of widely used No-Reference Image Quality Assessment (NR-IQA) metrics, namely BRISQUE
\cite{brisque}, PIQE \cite{piqe}, and NIQE \cite{niqe}. These metrics are particularly suitable for the present study, as obtaining accurately aligned LR–HR image pairs in Wireless Capsule Endoscopy (WCE) is inherently challenging.
The Blind/Referenceless Image Spatial Quality Evaluator (BRISQUE) assesses image quality by modeling Natural Scene Statistics (NSS) in the spatial domain. It captures deviations from statistical regularities observed in high-quality natural images, thereby quantifying distortions such as noise, blur, and compression artifacts without requiring a reference image. The Perception-based Image Quality Evaluator (PIQE) evaluates image quality based on block-wise distortion analysis, focusing on perceptually noticeable artifacts such as blocking effects, blurriness, and noise in localized regions. Unlike BRISQUE, PIQE does not rely on training and directly analyzes distortion in spatial patches, making it suitable for real-world degraded images. The Natural Image Quality Evaluator (NIQE) is another unsupervised NSS-based metric that estimates image quality by measuring the statistical deviation of an image from a model built on pristine natural images. NIQE operates without any human opinion scores and is widely used for evaluating blind image quality. For all these metrics, lower scores indicate better perceptual quality.
% In addition to the above-mentioned general-purpose NR-IQA metrics, we employ EndoQM, a domain-specific image quality assessment metric tailored for endoscopic imaging. Unlike generic NSS-based metrics, EndoQM is designed to capture the unique structural and textural characteristics of endoscopic data, such as mucosal patterns, vascular structures, and illumination variations commonly observed in WCE images. This makes it more suitable for evaluating clinically relevant image quality, particularly in scenarios where subtle anatomical details are critical for diagnosis. By combining general-purpose perceptual metrics (BRISQUE, PIQE, NIQE) with a domain-specific measure (EndoQM), we ensure a comprehensive evaluation of the proposed super-resolution framework in terms of both perceptual fidelity and clinical relevance. 

Additionally, to quantize the effective of the proposed SR method, we incorporate evaluation with novel no-reference image quality metric—Endoscopy Quality Metric (EndoQM)—specifically developed to assess the perceptual quality of SR endoscopic images. Traditional no-reference metric such as NIQE \cite{niqe} has reveal strong performance in evaluating natural images; however, they struggle to generalize to the unique characteristics of medical imagery, particularly endoscopic frames that differ significantly in texture, noise patterns, and color distributions. As illustrated Fig.~1 of the \emph{Supplementary material}, the constrained statistical distribution of WCE images poses challenges for the direct application of NIQE. To overcome this limitation, EndoQM is built upon the NIQE framework but trained on a curated dataset of endoscopic images, enabling it to capture modality-specific features and distortions that are clinically meaningful. This adaptation allows EndoQM to effectively assess high-frequency details, structural consistency, and texture fidelity without the need for reference HR images making it suitable for real-world SR scenarios in clinical settings. Compared to other metrics like BRISQUE, which depends on natural scene statistics, or PIQE, which tends to overemphasize local pixel distortions, EndoQM leverages the multivariate Gaussian modeling in NIQE to better align with the perceptual standards required in medical diagnostics. This makes EndoQM a robust and domain-aligned tool for evaluating SR performance in WCE and other endoscopic imaging applications.

\subsection{Qualitative Analysis}
This section presents a comprehensive qualitative analysis of the proposed method (i.e., \emph{CEM-TUDASR}) using the newly curated Kvasir-derived dataset, specifically designed for SR evaluation task. Further, to assess the generalization capability of the proposed model (i.e., \emph{CEM-TUDASR}), additional experiments are performed on the KID dataset \cite{KID} and GIANA \cite{GIANA}, which serves as an external benchmark comprising endoscopic images with distinct visual and structural characteristics. Importantly, the KID and GIANA dataset remains excluded from the training phase, ensuring an unbiased evaluation of the model’s cross-domain generalization capability. The performance of the proposed unsupervised SR model is rigorously evaluated and benchmarked against several state-of-the-art unsupervised SR methods, including ZSSR \cite{zssr}, DASR \cite{arr41}, dSRVAE \cite{dSRVAE}, DUSGAN \cite{dusgan}, BSRGAN \cite{bsrgan}, MDASR \cite{MDASR}, and TUDASR \cite{tudasr} with an upscaling factor of $\times 4$. It is important to highlight here that, among these baseline methods, MDASR and TUDASR was originally designed and trained specifically for WCE image enhancement, whereas the other methods— ZSSR, DASR, dSRVAE, DUSGAN, and BSRGAN—were initially developed for natural image unsupervised SR tasks. To ensure a fair and unbiased comparison, we re-trained these natural image-based models (ZSSR, DASR, dSRVAE, DUSGAN, and BSRGAN) using WCE image data prior to conducting the evaluations. The subsequent qualitative and quantitative comparisons presented herein systematically demonstrate the effectiveness and superior performance of the proposed method in preserving structural integrity and enhancing diagnostic perceptual quality in super-resolved WCE images.
\begin{figure*}[t!]
    \centering
    \renewcommand{\arraystretch}{1.2} 
    
    \begin{minipage}{0.20\linewidth} 
        \centering
        \subfloat[\scriptsize LR]{
            \includegraphics[width=\linewidth]{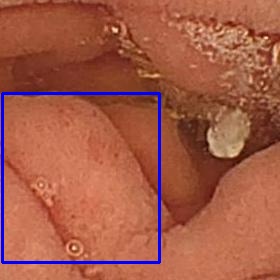}
        }
    \end{minipage}
    \begin{minipage}{0.78\linewidth}
        \centering
        % First row (4 images)
        \subfloat[\scriptsize ZSSR \cite{zssr}]{
            \includegraphics[width=0.24\linewidth]{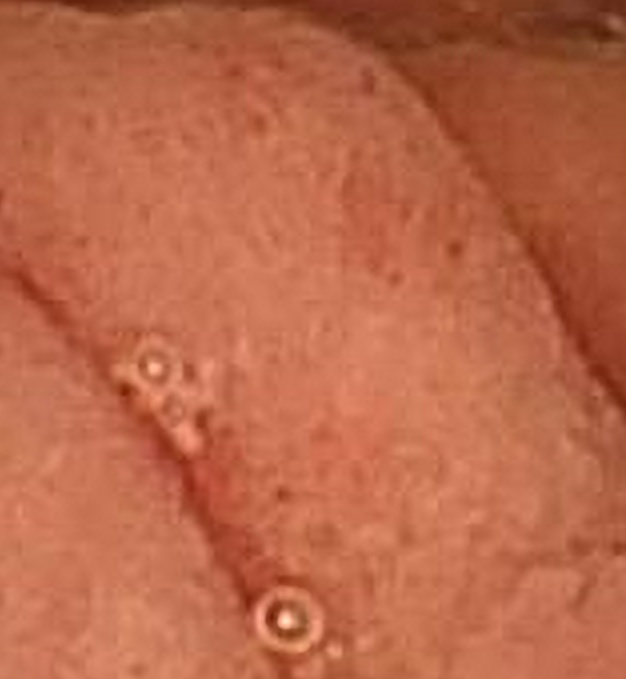}
        }
        \subfloat[\scriptsize DASR \cite{arr41}]{
            \includegraphics[width=0.24\linewidth]{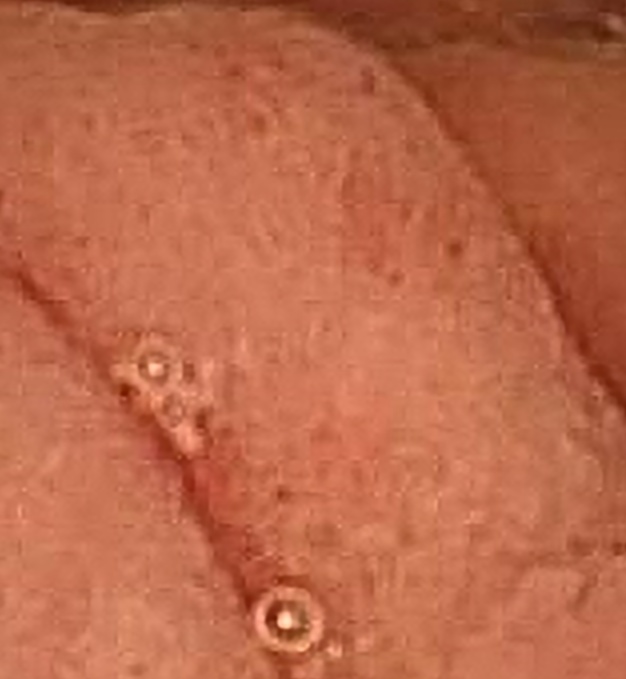}
        }
        \subfloat[\scriptsize dSRVAE \cite{dSRVAE}]{
            \includegraphics[width=0.24\linewidth]{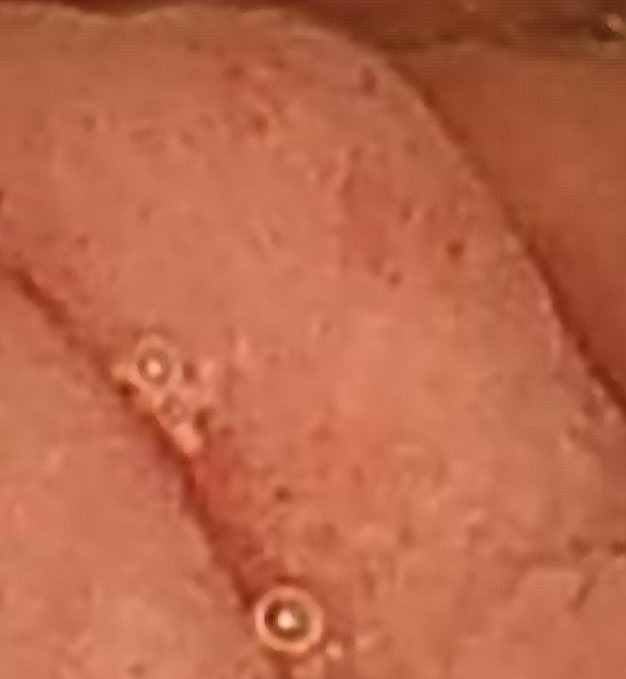}
        }
        \subfloat[\scriptsize DUSGAN \cite{dusgan}]{
            \includegraphics[width=0.24\linewidth]{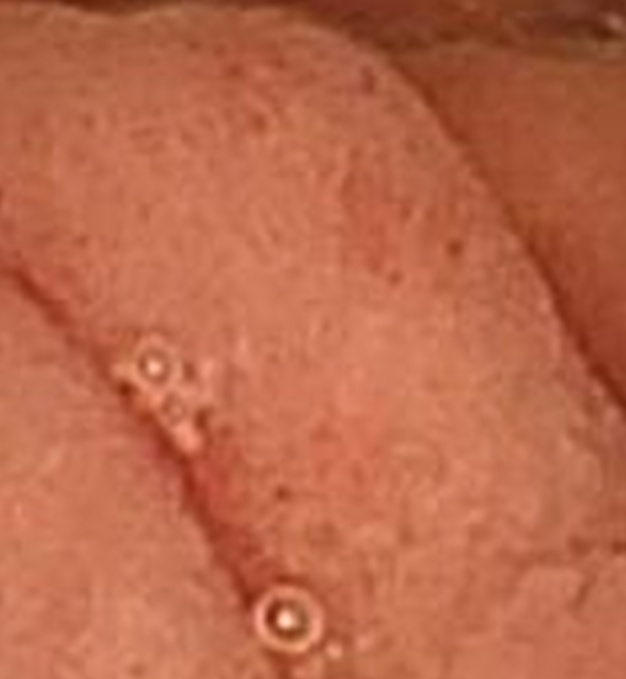}
        }
        \vspace{-0.3cm}\\
        % Second row (4 images)
        \subfloat[\scriptsize BSRGAN \cite{bsrgan}]{
            \includegraphics[width=0.24\linewidth]{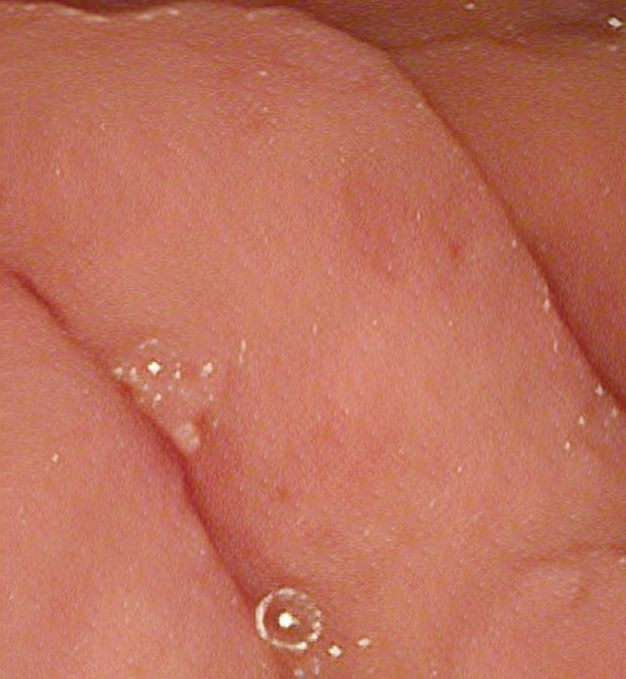}
        }
        \subfloat[\scriptsize MDASR \cite{MDASR}]{
            \includegraphics[width=0.24\linewidth]{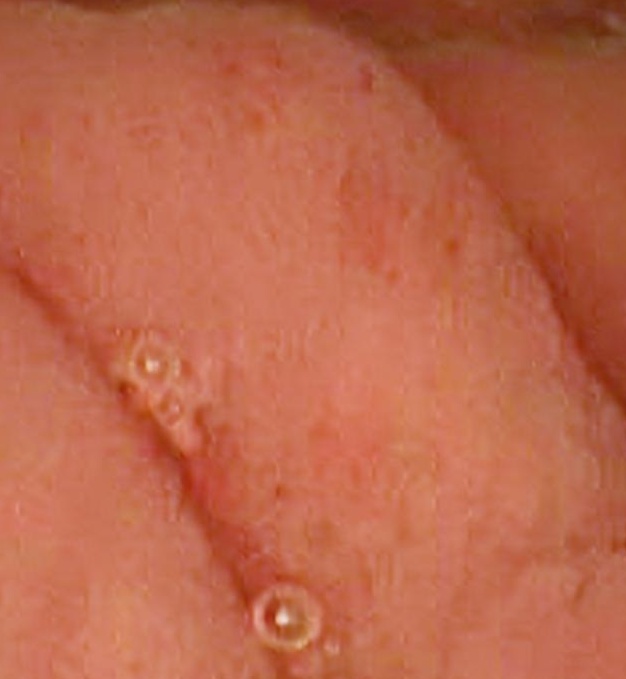}
        }
        \subfloat[\scriptsize TUDASR \cite{tudasr}]{
            \includegraphics[width=0.24\linewidth]{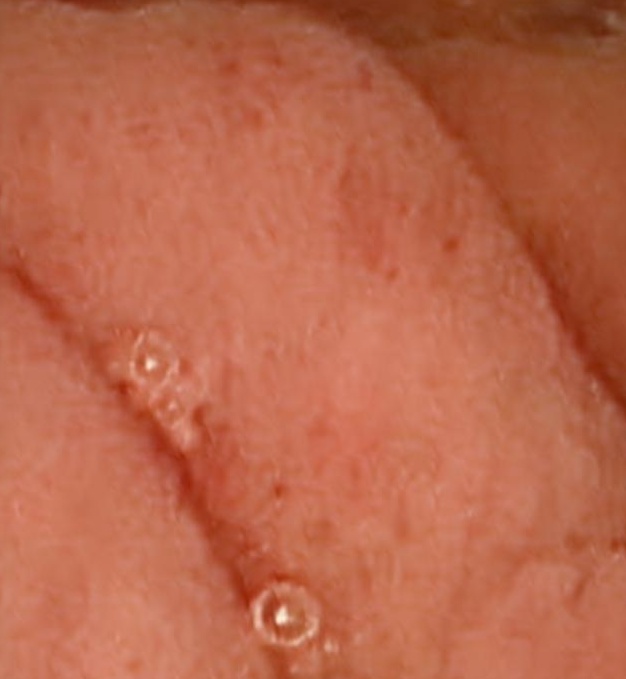}
        }
        \subfloat[\scriptsize Proposed]{
            \includegraphics[width=0.24\linewidth]{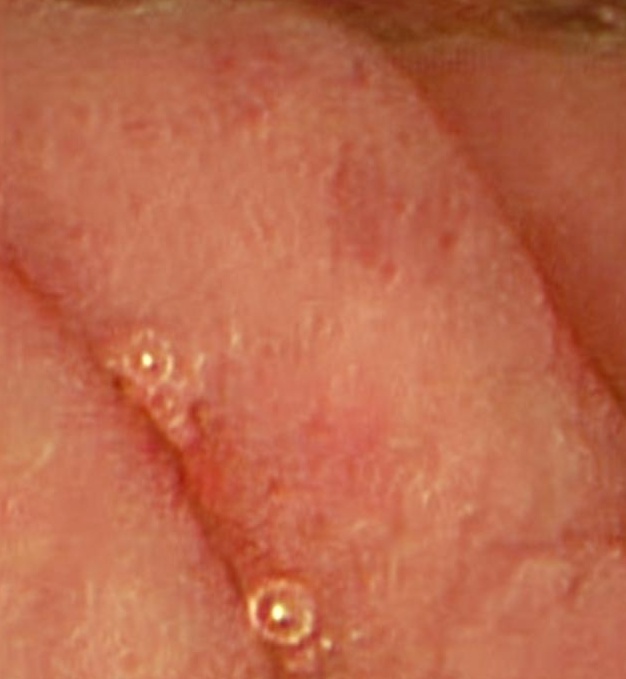}
        }
    \end{minipage}
\vspace{-0.2cm}
    \caption{The Qualitative evaluation of the proposed model against existing unsupervised SR techniques on the newly curated Kvasir dataset for a scaling factor of $\times 4$.}
    \label{fig:kvasir2}
    \vspace{-0.3cm}
\end{figure*}

Fig.~\ref{fig:kvasir2}\footnote{Due to space constraints, additional figures related to the newly edited Kvasitr dataset are included in the \emph{Supplementary material}} provides a visual comparison of SR outputs on the newly curated Kvasir dataset at an upscale factor of $\times4$, highlighting the performance of the proposed method in relation to several state-of-the-art unsupervised SR models. To facilitate clear visual assessment, an enlarged region of interest (highlighted in blue) from the LR input is provided. One can note that the LR input image (Fig.~\ref{fig:kvasir2}(a)) suffers from severe degradation, including blurr and loss of structural fidelity, which hinders the visibility of diagnostically relevant features. The unsupervised methods such as ZSSR and DASR (Fig.~\ref{fig:kvasir2}(b, c)) tend to produce overly smooth outputs, lacking the textural and structural richness necessary for perceptual quality. Similarly, dSRVAE and DUSGAN (see Fig.~\ref{fig:kvasir2}(d, e)) introduce visible distortions and failed to reconstruct fine tissue structures effectively, limiting their diagnostic usability. The SR output of BSRGAN displayed in Fig.~\ref{fig:kvasir2}(f) preserves some structural details; however, it suffers from color inaccuracies and over-smoothing. Further, the SR output obtained using MDASR and TUDASR methods (see Fig.~\ref{fig:kvasir2}(g, h)), although specifically developed for WCE images, tend to oversmooth the textures and miss fine anatomical details, resulting in moderately improved yet suboptimal reconstructions. In contrast, the SR result of the proposed method (i.e., \emph{CEM-TUDASR}) as displayed in Fig.~\ref{fig:kvasir2}(i) demonstrates visually superior performance, accurately recovering high-frequency details such as glandular edges, mucosal ridges, and textural gradients. Thus, the proposed model exhibits improved contrast, structural sharpness, and consistent color representation, enabling clearer visualization of the lesion region. These visual improvements can be attributed to the integration of transformer-based attention mechanisms and domain-adaptive training, which facilitate better localization of diagnostic features and contextual feature learning. The qualitative SR results demonstrate the effectiveness of the proposed approach (i.e., \emph{CEM-TUDASR}) in significantly enhancing the visual quality of WCE images, which is essential for reliable clinical diagnosis. 

\begin{figure*}[t!]
    \centering
    \renewcommand{\arraystretch}{1.2} 
    
    \begin{minipage}{0.20\linewidth} 
        \centering
        \subfloat[\scriptsize LR]{
            \includegraphics[width=\linewidth]{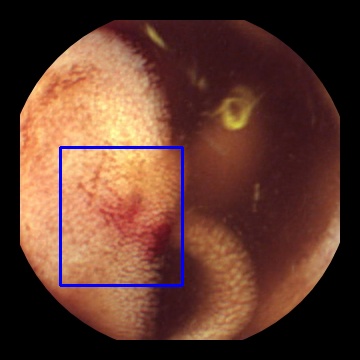}
        }
    \end{minipage}
    \begin{minipage}{0.78\linewidth}
        \centering
        % First row (4 images)
        \subfloat[\scriptsize ZSSR \cite{zssr}]{
            \includegraphics[width=0.24\linewidth]{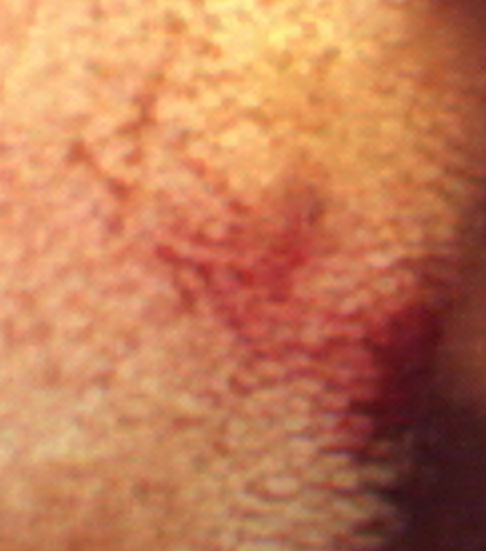}
            }
        \subfloat[\scriptsize DASR \cite{arr41}]{
            \includegraphics[width=0.24\linewidth]{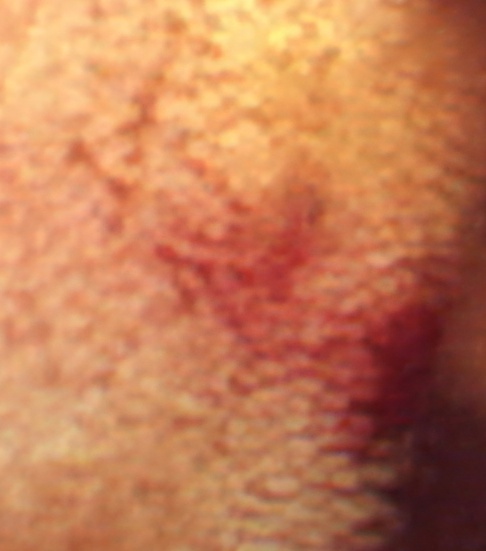}
        }
        \subfloat[\scriptsize dSRVAE \cite{dSRVAE}]{
            \includegraphics[width=0.24\linewidth]{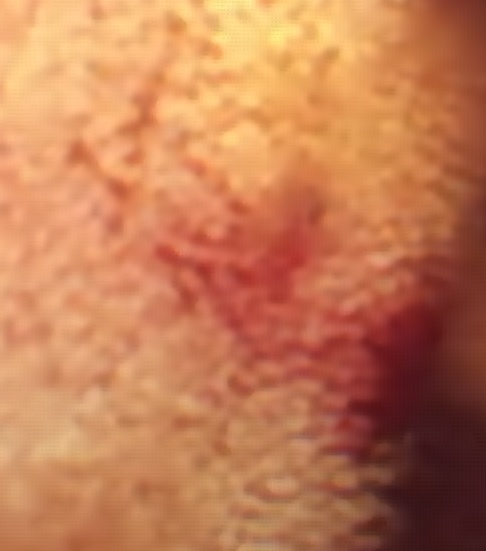}
        }
        \subfloat[\scriptsize DUSGAN \cite{dusgan}]{
            \includegraphics[width=0.24\linewidth]{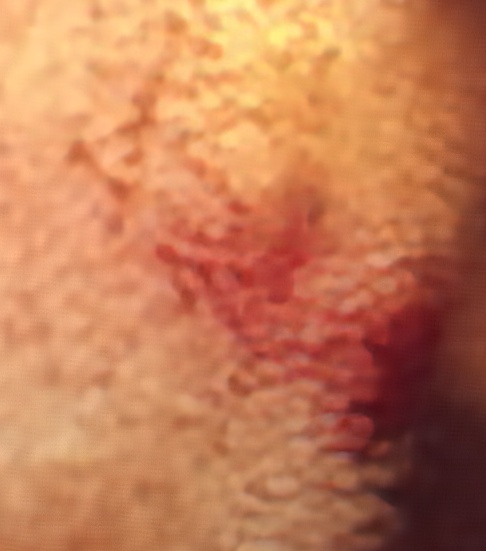}
        }
        \vspace{-0.3cm}\\
        % Second row (4 images)
        \subfloat[\scriptsize BSRGAN \cite{bsrgan}]{
            \includegraphics[width=0.24\linewidth]{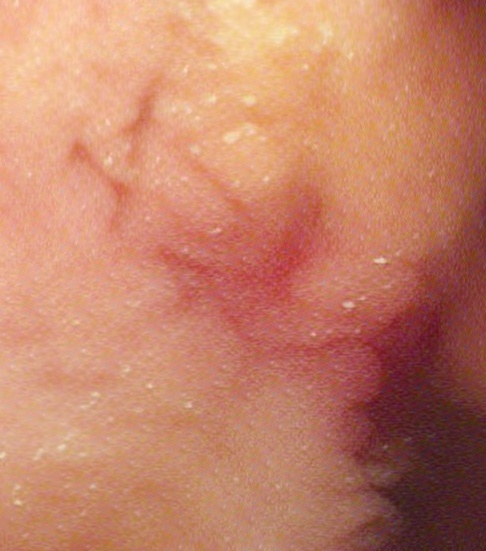}
        }
        \subfloat[\scriptsize MDASR \cite{MDASR}]{
            \includegraphics[width=0.24\linewidth]{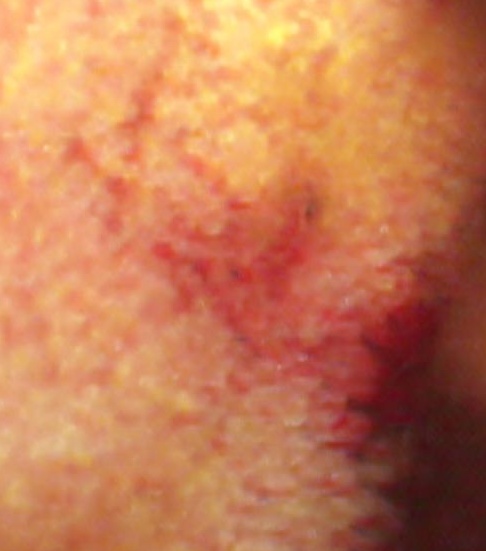}
        }
        \subfloat[\scriptsize TUDASR \cite{tudasr}]{
            \includegraphics[width=0.24\linewidth]{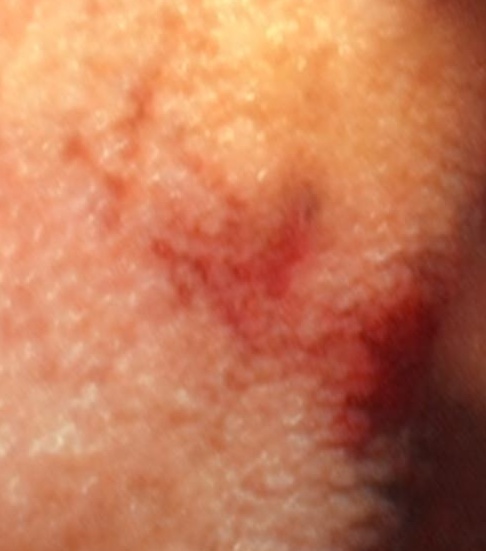}
        }
        \subfloat[\scriptsize Proposed]{
            \includegraphics[width=0.24\linewidth]{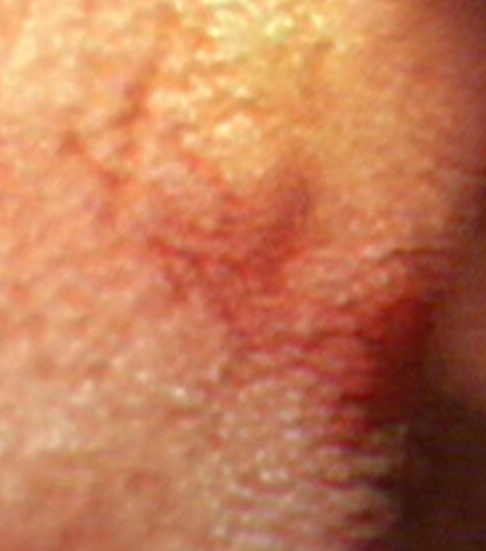}
        }
    \end{minipage}
\vspace{-0.2cm}
    \caption{The Qualitative evaluation of the proposed model against existing unsupervised SR techniques on the KID dataset for a scaling factor of $\times 4$.}
    \label{fig:kid1}
    \vspace{-0.7cm}
\end{figure*}

The qualitative comparison of the proposed model (i.e., \emph{CEM-TUDASR}) against existing state-of-the-art methods on the KID dataset for an upscaling factor of $\times 4$ is illustrated in Fig.~\ref{fig:kid1}\footnote{Due to space constraints, additional figures related to the KID dataset are included in the \emph{Supplementary material}}. The LR input shown in Fig.~\ref{fig:kid1}(a) exhibits substantial degradation, including diminished contrast, blurred lesion boundaries, and a significant loss of high-frequency components such as capillary textures and mucosal structures. The Fig.~\ref{fig:kid1}(b, c)) show the SR outputs of ZSSR and DASR methods that show moderate improvement in lesion boundary sharpness and surface texture enhancement; however, these models introduce noticeable artifacts and suffer from over-smoothing effects that compromise the visibility of vascular features and micro-structural cues. Similarly, dSRVAE (Fig.~\ref{fig:kid1}(d)) produces a relatively smooth output, yet it fails to preserve chromatic consistency and blurs critical diagnostic regions, particularly in areas with reddish lesions. The SR result of DUSGAN approach displayed in Fig.~\ref{fig:kid1}(e) enhances broader lesion structures; however, it lacks the capacity to accurately reconstruct fine vascular networks and subtle tissue irregularities, limiting its clinical interpretability. Further, SR output of BSRGAN method (see Fig.~\ref{fig:kid1}(f)) results in a heavily smoothed image with substantial loss of chromatic information and low contrast across lesion boundaries. Although MDASR and TUDASR (Fig.~\ref{fig:kid1}(g, h)) generate visually cleaner outputs with reduced noise and improved structural integrity, they tend to over-smooth diagnostically critical regions, suppressing vascular granularity and textural features that are essential for early-stage lesion characterization and complex vascular pattern analysis. Comparing to all above results, \emph{CEM-TUDASR} as displayed in Fig.~\ref{fig:kid1}(i) demonstrates marked superiority by effectively reconstructing high-frequency textural components and preserving vascular topology with minimal artifact generation. The integration of degradation-aware learning and attention-guided feature refinement enables the proposed method to retain subtle diagnostic cues, enhance lesion boundary definition, and restore chromatic fidelity—thereby producing super-resolved outputs that are both perceptually natural and clinically informative. 

\begin{figure*}[t!]
    \centering
    \renewcommand{\arraystretch}{1.2} 
    
    \begin{minipage}{0.20\linewidth} 
        \centering
        \subfloat[\scriptsize LR]{
            \includegraphics[width=\linewidth]{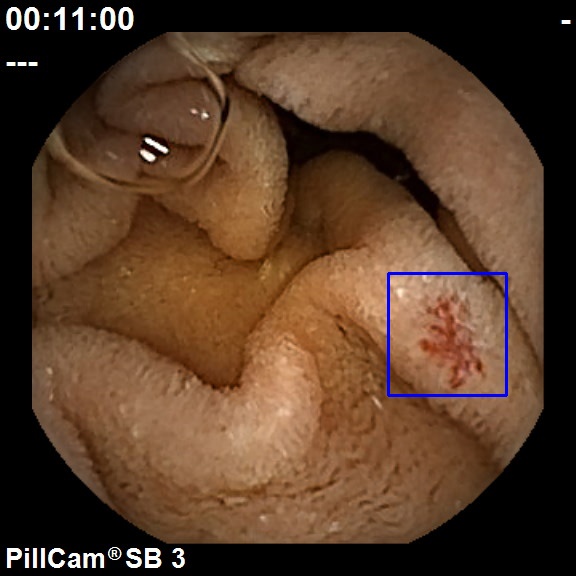}
        }
    \end{minipage}
    \begin{minipage}{0.78\linewidth}
        \centering
        % First row (4 images)
        \subfloat[\scriptsize ZSSR \cite{zssr}]{
            \includegraphics[width=0.24\linewidth]{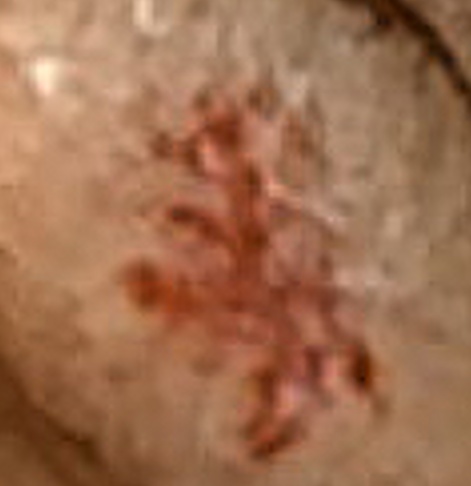}
        }
        \subfloat[\scriptsize DASR \cite{arr41}]{
            \includegraphics[width=0.24\linewidth]{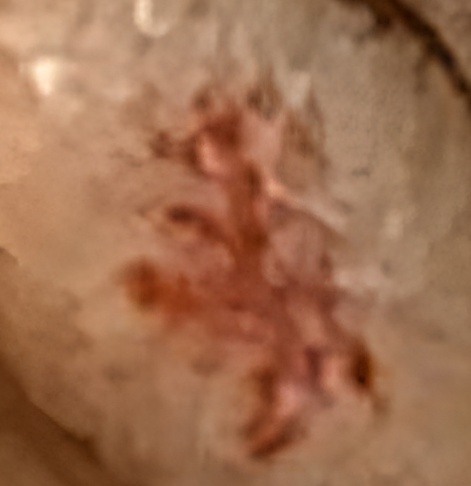}
        }
        \subfloat[\scriptsize dSRVAE \cite{dSRVAE}]{
            \includegraphics[width=0.24\linewidth]{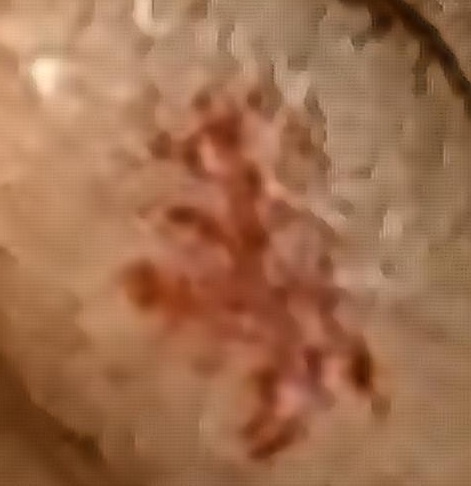}
        }
        \subfloat[\scriptsize DUSGAN \cite{dusgan}]{
            \includegraphics[width=0.24\linewidth]{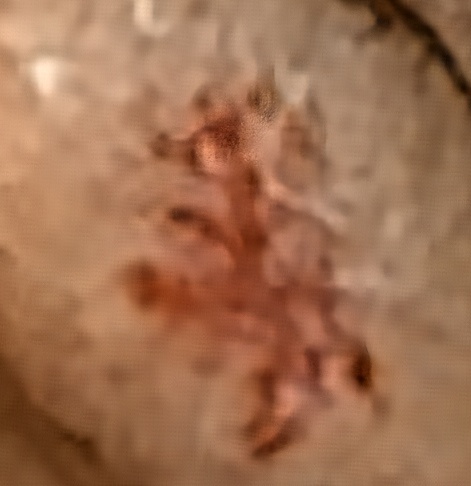}
        }
        \vspace{-0.3cm}
         \\
        
        % Second row (4 images)
        \subfloat[\scriptsize BSRGAN \cite{bsrgan}]{
            \includegraphics[width=0.24\linewidth]{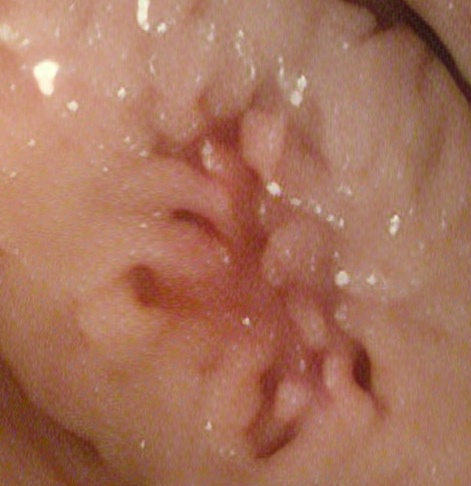}
        }
        \subfloat[\scriptsize MDASR \cite{MDASR}]{
            \includegraphics[width=0.24\linewidth]{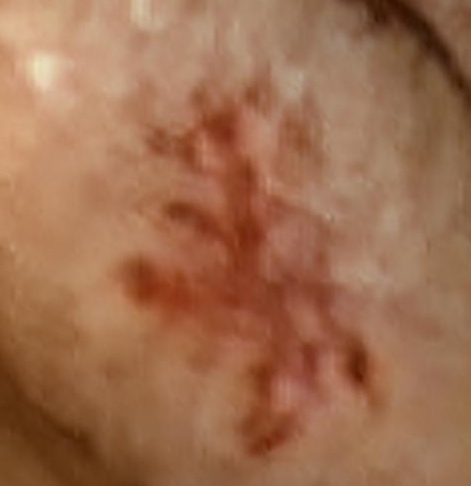}
        }
        \subfloat[\scriptsize TUDASR \cite{tudasr}]{
            \includegraphics[width=0.24\linewidth]{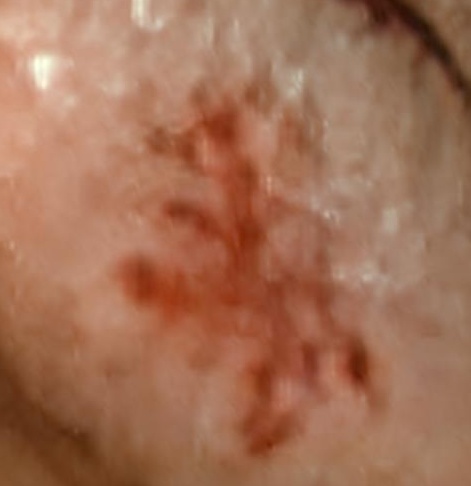}
        }
        \subfloat[\scriptsize Proposed]{
            \includegraphics[width=0.24\linewidth]{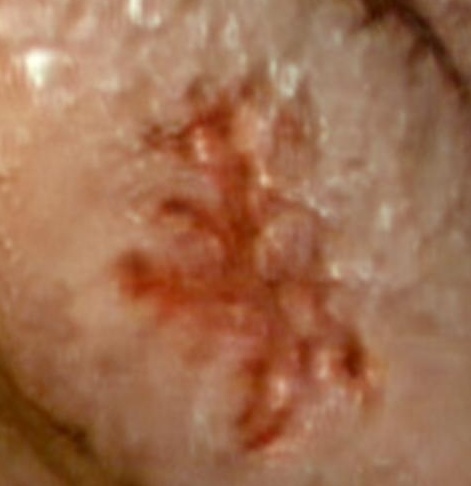}
        }
    \end{minipage}
\vspace{-0.2cm}
    \caption{The Qualitative evaluation of the proposed model against existing unsupervised SR techniques on the GIANA dataset for a scaling factor of $\times 4$.}
    \label{fig:giana3}
    % \vspace{-0.7cm}
\end{figure*}

The visual SR results of the proposed model are compared with other state-of-the-art models on GIANA dataset in Fig.~\ref{fig:giana3}\footnote{Due to space constraints, additional figures on GIANA have been included in the \emph{Supplementary material}}. The LR input is displayed in Fig.~\ref{fig:giana3}(a) that exhibits blurred lesion boundaries and loss of fine anatomical structures, limiting its diagnostic clarity. Among the baseline methods, ZSSR and DASR whose SR images are displayed in Fig.~\ref{fig:giana3}(b,c) produce over-smoothed textures with poor structural delineation, making lesion localization difficult. Additionally, Fig.~\ref{fig:giana3}(d,e) depict the SR results of dSRVAE and DUSGAN methods that offer moderate improvements in structural sharpness; however, fail to recover intricate mucosal textures and introduce mild artifacts. Moreover, the BSRGAN SR method (see Fig.~\ref{fig:giana3}(f)) exhibits sharper reconstruction than other baselines but introduces unnatural color tones and noise amplification. In the similar line, one can note by looking at the SR results obtained using MDASR and TUDASR methods (see Fig.~\ref{fig:giana3}(g,h)) that they yield relatively cleaner outputs tailored for endoscopic domains; however, they still lack precision in reconstructing fine vascular and glandular structures, resulting in limited enhancement in clinical interpretability. Conversely, the SR obtained using the proposed method i.e., \emph{CEM-TUDASR} is depicted in Fig.~\ref{fig:giana3}(i) which reconstructs clear superior visual result by faithfully restoring high-frequency details such as mucosal folds, lesion textures, and vascular patterns. Additionally, it demonstrates improved sharpness, structural integrity, and color fidelity, providing enhanced visibility of the diagnostically critical region. The qualitative analysis across the Kvasir, KID, and GIANA datasets is conducted in accordance with the objectives of the proposed framework, particularly emphasizing the preservation of fine mucosal textures, vascular structures, perceptual fidelity, and structural consistency under diverse imaging conditions.

\subsection{Quantitative Analysis}
\begin{table*}[!t]
\caption{The quantitative evaluation of the proposed model:\emph{CEM-TUDASR} against existing unsupervised SR techniques on the Kvasir, KID and GIANA dataset for a scaling factor of $\times 4$. Here, first and second highest values are highlighted with
red and blue colors, respectively.}
% \vspace{-0.2cm}
\resizebox{\textwidth}{!}{%
\begin{tabular}{lcccc|llll|llll}
\hline \hline
\multicolumn{1}{c}{\multirow{2}{*}{Method ($\times 4)$}} & \multicolumn{4}{c}{Newly edited Kvasir Dataset}                    & \multicolumn{4}{c}{KID dataset } & \multicolumn{4}{c}{GIANA Dataset } \\ \cline{2-13} 
\multicolumn{1}{c}{}                        & BRISQUE$\downarrow$ & \multicolumn{1}{l}{PIQE$\downarrow$} & NIQE $\downarrow$   & EndoQM $\downarrow$     & BRISQUE$\downarrow$   & PIQE$\downarrow$     & NIQE$\downarrow$    & EndoQM $\downarrow$     & BRISQUE$\downarrow$   & PIQE$\downarrow$      & NIQE $\downarrow$    & EndoQM $\downarrow$     \\ \hline \hline
% SRResCGAN \cite{SRResCGAN}                                  & 66.1341 & 58.1722                  & \textcolor{red}{4.4847} & 10.3484 & 88.7592   & 75.7123  & 5.1225  & 10.7286  & 86.5817   & 82.1627   & 4.9225   & 9.7258   \\ 
ZSSR \cite{zssr}                                       & 64.7164 & 85.2007                  & 6.0950 & 13.1060 & 83.7318   & 81.1143  & 4.9983  & 11.3095  & 86.9541   & 81.2241   & 5.0458   & 8.4063   \\ 
DASR\cite{arr41}                                       & 63.7744 & 91.8829                  & 6.0622 & 14.2492 & 71.3107   & \textcolor{red}{42.6297}  & 5.4863  & 13.5945  & 63.1576   & 53.8924   & \textcolor{blue}{4.2092}   & 10.0339  \\ 
dSRVAE \cite{dSRVAE}                                      & 56.4932 & 64.9396                  & \textcolor{red}{4.8916} & 10.8352 & \textcolor{red}{56.6771}  & 47.8164  & 5.9949  & 13.5611  & \textcolor{red}{46.6371}   & \textcolor{red}{43.2417}   & 5.6909   & 10.0163  \\ 
DUSGAN \cite{dusgan}                                     & 72.1658 & 87.5359                  & 7.3791 & 13.8334 & 59.9603   & 54.9143  & 6.6590  & 17.3653  & \textcolor{blue}{58.0079}   & \textcolor{blue}{51.3542}   & 5.9432   & 13.4425  \\ 

BSRGAN \cite{dusgan}                                     & 60.9487 & 70.3330                  & 5.7289 & 9.6295 & 75.4041   & 60.5074  & 5.6630  & 9.0223  & 65.8627   & 70.1311   & 5.6629   & 11.2288  \\ 

MDASR \cite{MDASR}                                       & 59.3752 & 62.8371                 & 5.4123 & 8.9283  & 76.7706   & \textcolor{blue}{47.4550}  & \textcolor{blue}{4.8297}  & 7.6567   & 71.0476   & 68.4324   & 4.8809   & 7.7718  \\

TUDASR \cite{tudasr}                                     & \textcolor{red}{55.2830} & \textcolor{blue}{59.4121}                  & 5.2104 & \textcolor{blue}{7.6216} & 60.2078   & 53.4644  & 4.8630  & \textcolor{blue}{6.5706}  & 71.9798   & 69.1666   & 4.4672   & \textcolor{blue}{6.8840}  \\

Proposed                                    & \textcolor{blue}{56.2884} & \textcolor{red}{53.1951}                  & \textcolor{blue}{5.1846} &\textcolor{red}{6.1204 }  & \textcolor{blue}{57.3572}   & 50.3821  & \textcolor{red}{4.5140}  & \textcolor{red}{5.8679}   & 68.0625   & 72.3912   & \textcolor{red}{4.1987}   & \textcolor{red}{5.9021}   \\ \hline
\hline
\end{tabular}%
}
\label{tab:quantitative}
\vspace{-0.5cm}
\end{table*}

To further validate the effectiveness of the proposed \emph{CEM-TUDASR} model, we conduct an extensive quantitative analysis by comparing its performance with several state-of-the-art unsupervised SR methods. The evaluation is carried out using widely adopted no-reference image quality assessment metrics, namely BRISQUE, PIQE, and NIQE, along with EndoQM, a domain-specific perceptual quality metric tailored for endoscopic image evaluation. Lower values across all four metrics indicate better perceptual quality and fewer visual artifacts. These metrics provide insights into the perceptual quality of the super-resolved WCE images without relying on ground-truth HR images. In Table~\ref{tab:quantitative}, the top two values are highlighted with red and blue colored texts, respectively, for the convenience of the reader. It is evident from this table that, the proposed method achieves a second highest BRISQUE score among all compared approaches, slightly behind TUDASR. This indicates that the proposed model effectively suppresses spatial distortions and enhances perceptual quality in terms of natural image statistics. Notably, the PIQE score obtained by the proposed method is the lowest among all methods evaluated, demonstrating superior robustness in preserving perceptual texture structure and minimizing pixel-level degradations across the WCE image surface. The NIQE score is also the best among all methods, suggesting that the SR outputs generated by the proposed model closely match the statistical distribution of high-quality natural images, which is critical for clinical interpretability. Furthermore, the proposed method yields an EndoQM score which is the lowest across all models. Since EndoQM is specifically trained to reflect perceptual quality tailored to endoscopic image characteristics (i.e., vascular patterns, mucosal textures, and lighting conditions), this result strongly validates the model’s capacity to reconstruct diagnostically important features while avoiding artificial enhancement or detail hallucination. By contrast, other models such as ZSSR, DASR, and DUSGAN exhibit significantly higher EndoQM scores, indicating their limited effectiveness in domain-specific quality restoration. Overall, \emph{CEM-TUDASR} not only outperforms all baseline approaches in terms of PIQE, NIQE, and EndoQM, but also achieves a highly competitive BRISQUE score. These findings confirm the model’s effectiveness in reconstructing perceptually faithful and diagnostically reliable high-resolution WCE images, despite the challenges posed by unpaired training and the absence of ground-truth HR images in clinical scenarios.

The quantitative evaluation of the proposed method (\emph{CEM-TUDASR}) on the KID dataset (Table~\ref{tab:quantitative}) demonstrates substantial improvements over state-of-the-art methods. Traditional approaches, such as ZSSR, BSRGAN, and MDASR, yield higher BRISQUE and PIQE scores due to their limited ability to reconstruct fine anatomical details, introducing noticeable perceptual distortions. While TUDASR performs better owing to its endoscopy-specific architecture, it still struggles to accurately restore fine textures. In contrast, the proposed method achieves significantly lower BRISQUE, PIQE, and NIQE scores, indicating superior perceptual quality and structural coherence. Most notably, the proposed model attains the lowest EndoQM scores, highlighting its ability to preserve diagnostically crucial textures and vascular patterns. This enhanced performance stems from its Transformer-based architecture and domain adaptation capability, effectively capturing long-range dependencies essential for clinical endoscopic imagery.

The quantitative evaluation on the GIANA dataset (Table~\ref{tab:quantitative}) confirms the superior generalization of the proposed method. Across all evaluated no-reference metrics, the proposed approach consistently outperforms existing unsupervised SR techniques, achieving the lowest NIQE and EndoQM scores. This indicates its enhanced ability to reconstruct clinically relevant textures and subtle anatomical features, compared to methods such as ZSSR, BSRGAN, and DASR, which introduce more perceptual artifacts. Although MDASR and TUDASR, specialized for endoscopy images, exhibit competitive results, they still underperform in preserving intricate details. Leveraging a degradation-aware Transformer and attention-driven refinement, the proposed method effectively adapts to new domains, delivering high-quality outputs suitable for practical WCE clinical scenarios.

\subsection{Cross Domain Adaptability Analysis}
\begin{figure*}[t!]
    \centering
    \renewcommand{\arraystretch}{1.2} 
    
    \begin{minipage}{0.20\linewidth} 
        \centering
        \subfloat[\scriptsize LR]{
            \includegraphics[width=\linewidth]{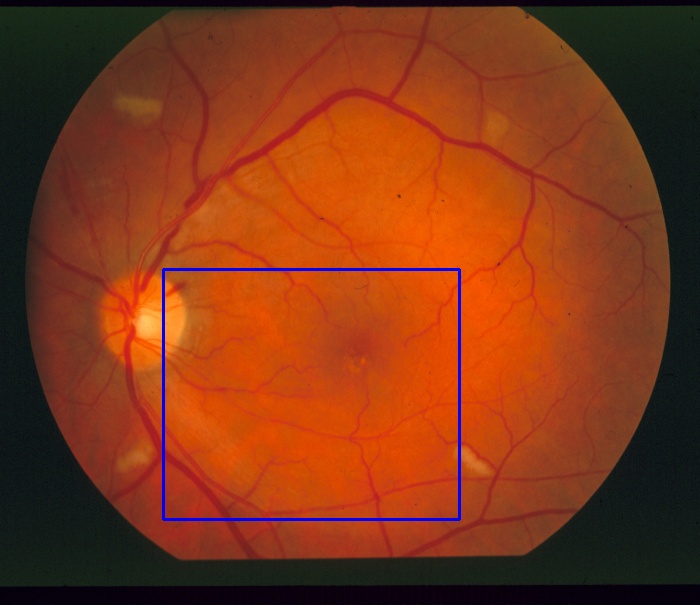}
        }
    \end{minipage}
    \begin{minipage}{0.78\linewidth}
        \centering
        % First row (4 images)
        \subfloat[\scriptsize ZSSR \cite{zssr}]{
            \includegraphics[width=0.24\linewidth]{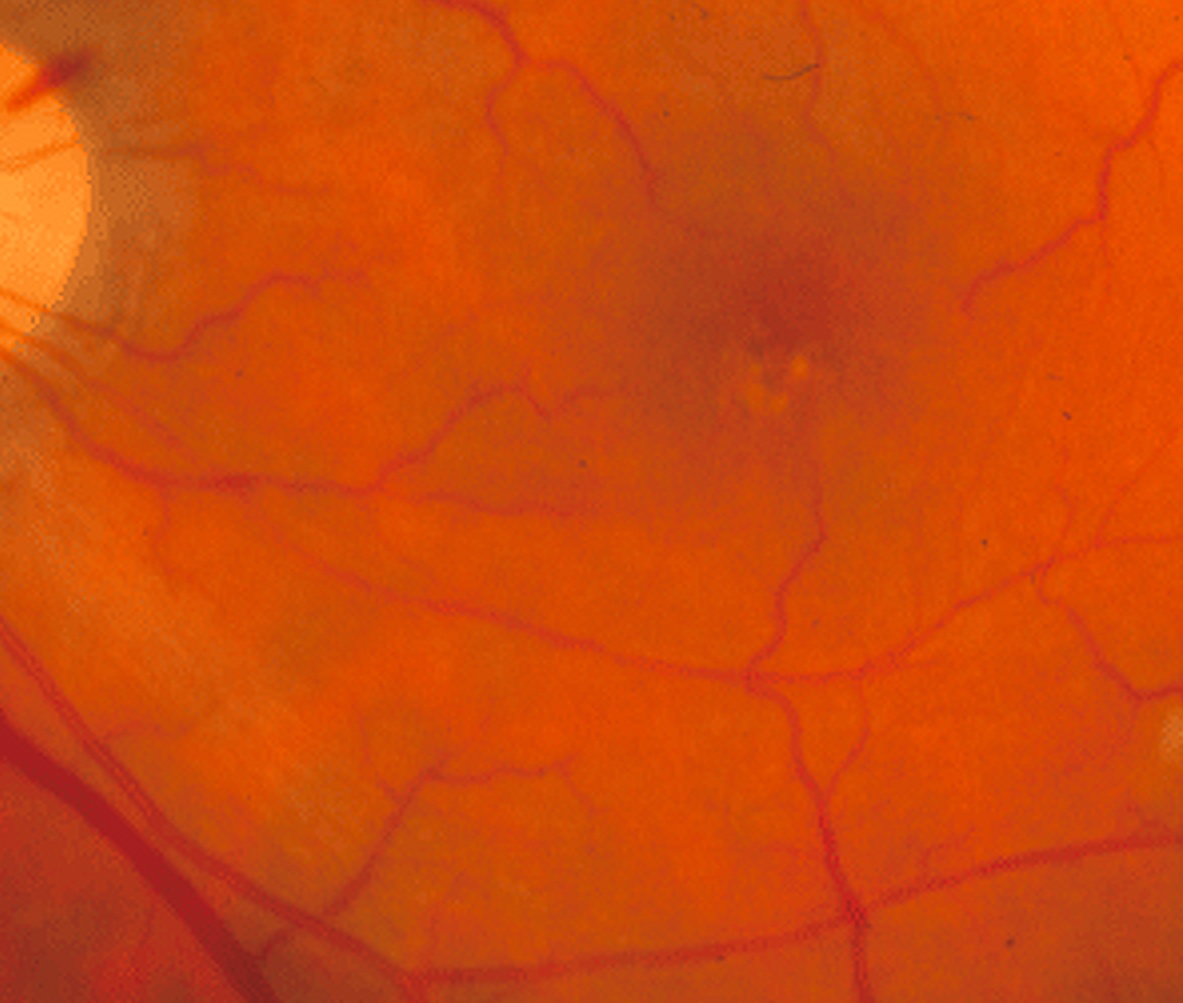}
        }
        \subfloat[\scriptsize DASR \cite{arr41}]{
            \includegraphics[width=0.24\linewidth]{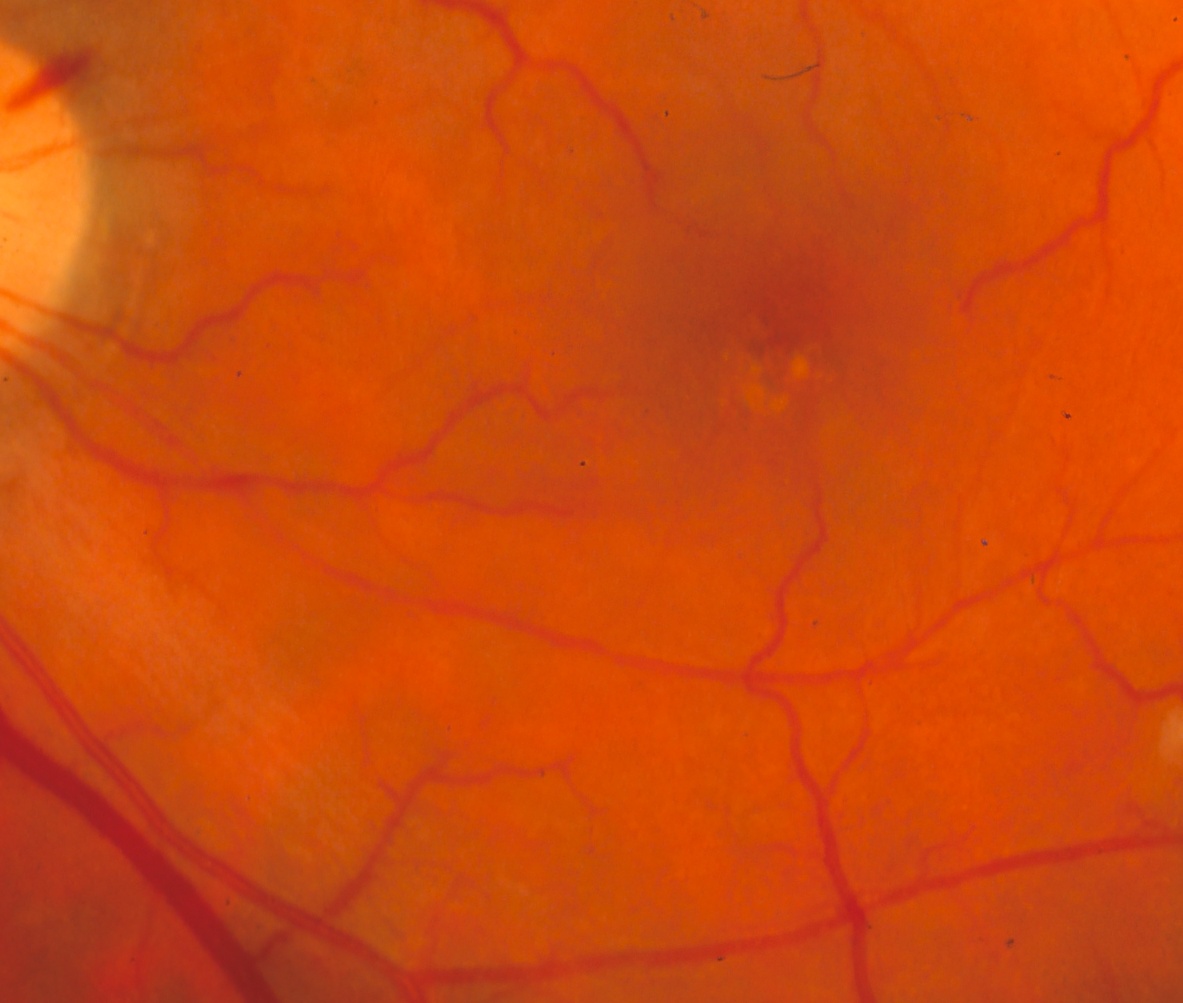}
        }
        \subfloat[\scriptsize dSRVAE \cite{dSRVAE}]{
            \includegraphics[width=0.24\linewidth]{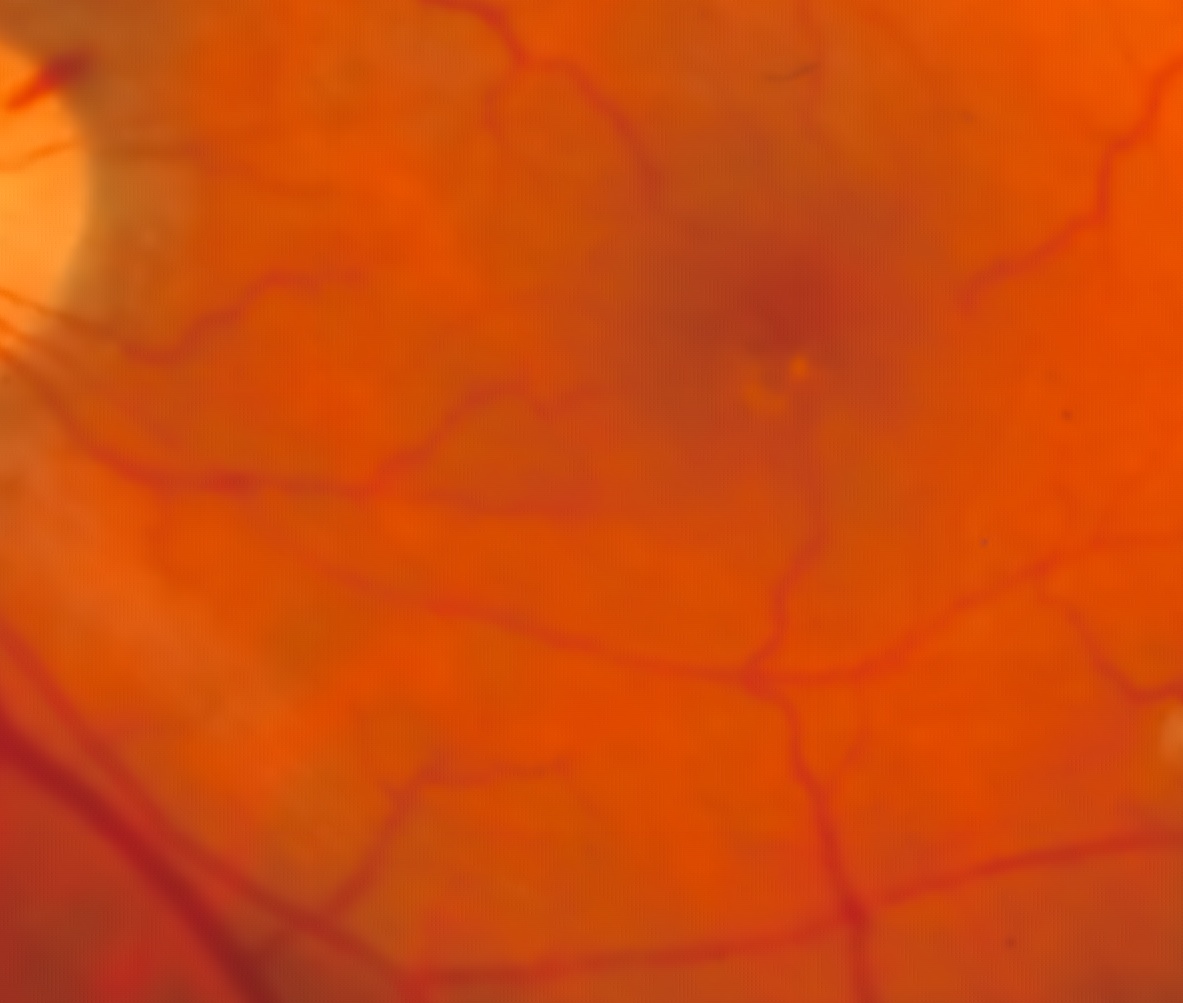}
        }
        \subfloat[\scriptsize DUSGAN \cite{dusgan}]{
            \includegraphics[width=0.24\linewidth]{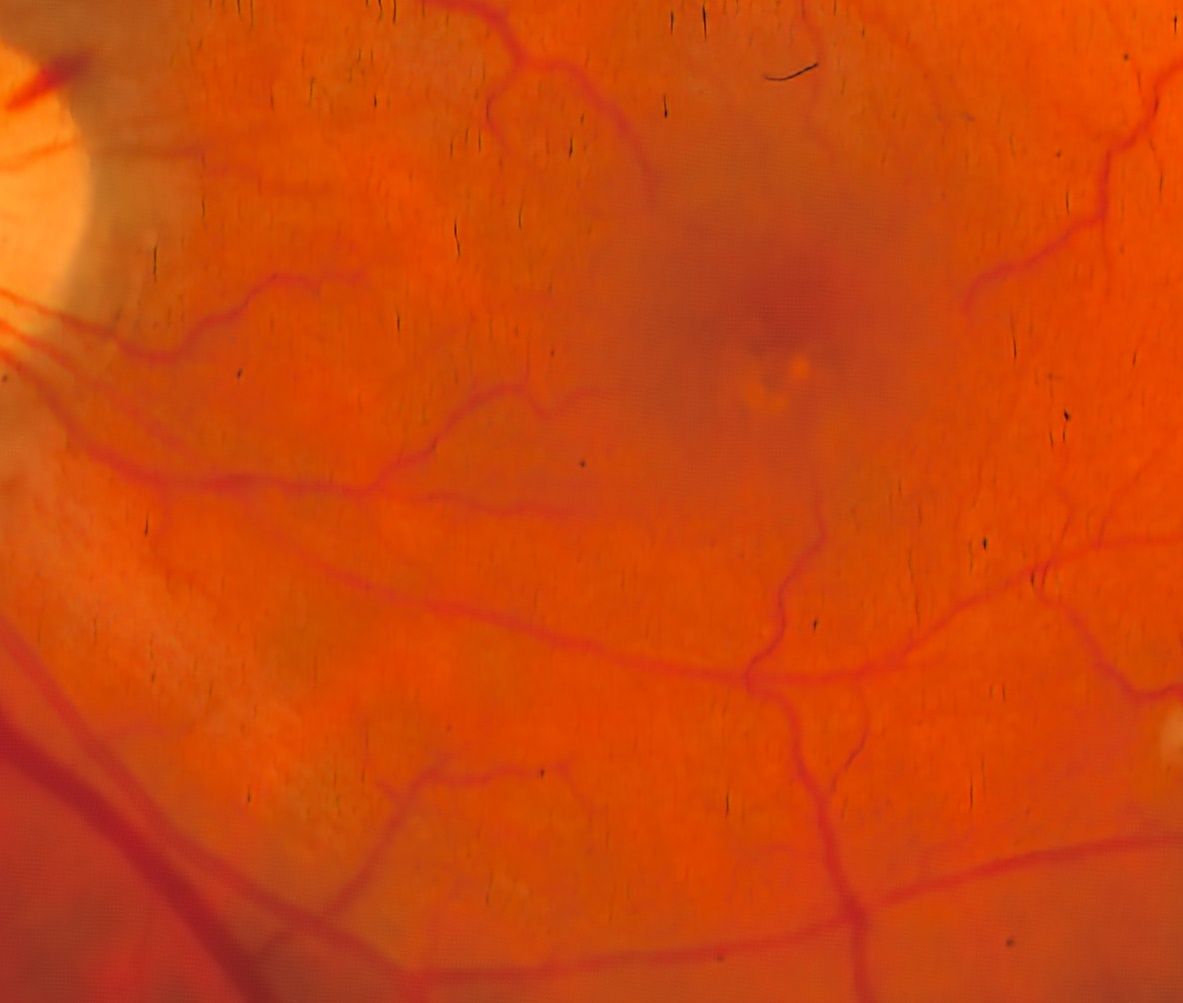}
        }
         \\ \vspace{-0.3cm}
        
        % Second row (4 images)
        \subfloat[\scriptsize BSRGAN \cite{bsrgan}]{
            \includegraphics[width=0.24\linewidth]{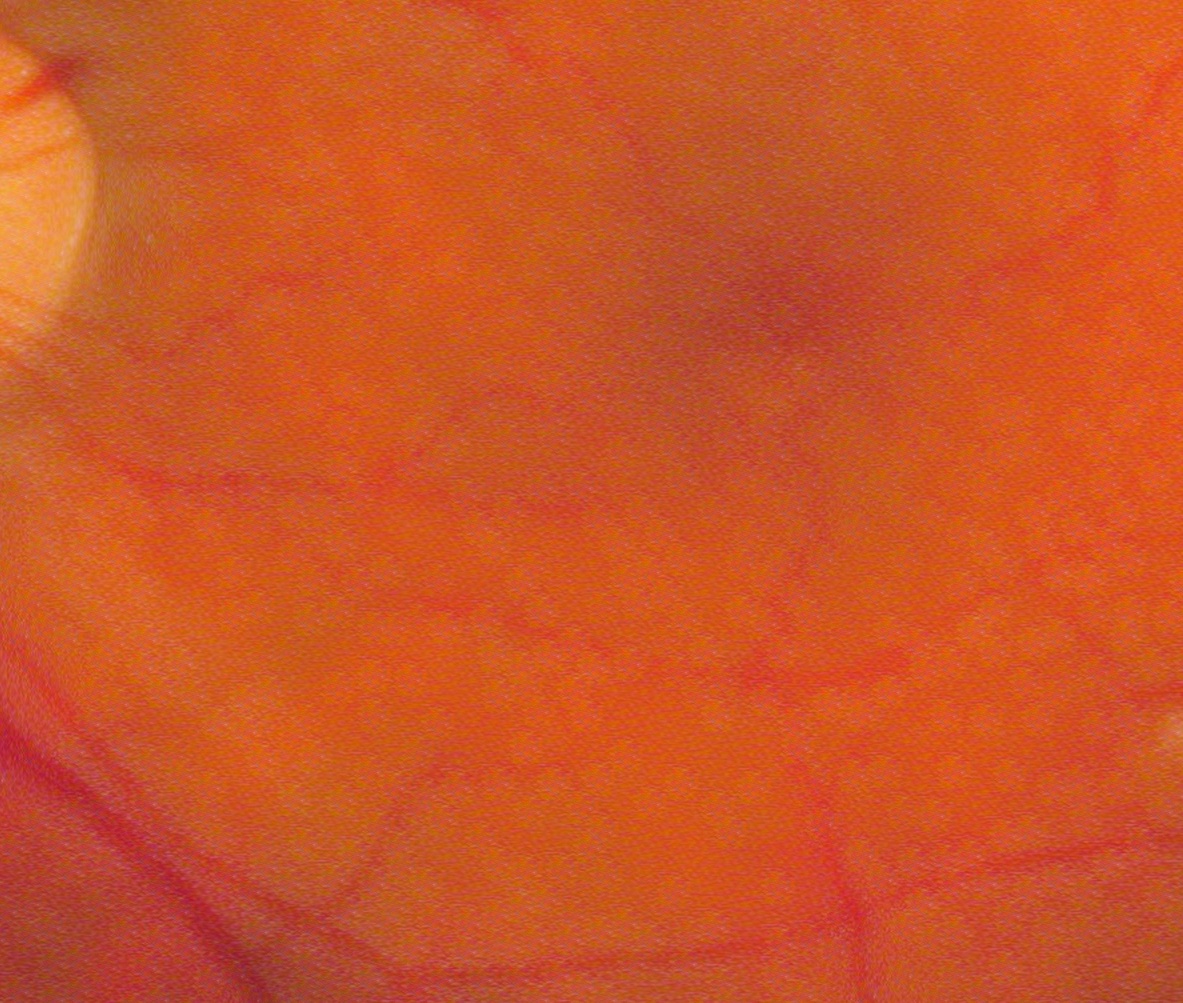}
        }
        \subfloat[\scriptsize MDASR \cite{MDASR}]{
            \includegraphics[width=0.24\linewidth]{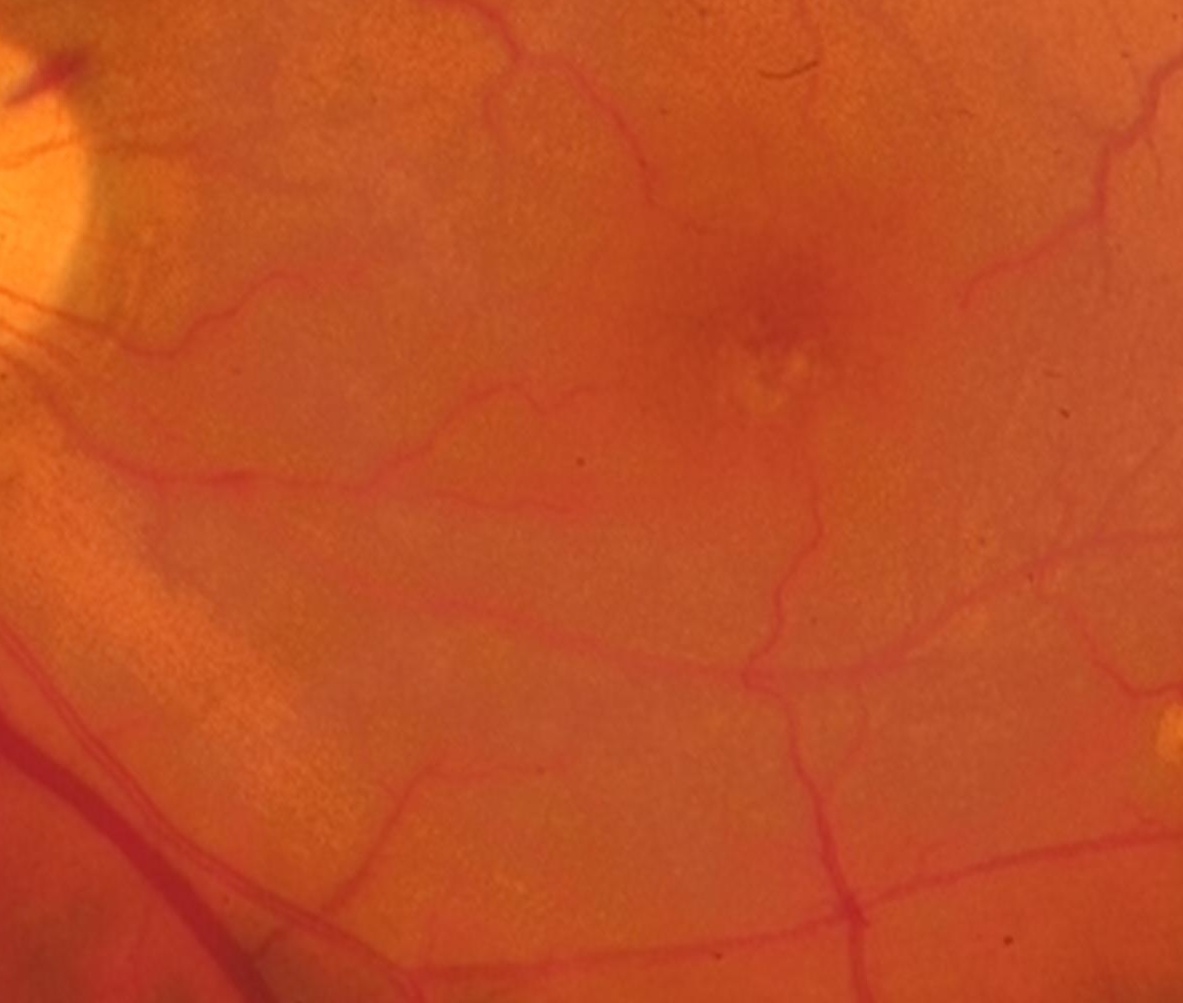}
        }
        \subfloat[\scriptsize TUDASR \cite{tudasr}]{
            \includegraphics[width=0.24\linewidth]{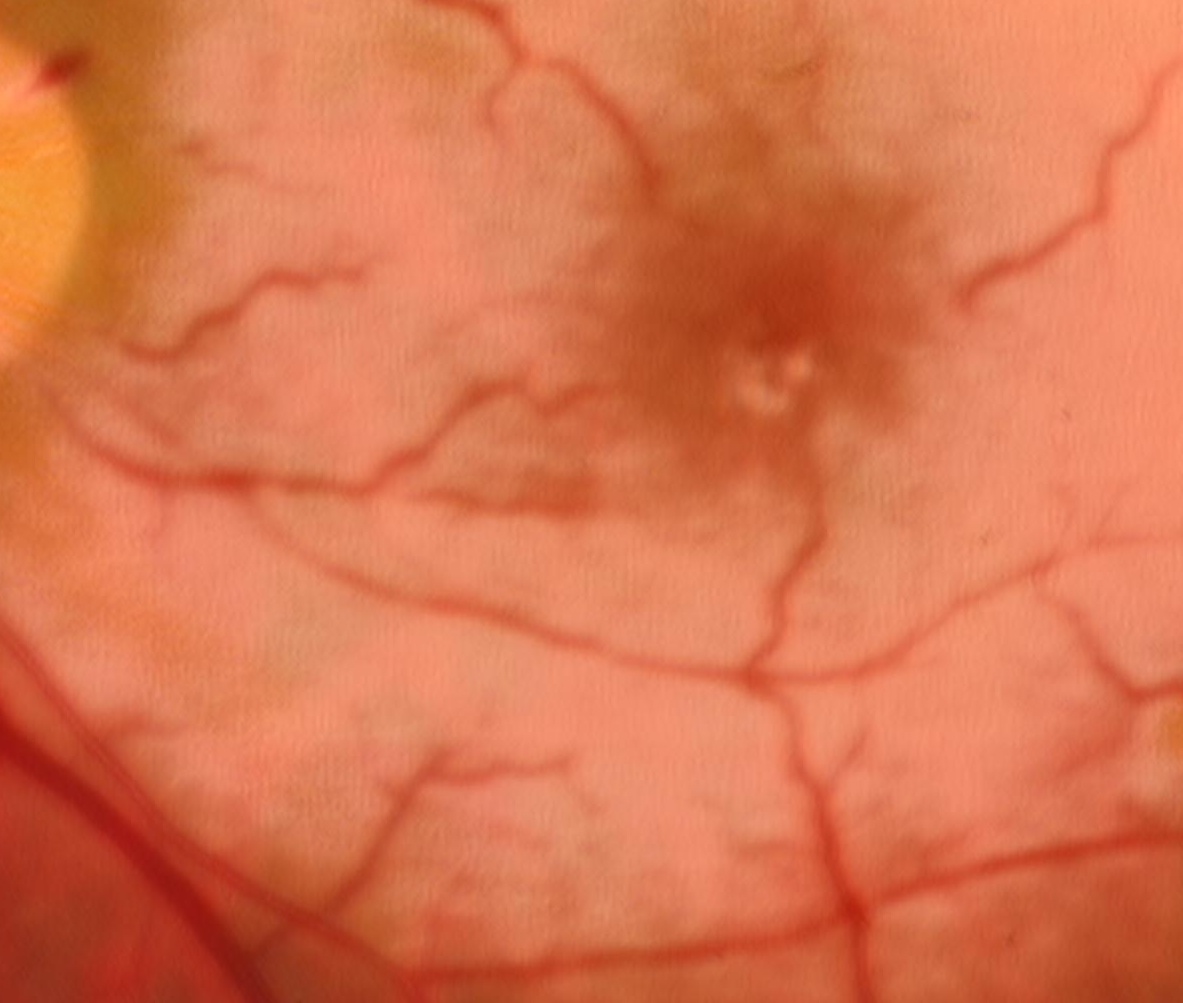}
        }
        \subfloat[\scriptsize Proposed]{
            \includegraphics[width=0.24\linewidth]{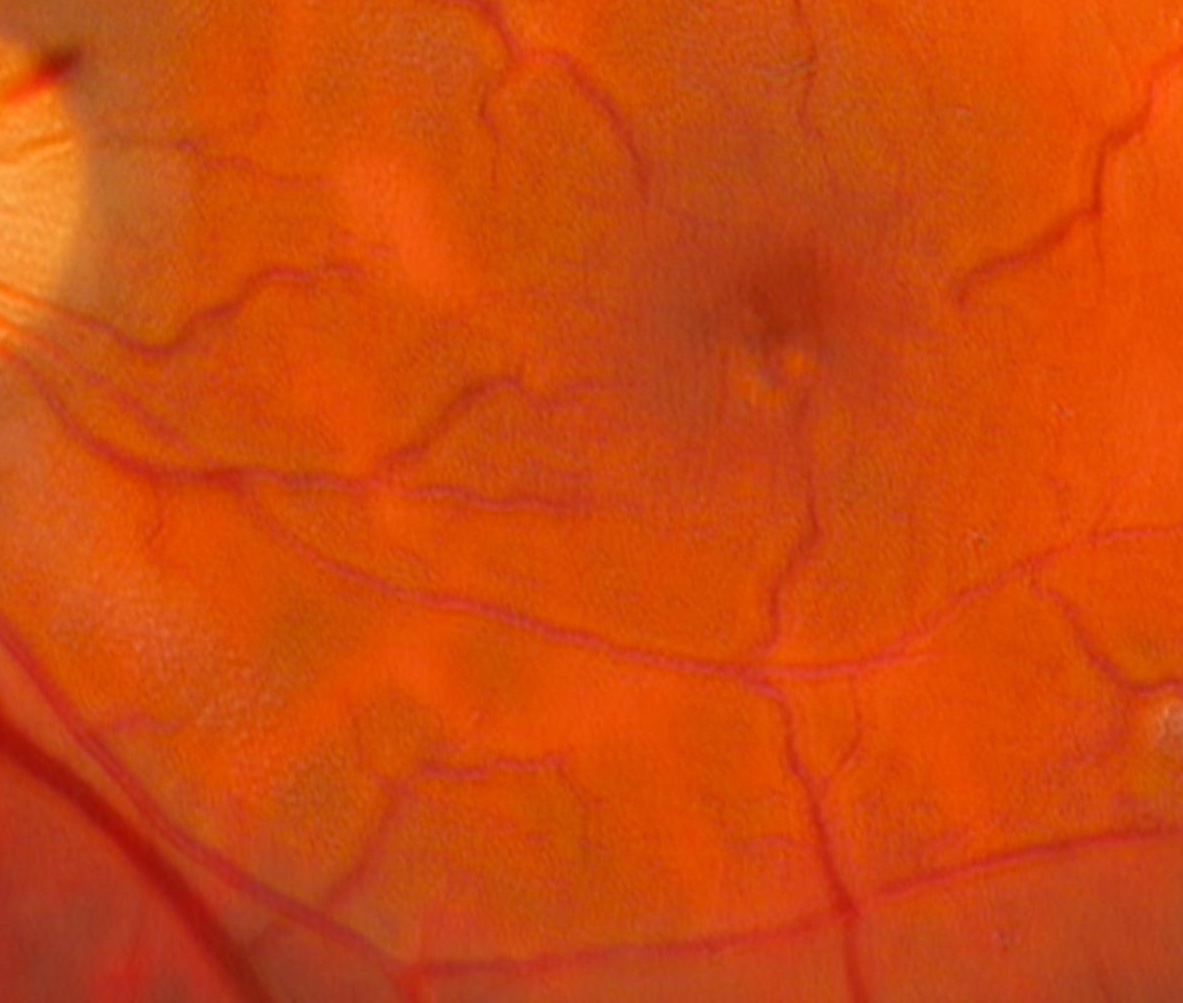}
        }
    \end{minipage}
\vspace{-0.2cm}
    \caption{Qualitative evaluation of the proposed model against existing unsupervised SR techniques on the Retinal dataset for a scaling factor of $\times 4$.}
    \label{fig:retina4}
    %\vspace{-0.7cm}
\end{figure*}

% \begin{table}[]
% \centering
% \caption{The Quantitative evaluation of the proposed model against existing unsupervised SR techniques on the Retinal dataset for a scaling factor of $\times 4$. Here, first and second highest values are highlighted with red and blue colors, respectively.}
% \label{tab:retinal}
% \begin{tabular}{|c|c|c|c|}
% \hline
% \textbf{Method ($\times 4$)} & \textbf{BRISQUE $\downarrow$} & \textbf{PIQE $\downarrow$} & \textbf{NIQE $\downarrow$} \\ \hline
% ZSSR \cite{zssr}                 & 58.0477          & 65.7772       & 5.4189        \\ \hline
% DASR \cite{arr41}                 & \textcolor{blue}{54.9495}          & 59.4072       & \textcolor{red}{5.0094}        \\ \hline
% dSRVAE \cite{dSRVAE}              & 62.6591          & \textcolor{blue}{42.9949}       & 5.7565        \\ \hline
% DUSGAN \cite{dusgan}               & 55.7971          & 44.5746       & 5.7259        \\ \hline
% BSRGAN \cite{bsrgan}               & 57.0901           & 51.9623       & 5.1887        \\ \hline
% MDASR \cite{MDASR}                & 57.0023          & 49.9143       & 5.4215        \\ \hline
% TUDASR \cite{tudasr}               & 56.3710          & 49.9854       & 5.7847        \\ \hline
% Proposed             & \textcolor{red}{54.3320}          & \textcolor{red}{42.0714}       & \textcolor{blue}{5.1449}        \\ \hline
% \end{tabular}
% \end{table}

To further assess the robustness and generalization capability of the proposed unsupervised SR framework (i.e., \emph{CEM-TUDASR}), we conduct additional cross-domain experiments on retinal images \cite{retinal}. This evaluation aims to investigate the model’s ability to transfer knowledge beyond Gastrointestinal (GI) endoscopy and operate effectively on medical images that differ significantly in anatomical structure, texture distribution, and acquisition conditions. Retinal images were specifically chosen as the target domain for cross-modal testing instead of volumetric modalities such as MRI or CT for several practical and methodological reasons. First, retinal fundus images are two-dimensional (2D) in nature, similar to WCE and conventional endoscopy images. This structural compatibility ensures a fair and technically feasible evaluation without requiring significant architectural or dimensional adjustments to the model, which is optimized for 2D spatial representations. In contrast, MRI and CT data are inherently volumetric (3D) and often require specialized pre-processing, architectural extensions (i.e., 3D convolutions), and domain-specific training strategies, which fall outside the scope of the current 2D SR framework. Second, retinal images share certain diagnostic characteristics with endoscopic images, such as the importance of high-frequency details (i.e., vascular structures, lesion boundaries) and color-based tissue differentiation. This makes them a challenging yet relevant testbed for assessing the model’s ability to preserve clinically meaningful features under a domain shift. Notably, the proposed model, trained without access to retinal data, demonstrates strong qualitative performance in enhancing vascular clarity and edge sharpness in retinal images, accurately recovering fine structures like microaneurysms and vessel branches critical for early diagnosis of ocular diseases (See Fig.~\ref{fig:retina4}). The additional qualitative and quantitative analysis of the proposed method on retinal images is discussed in Section~V-D in \emph{Supplementary material}. The cross-domain qualitative analysis validates the objective of the proposed framework in achieving robust generalization capability and perceptually consistent reconstruction across different medical imaging modalities beyond WCE images.

\subsection{Statistical Analysis}

\begin{figure*}[!t]
    \centering
    \subfloat[BRISQUE]{\includegraphics[width=9 cm, height=7cm]{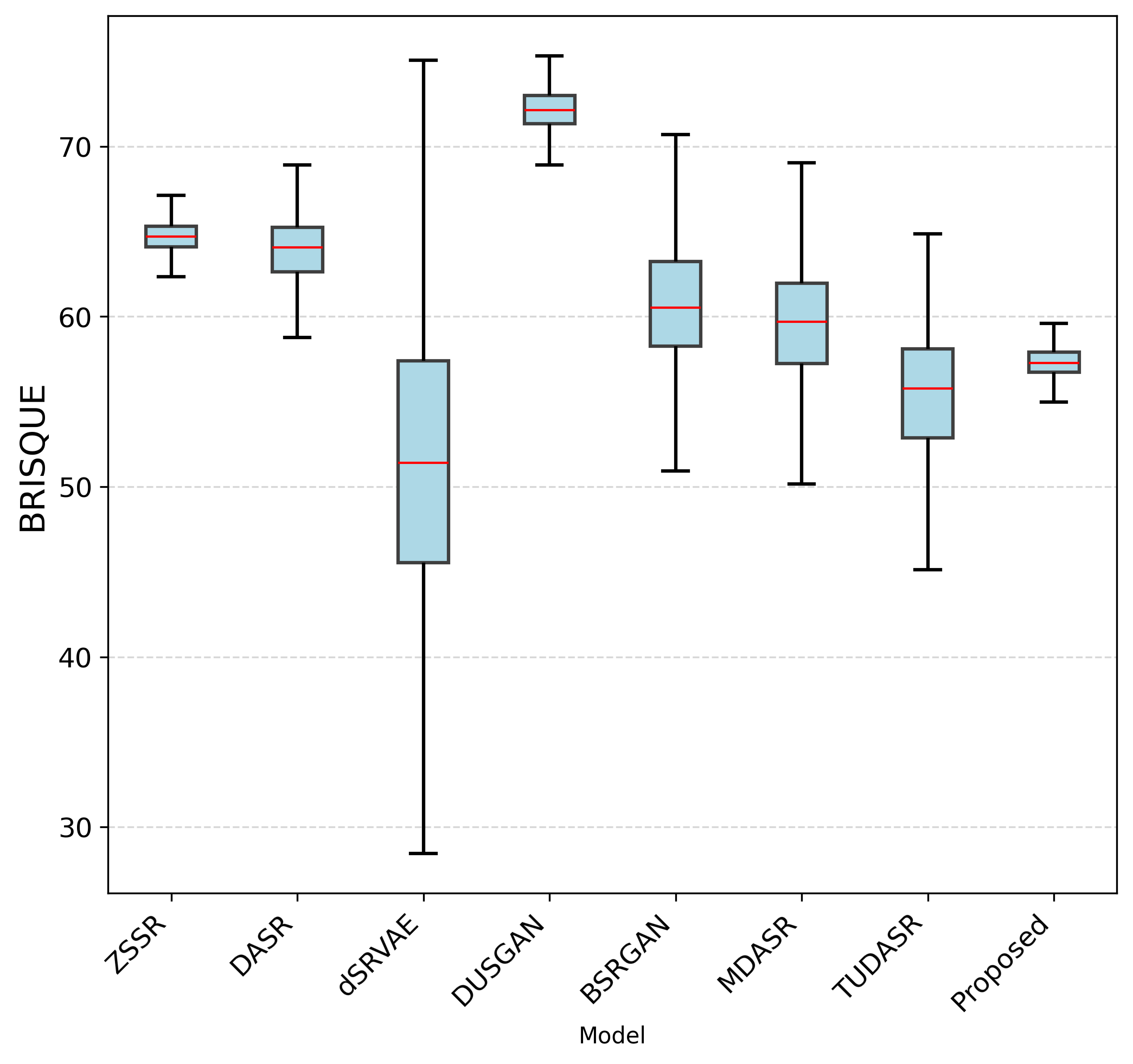}}
    \subfloat[PIQE]{\includegraphics[width=9cm, height=7cm]{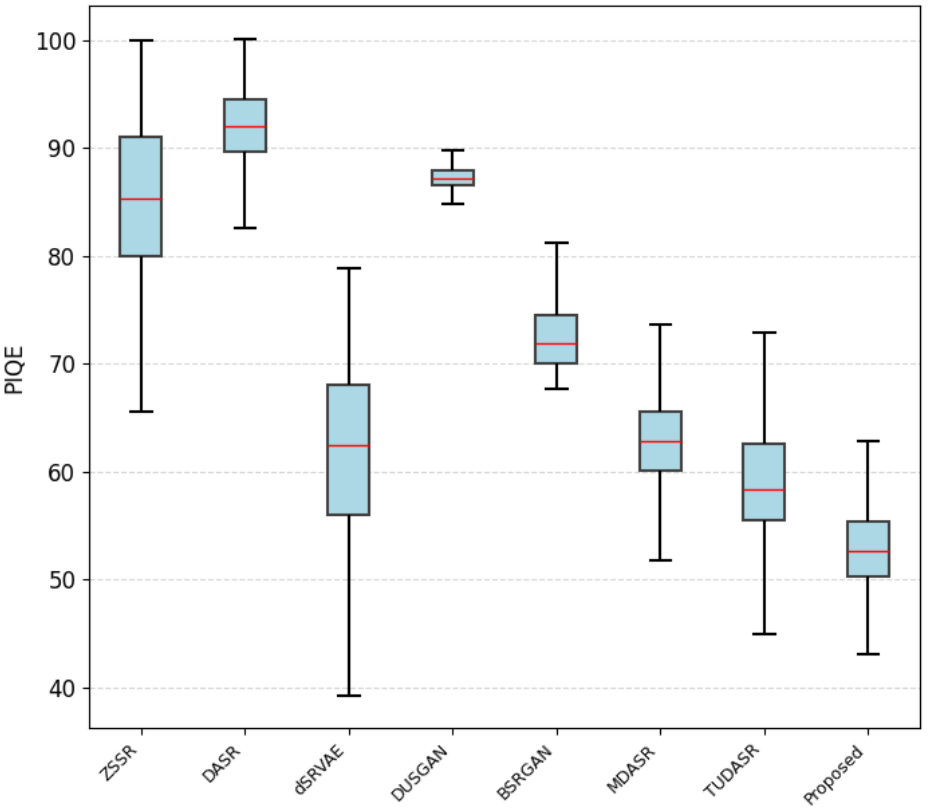}}\\
    \subfloat[NIQE]{\includegraphics[width=9 cm, height=7cm]{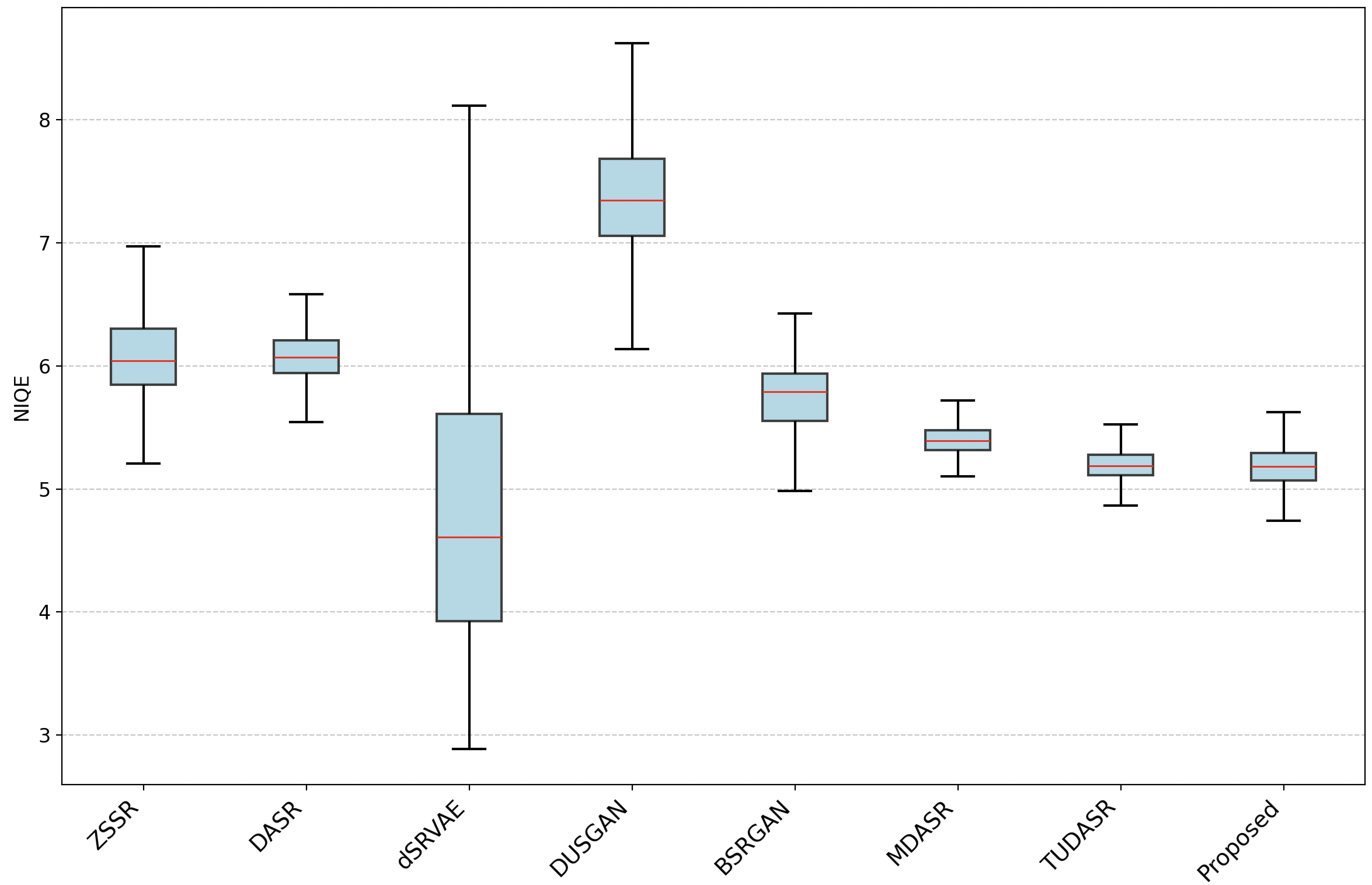}}
    \subfloat[EndoQM]{{\includegraphics[width=9 cm, height=7cm]{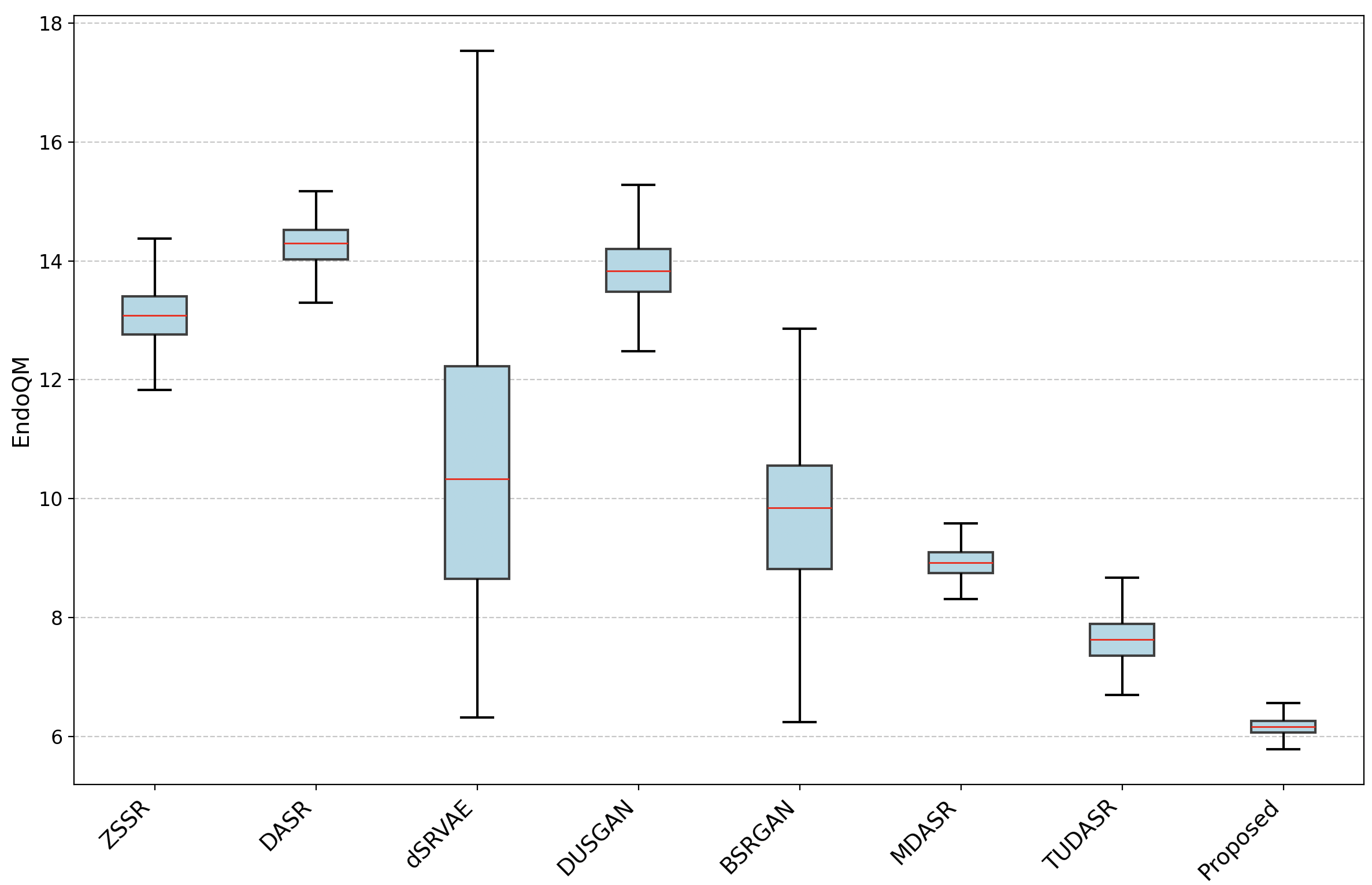}}}
   \caption{The box plot representation of BRISQUE, PIQE, 
NIQE and EndoQM  measurements obtained using the proposed
and other methods on newly edited Kvasir Capsule test dataset.
}
    \label{fig:anova}
      
\end{figure*}

\begin{table*}[]
\caption{The quantitative results in terms of Confidence Interval (CI) on BRISQUE, PIQE, NIQE and EndoQM measures of different unsupervised network. Here, first and second highest values are highlighted with \textcolor{red}{red} and \textcolor{blue}{blue} colors, respectively.}
\label{tab:anova}
\resizebox{\textwidth}{!}{
\begin{tabular}{|l|cc|cc|cc|cc|}
\hline
\multicolumn{1}{|c|}{\multirow{2}{*}{\textbf{Method ($\times 4$)}}} & \multicolumn{2}{c|}{\textbf{BRISQUE$\downarrow$}}                    & \multicolumn{2}{c|}{\textbf{PIQE$\downarrow$}}                       & \multicolumn{2}{c|}{\textbf{NIQE$\downarrow$}}                     & \multicolumn{2}{c|}{\textbf{EndoQM$\downarrow$}}                     \\ \cline{2-9} 
\multicolumn{1}{|c|}{}                                              & \multicolumn{1}{c|}{\textbf{Mean}} & \textbf{95\% CI}                & \multicolumn{1}{c|}{\textbf{Mean}} & \textbf{95\% CI}                & \multicolumn{1}{c|}{\textbf{Mean}} & \textbf{95\% CI}              & \multicolumn{1}{c|}{\textbf{Mean}} & \textbf{95\% CI}                \\ \hline
\textbf{ZSSR \cite{zssr}} & 
\multicolumn{1}{c|}{64.7164}  & {[}64.6051 - 64.8148{]} & 
\multicolumn{1}{c|}{85.2007} & {[}85.1398 - 85.2790{]} & 
\multicolumn{1}{c|}{6.0950}  & {[}6.0733 - 6.1169{]} & 
\multicolumn{1}{c|}{13.1060}   & {[}13.0715 - 13.1407{]} \\ \hline
\textbf{DASR \cite{arr41}}    &
\multicolumn{1}{c|}{63.7744}  & {[}63.6153 - 63.7983{]} & 
\multicolumn{1}{c|}{91.8829} & {[}91.7645 - 91.9571{]} &
\multicolumn{1}{c|}{6.0622}   & {[}6.0479 - 6.0765{]} & 
\multicolumn{1}{c|}{14.2492} & {[}14.2251 - 14.2733{]} \\ \hline
\textbf{dSRVAE \cite{dSRVAE}}  & 
\multicolumn{1}{c|}{56.4932}       & {[}54.7154 - 57.8896{]} &
\multicolumn{1}{c|}{64.9396}       & {[}64.8250 - 65.0689{]} & 
\multicolumn{1}{c|}{\textcolor{red}{4.8916}}        & \textcolor{red}{{[}4.8103 - 4.9731{]}} & 
\multicolumn{1}{c|}{10.8352}       & {[}10.6590 - 11.0115{]} \\ \hline
\textbf{DUSGAN \cite{dusgan}}   & 
\multicolumn{1}{c|}{72.1658}       & {[}72.1032 - 72.1945{]} & 
\multicolumn{1}{c|}{87.5359}       & {[}87.4176 - 87.6289{]} & 
\multicolumn{1}{c|}{7.3791}        & {[}7.3488 - 7.4094{]} & 
\multicolumn{1}{c|}{13.8334}       & {[}13.7974 - 13.8696{]} \\ \hline
\textbf{BSRGAN \cite{bsrgan}}   & 
\multicolumn{1}{c|}{60.9487}       & {[}59.8745 - 61.0132{]}         & \multicolumn{1}{c|}{70.3330}       & {[}70.1321 - 70.3897{]}         & \multicolumn{1}{c|}{5.7289}        & {[}5.5367 - 5.9154{]}         & \multicolumn{1}{c|}{9.6295}        & {[}8.5486 - 9.8756{]}           \\ \hline
\textbf{MDASR \cite{MDASR}}  & 
\multicolumn{1}{c|}{59.3752}       & {[}59.0156 - 59.3967{]}         & \multicolumn{1}{c|}{62.8271}       & {[}62.6256 - 62.8943{]}         & \multicolumn{1}{c|}{5.4123}        & {[}5.3901 - 5.4237{]}         & \multicolumn{1}{c|}{8.9283}        & {[}8.8743 - 8.9867{]}           \\ \hline
\textbf{TUDASR \cite{tudasr}}  &
\multicolumn{1}{c|}{\textcolor{red}{55.2830}} & \textcolor{red}{{[}54.9381 - 56.4579{]}}         & \multicolumn{1}{c|}{\textcolor{blue}{59.4121}}       & \textcolor{blue}{{[}58.7976 - 60.4894{]}}         & \multicolumn{1}{c|}{5.2104}        & {[}5.1831 - 5.2210{]}         & \multicolumn{1}{c|}{\textcolor{blue}{7.6216}}        & \textcolor{blue}{{[}7.5976 - 7.6456{]}}           \\ \hline
\textbf{Proposed}                                                   & \multicolumn{1}{c|}{\textcolor{blue}{56.2884}}  & \textcolor{blue}{{[}56.2734 - 56.2917{]}}         & \multicolumn{1}{c|}{\textcolor{red}{53.1951}}       & \textcolor{red}{{[}53.0953 - 53.2130{]}}         & \multicolumn{1}{c|}{\textcolor{blue}{5.1846}}       & \textcolor{blue}{{[}5.1727 - 5.1966{]}}         & \multicolumn{1}{c|}{\textcolor{red}{6.1204}}        & \textcolor{red}{{[}6.1012 - 6.1396{]}}           \\ \hline
\end{tabular}
}
\end{table*}

To systematically evaluate the statistical significance and reliability of the proposed method (i.e., \emph{CEM-TUDASR}) in comparison to existing state-of-the-art unsupervised SR techniques, we also performed a comprehensive statistical analysis using Analysis of Variance (ANOVA) test and calculated the corresponding $95\%$ confidence intervals (CIs) for various no-reference quality metrics, namely BRISQUE, PIQE, NIQE, and the domain-specific EndoQM. The summarized mean values and associated CIs for each evaluated method on the newly curated Kvasir dataset are presented in Table~\ref{tab:anova}. Moreover, the statistical differences among these methods are visually illustrated via boxplots depicted in Fig.~\ref{fig:anova}. For the BRISQUE metric (see Fig.\ref{fig:anova}(a)), the proposed approach consistently yields lower (better) perceptual quality scores with narrower confidence intervals, highlighting its superior stability and effectiveness compared to competing models. Conversely, methods like dSRVAE, BSRGAN, MDASR, and TUDASR exhibit wider intervals and higher mean values, indicating reduced reliability and perceptual quality consistency. Analyzing the PIQE metric (Fig.~\ref{fig:anova}(b)), the proposed method again surpasses the comparison methods, achieving the lowest mean PIQE value and exhibiting the tightest $95\%$ CI. This strongly suggests reduced perceptual distortion and improved visual coherence. In contrast, methods such as ZSSR and dSRVAE generate notably higher and more variable PIQE scores, indicating limitations in reliability when applied to clinical scenarios. Furthermore, evaluation of the NIQE scores (Fig.\ref{fig:anova}(c)) confirms the superior consistency and reliability of the proposed method, which obtains significantly improved mean NIQE values. This demonstrates its capacity to preserve natural image characteristics effectively. Although the TUDASR and MDASR methods provide competitive performance, their relatively higher mean scores reflect moderate reliability compared to our proposed framework.

The most substantial validation of clinical efficacy is demonstrated through the EndoQM metric (Fig.~\ref{fig:anova}(d)), specifically tailored for endoscopic image assessment. The proposed model achieves the lowest mean EndoQM score accompanied by the narrowest confidence interval, underlining its exceptional capability to reconstruct diagnostically significant details, including subtle textures and intricate vascular structures. Overall, the statistical results derived from ANOVA and the reported confidence intervals affirm that our proposed unsupervised SR method significantly outperforms existing approaches, confirming its superior reliability, robustness, and practical suitability for clinical applications in wireless capsule endoscopy imaging.

\subsection{Ablation Study}

\label{sec:ablation}
\begin{figure*}[t!]
    \centering
    \subfloat[\scriptsize\centering LR]{{\includegraphics[width=3.5cm,height=3.5cm]{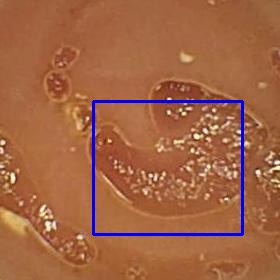} }}%
    \subfloat[\scriptsize\centering Case 1]{{\includegraphics[width=3.5cm,height=3.5cm]{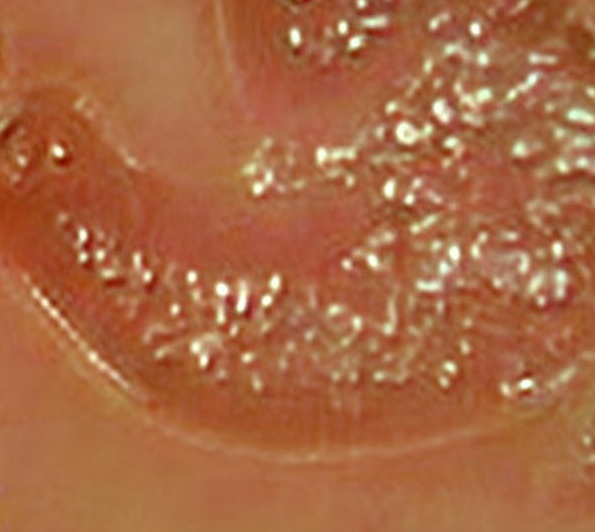} }}%
    \subfloat[\scriptsize\centering Case 2]{{\includegraphics[width=3.5cm,height=3.5cm]{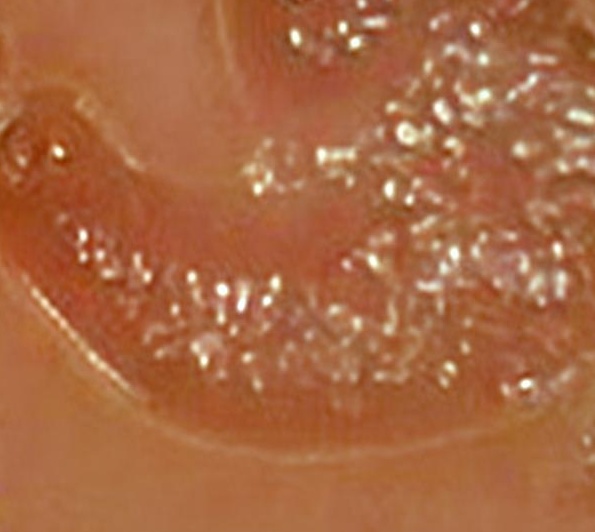}
    }}%
    \subfloat[\scriptsize\centering Case 3 ]{{\includegraphics[width=3.5cm,height=3.5cm]{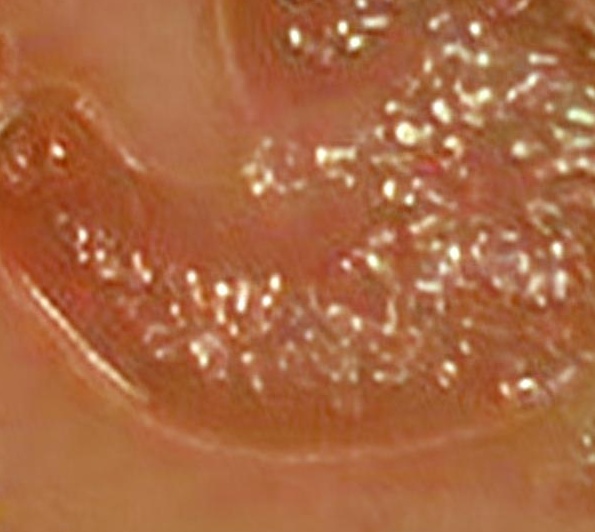} }} %
    \subfloat[\scriptsize\centering Case 4]{{\includegraphics[width=3.5cm,height=3.5cm]{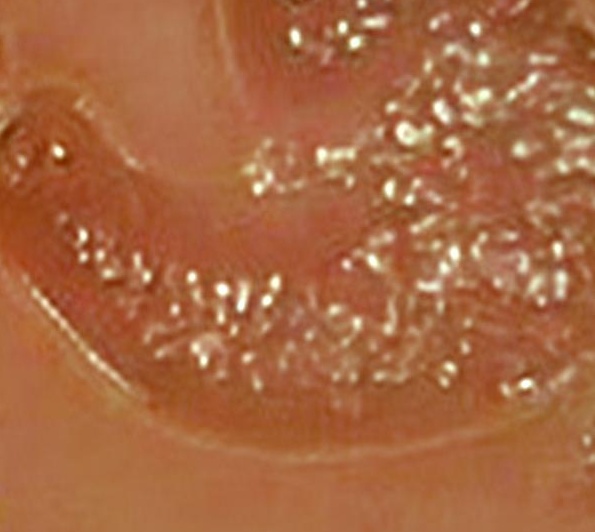} }}\\%
    \subfloat[\scriptsize\centering Case 5 ]{{\includegraphics[width=3.5cm,height=3.5cm]{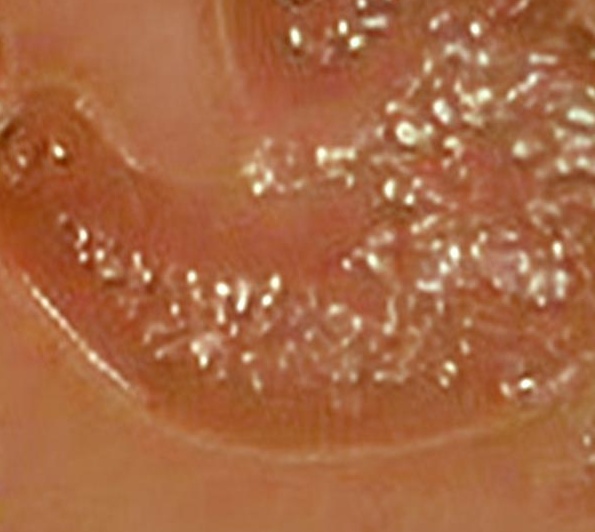} }} %\\
    \subfloat[\scriptsize\centering Case 6]{{\includegraphics[width=3.5cm,height=3.5cm]{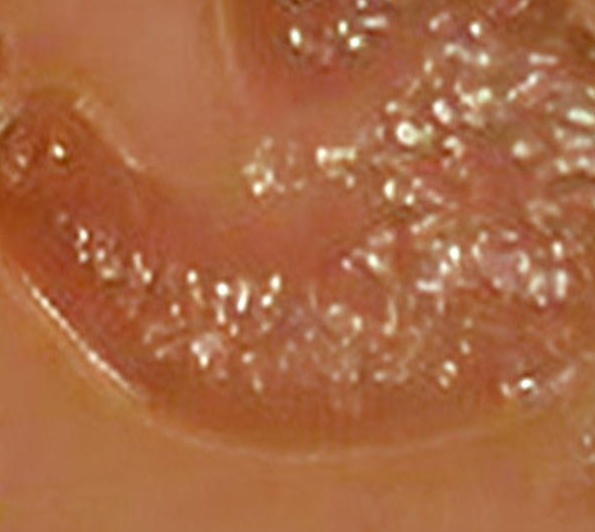} }}%
    \subfloat[\scriptsize\centering Case 7]{{\includegraphics[width=3.5cm,height=3.5cm]{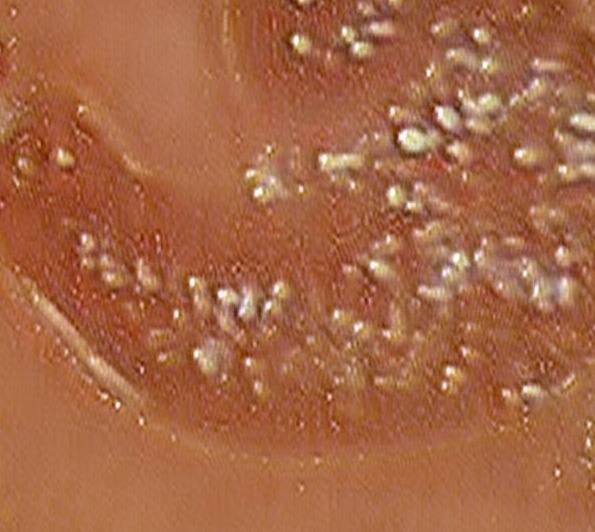} }}%
    \subfloat[\scriptsize\centering Case 8]{{\includegraphics[width=3.5cm,height=3.5cm]{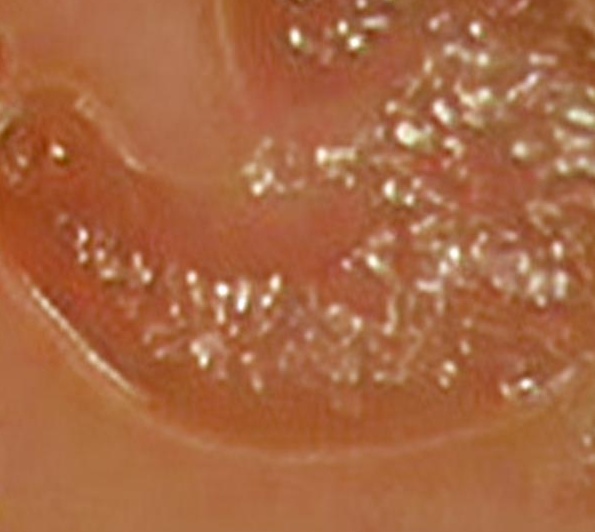} }}%
    \subfloat[\scriptsize\centering Proposed]{{\includegraphics[width=3.5cm,height=3.5cm]{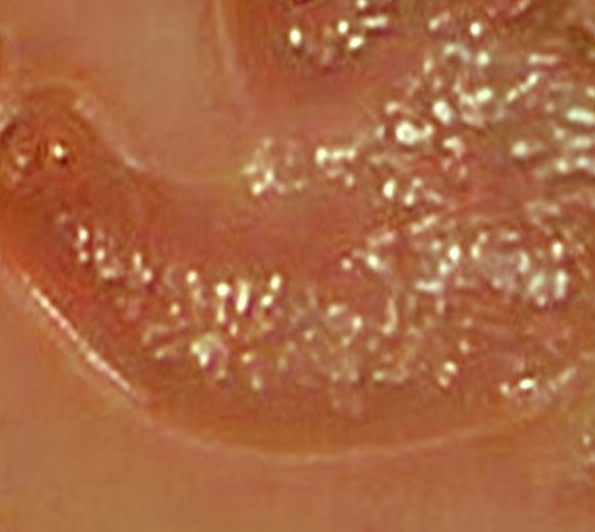} }}%
    \hspace{5cm} % optional: center last 3 images if needed
    \caption{The SR images obtained using the different hyper-parameter settings of the proposed method on the newly derived Kvasir dataset. Also, the details of different cases are mentioned in Table~\ref{tab:ablation}. }%
    \label{fig:ablation1}%
\end{figure*}

\begin{figure*}[t!]
    \centering
    \subfloat[\scriptsize\centering LR]{{\includegraphics[width=3.5cm,height=3.5cm]{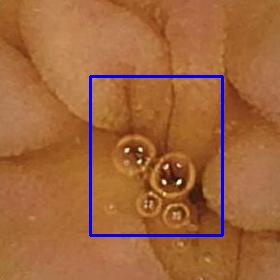} }}%
    \subfloat[\scriptsize\centering Case 1 ]{{\includegraphics[width=3.5cm,height=3.5cm]{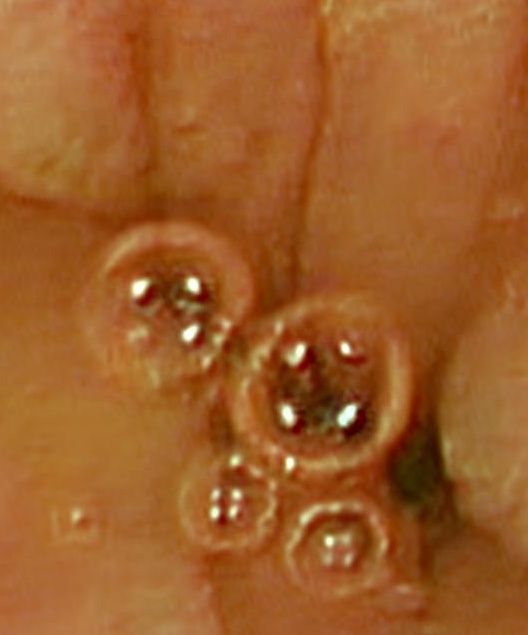} }}%
    \subfloat[\scriptsize\centering Case 2 ]{{\includegraphics[width=3.5cm,height=3.5cm]{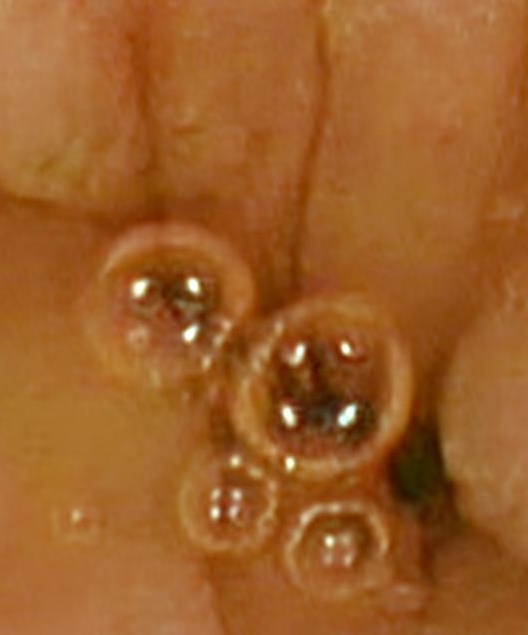} }}%
    \subfloat[\scriptsize\centering Case 3 ]{{\includegraphics[width=3.5cm,height=3.5cm]{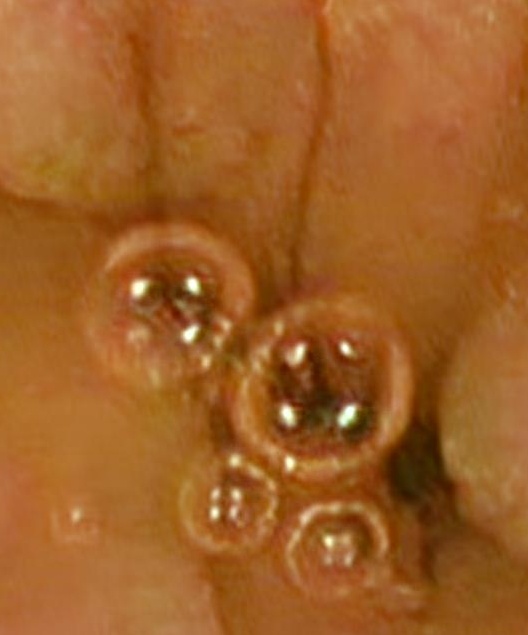} }}
    \subfloat[\scriptsize\centering Case 4]{{\includegraphics[width=3.5cm,height=3.5cm]{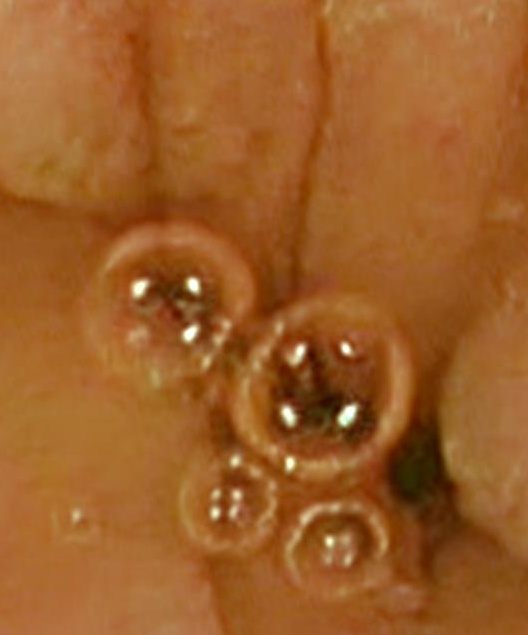} }} \\%
    \subfloat[\scriptsize\centering Case 5 ]{{\includegraphics[width=3.5cm,height=3.5cm]{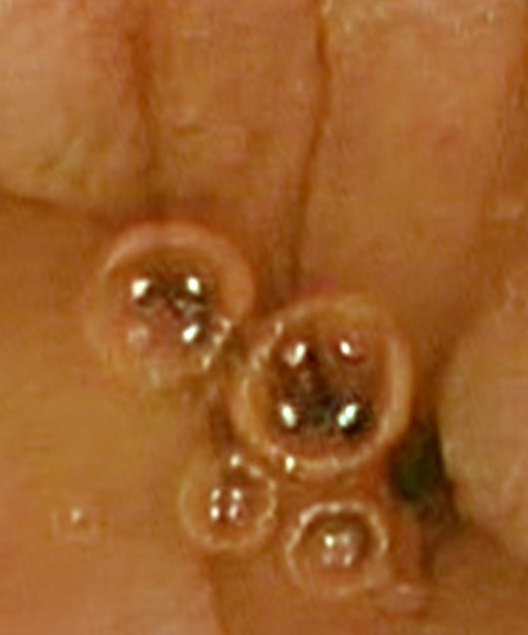} }}%
    \subfloat[\scriptsize\centering Case 6]{{\includegraphics[width=3.5cm,height=3.5cm]{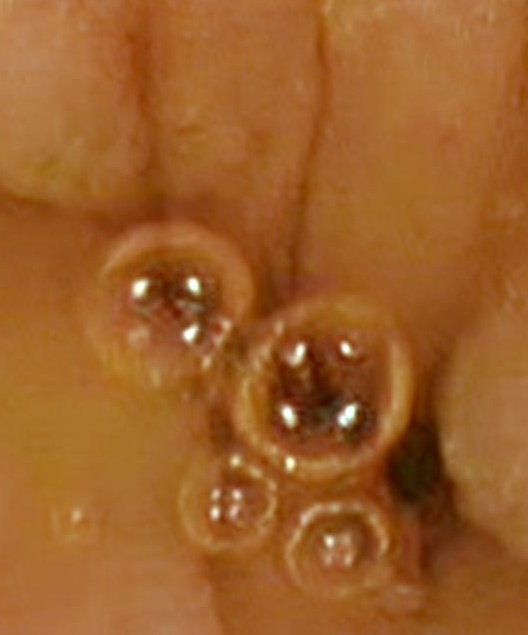} }}%
    \subfloat[\scriptsize\centering Case 7]{{\includegraphics[width=3.5cm,height=3.5cm]{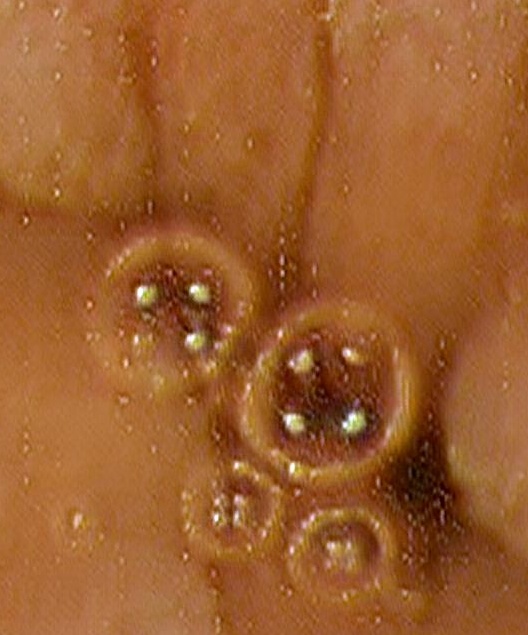} }}%
    \subfloat[\scriptsize\centering Case 8]{{\includegraphics[width=3.5cm,height=3.5cm]{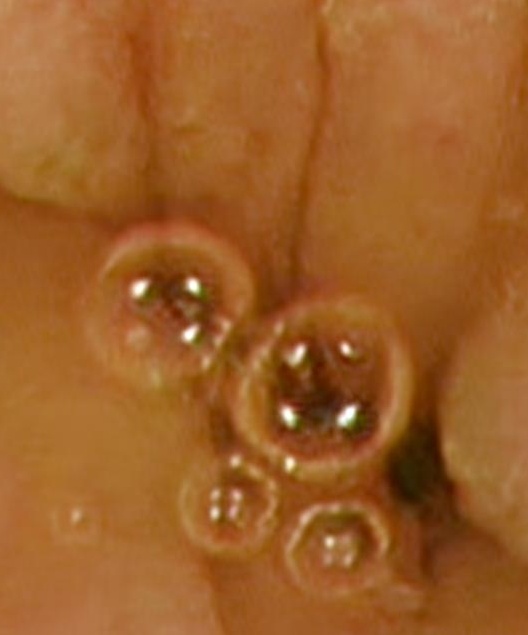} }}% 
    \subfloat[\scriptsize\centering Proposed]{{\includegraphics[width=3.5cm,height=3.5cm]{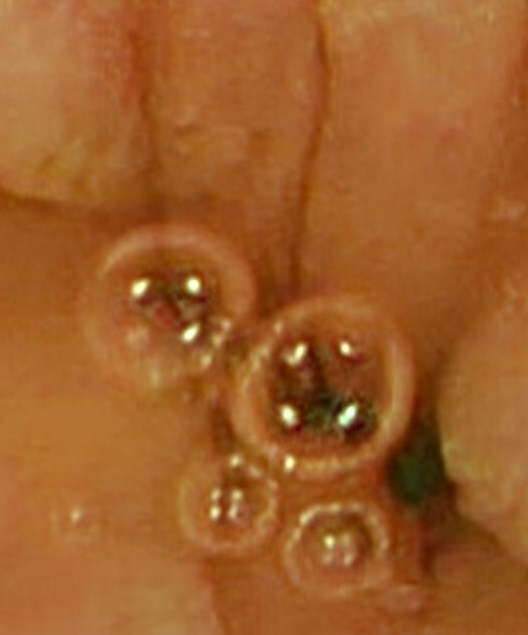} }}%
    \hspace{5cm} % optional: center last 3 images if needed
    \caption{The SR images obtained using the different hyper-parameter settings of the proposed method on the newly derived Kvasir dataset. Also, the details of different cases are mentioned in Table~\ref{tab:ablation}.}%
    \label{fig:ablation2}%
\end{figure*}

\begin{figure*}[t!]
    \centering
    \subfloat[\scriptsize\centering LR]{{\includegraphics[width=3.5cm,height=3.5cm]{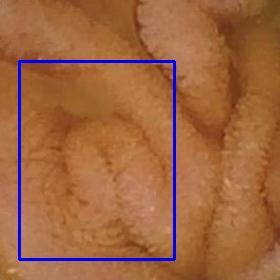} }}%
    \subfloat[\scriptsize\centering Case 1 ]{{\includegraphics[width=3.5cm,height=3.5cm]{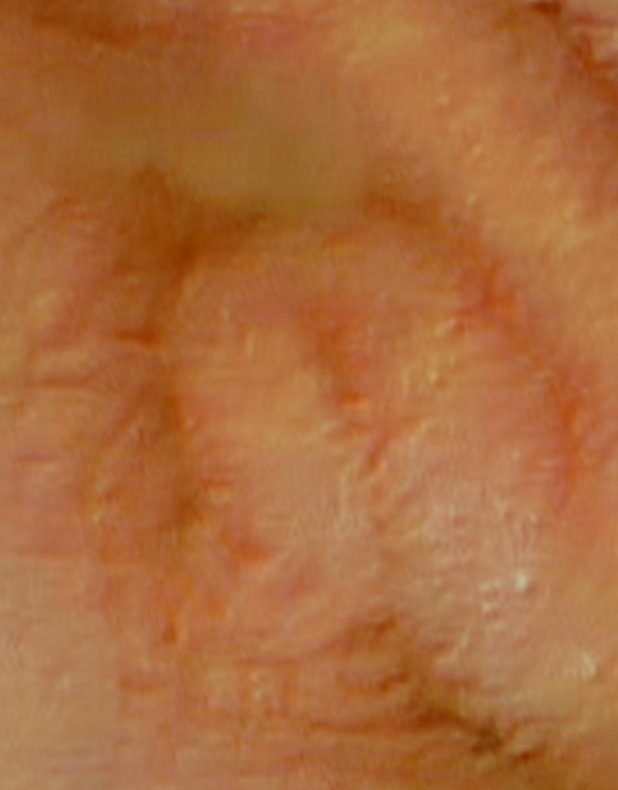} }}%
    \subfloat[\scriptsize\centering Case 2 ]{{\includegraphics[width=3.5cm,height=3.5cm]{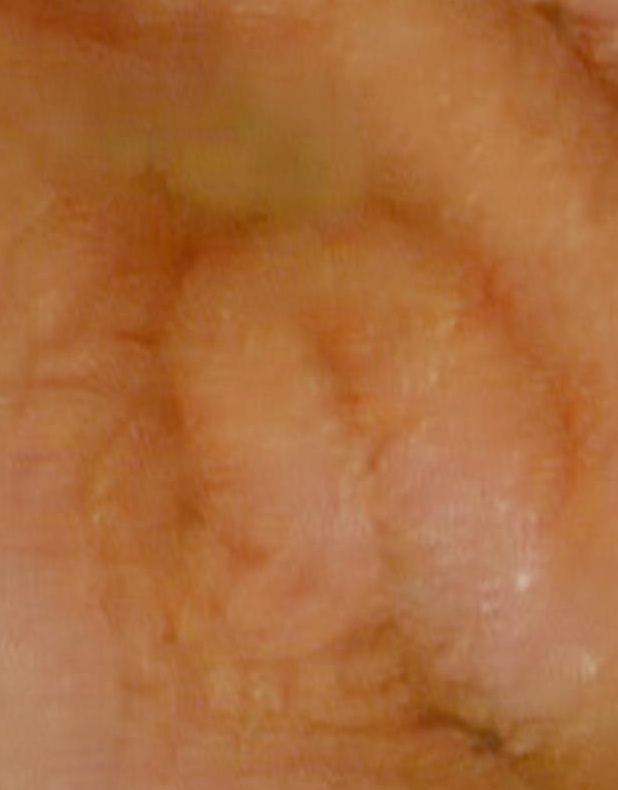} }}%
    \subfloat[\scriptsize\centering Case 3 ]{{\includegraphics[width=3.5cm,height=3.5cm]{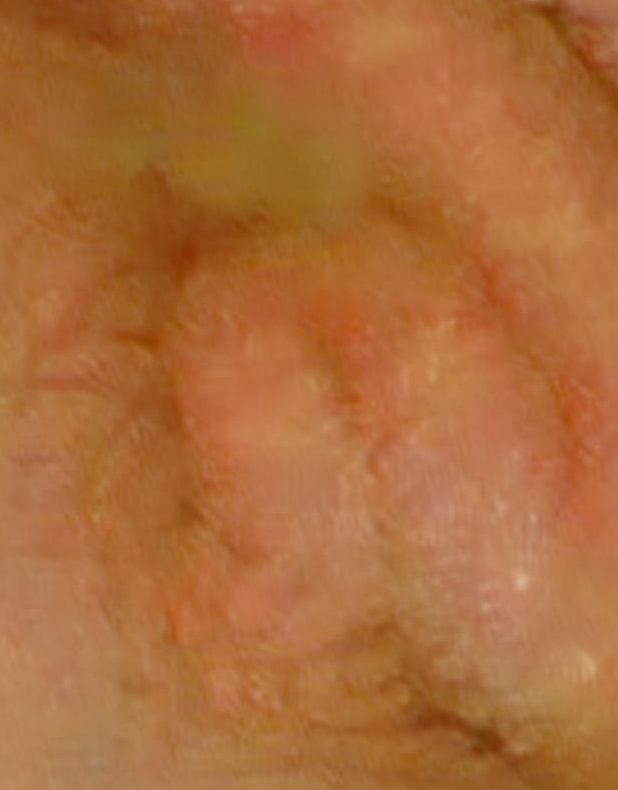} }}
    \subfloat[\scriptsize\centering Case 4]{{\includegraphics[width=3.5cm,height=3.5cm]{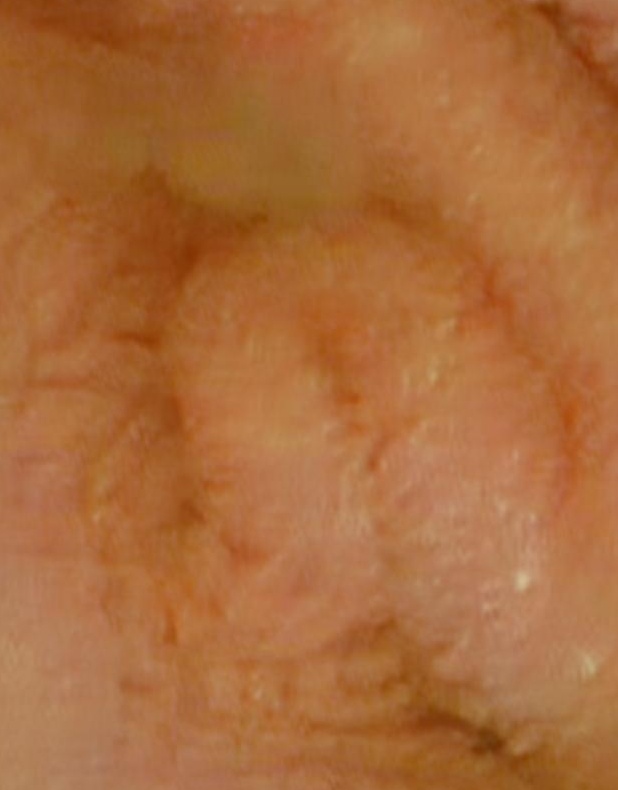} }} \\%
    \subfloat[\scriptsize\centering Case 5 ]{{\includegraphics[width=3.5cm,height=3.5cm]{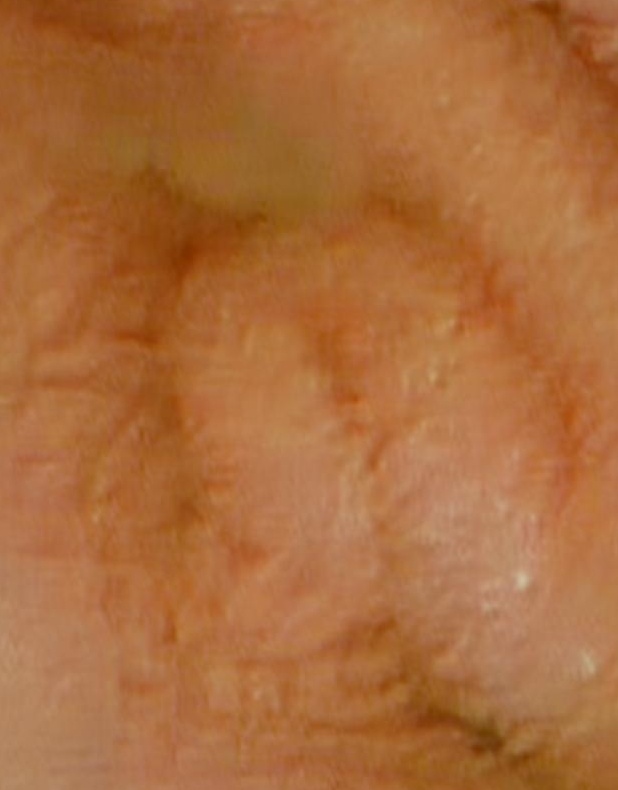} }}%
    \subfloat[\scriptsize\centering Case 6]{{\includegraphics[width=3.5cm,height=3.5cm]{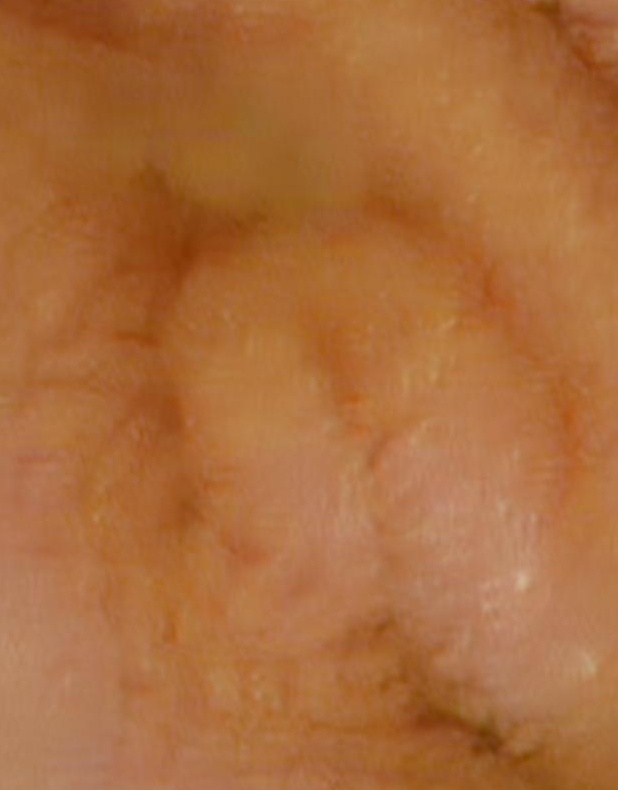} }}%
    \subfloat[\scriptsize\centering Case 7]{{\includegraphics[width=3.5cm,height=3.5cm]{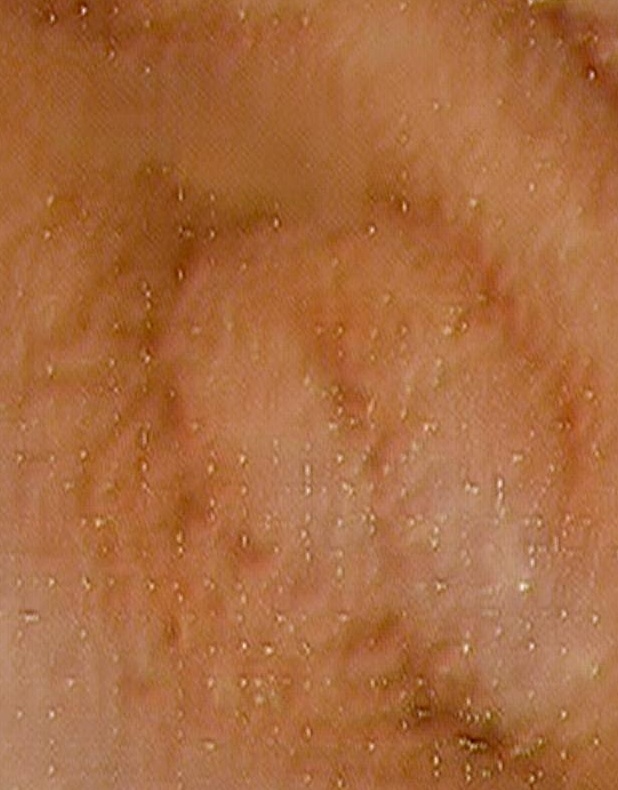} }}%  
    \subfloat[\scriptsize\centering Case 8]{{\includegraphics[width=3.5cm,height=3.5cm]{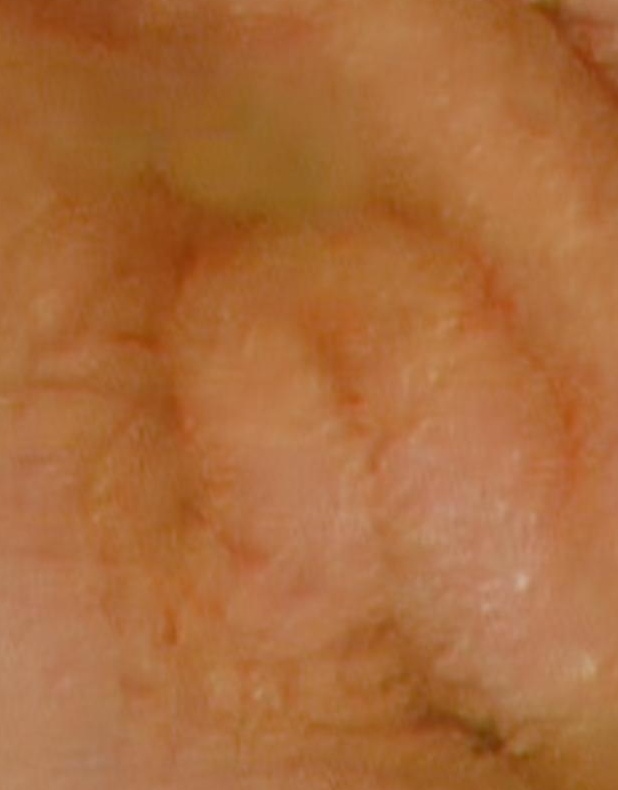} }}%
    \subfloat[\scriptsize\centering Proposed]{{\includegraphics[width=3.5cm,height=3.5cm]{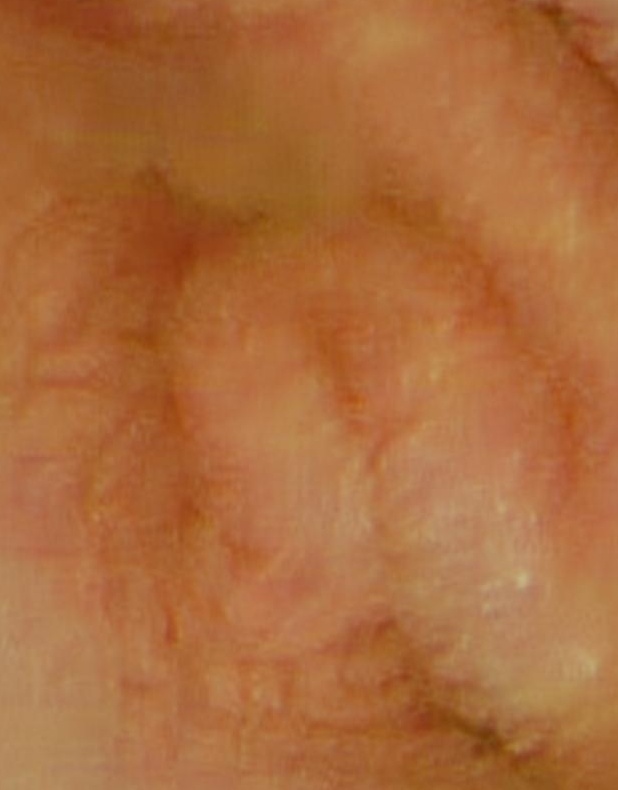} }}%
    \hspace{5cm} % optional: center last 3 images if needed
    \caption{The SR images obtained using the different hyper-parameter settings of the proposed method on the newly derived Kvasir dataset. Also, the details of different cases are mentioned in Table~\ref{tab:ablation}.}%
    \label{fig:ablation3}%
\end{figure*}

\begin{table*}[!t]
\caption{The quantitative analysis under different hyper-parameter settings with no-reference quality assessment measurements. Here, the best value for each measurement is highlighted with bold text.}
\resizebox{\textwidth}{!}{%
\begin{tabular}{|cccccccccccccc|}
\hline
\multicolumn{1}{|c|}{}  & \multicolumn{1}{c|}{}  & \multicolumn{4}{c|}{Newly edited Kvasir Dataset }   & \multicolumn{4}{c|}{KID Dataset}  & \multicolumn{4}{c|}{GIANA Dataset} 
\\ \cline{3-14} 
\multicolumn{1}{|c|}{\multirow{-2}{*}{Case}} & 
\multicolumn{1}{c|}{\multirow{-2}{*}{Configuration}}  & 
\multicolumn{1}{c|}{BRISQUE $\downarrow$}  & 
\multicolumn{1}{c|}{PIQE $\downarrow$} & 
\multicolumn{1}{c|}{NIQE $\downarrow$}  & 
\multicolumn{1}{c|}{EndoQM $\downarrow$} & 
\multicolumn{1}{c|}{BRISQUE $\downarrow$} & 
\multicolumn{1}{c|}{PIQE $\downarrow$} & 
\multicolumn{1}{c|}{NIQE $\downarrow$}  & 
\multicolumn{1}{c|}{EndoQM $\downarrow$}  & 
\multicolumn{1}{c|}{BRISQUE $\downarrow$} & 
\multicolumn{1}{c|}{PIQE $\downarrow$} & 
\multicolumn{1}{c|}{NIQE $\downarrow$} & 
EndoQM $\downarrow$ \\ \hline
\multicolumn{14}{|l|}{ \textbf{ Based on Network Configuration}} 
\\ \hline

\multicolumn{1}{|c|}{Case 1}
& \multicolumn{1}{c|}{Without ESA} 
& \multicolumn{1}{c|}{56.9043}  
& \multicolumn{1}{c|}{55.0374}  
& \multicolumn{1}{c|}{5.2897}  
& \multicolumn{1}{c|}{6.2219}  
& \multicolumn{1}{c|}{66.8553}     
& \multicolumn{1}{c|}{56.6072} 
& \multicolumn{1}{c|}{{4.3621}} 
& \multicolumn{1}{c|}{{5.8883}}
& \multicolumn{1}{c|}{67.6908} 
& \multicolumn{1}{c|}{72.7603}        
& \multicolumn{1}{c|}{4.2665} 
&      6.1006  \\ \hline

\multicolumn{1}{|c|}{Case 2} 
& \multicolumn{1}{c|}{Without EA}                
& \multicolumn{1}{c|}{{57.5895}} 
& \multicolumn{1}{c|}{{57.6048}} 
& \multicolumn{1}{c|}{5.3269} 
& \multicolumn{1}{c|}{6.3485} 
& \multicolumn{1}{c|}{{65.7491 }} 
& \multicolumn{1}{c|}{59.8080} 
& \multicolumn{1}{c|}{{4.3697 }} 
& \multicolumn{1}{c|}{{6.0534}} 
& \multicolumn{1}{c|}{{68.1151}} 
& \multicolumn{1}{c|}{{73.8030}} 
& \multicolumn{1}{c|}{{{4.2499}}} 
& {6.2229} \\ \hline

\multicolumn{1}{|c|}{Case 3} 
& \multicolumn{1}{c|}{Without Fusion Attention  block}
& \multicolumn{1}{c|}{56.5844}
& \multicolumn{1}{c|}{{57.3842 }} 
& \multicolumn{1}{c|}{5.4969} 
& \multicolumn{1}{c|}{6.3332}  
& \multicolumn{1}{c|}{{{60.5608}}} 
& \multicolumn{1}{c|}{53.6222}
& \multicolumn{1}{c|}{{\textbf{4.2931}}} 
& \multicolumn{1}{c|}{5.8749}  
& \multicolumn{1}{c|}{{{\textbf{62.3201}}}} 
& \multicolumn{1}{c|}{{72.6567}} 
& \multicolumn{1}{c|}{{\textbf{4.0879}}} 
& {6.4360} \\ \hline

\multicolumn{1}{|c|}{Case 4}  
& \multicolumn{1}{c|}{With average pooling} 
& \multicolumn{1}{c|}{56.7365} 
& \multicolumn{1}{c|}{55.5568}  
& \multicolumn{1}{c|}{5.9235}
& \multicolumn{1}{c|}{{6.1646}}
& \multicolumn{1}{c|}{{68.1405}}
& \multicolumn{1}{c|}{{ 54.3761}}
& \multicolumn{1}{c|}{4.4659}
& \multicolumn{1}{c|}{{5.8721}}  
& \multicolumn{1}{c|}{{69.4992}}
& \multicolumn{1}{c|}{{72.4371}} 
& \multicolumn{1}{c|}{4.2612} 
& 6.1771  \\ \hline

\multicolumn{1}{|c|}{Case 5} 
& \multicolumn{1}{c|}{With more DAB}
& \multicolumn{1}{c|}{{57.3521}}  
& \multicolumn{1}{c|}{56.0854} 
& \multicolumn{1}{c|}{{5.3050}} 
& \multicolumn{1}{c|}{{6.3949}}
& \multicolumn{1}{c|}{{{63.7071}}}
& \multicolumn{1}{c|}{{55.7091}}
& \multicolumn{1}{c|}{4.4890}  
& \multicolumn{1}{c|}{5.9652} 
& \multicolumn{1}{c|}{{{68.2038}}}
& \multicolumn{1}{c|}{74.2728} 
& \multicolumn{1}{c|}{4.3575}  
& 6.2351 \\ \hline

\multicolumn{1}{|c|}{Case 6} 
& \multicolumn{1}{c|}{With less DAB} 
& \multicolumn{1}{c|}{58.0501} 
& \multicolumn{1}{c|}{57.2533} 
& \multicolumn{1}{c|}{5.9629}
& \multicolumn{1}{c|}{{6.4154}} 
& \multicolumn{1}{c|}{{67.1914}} 
& \multicolumn{1}{c|}{{56.4973}} 
& \multicolumn{1}{c|}{4.4609} 
& \multicolumn{1}{c|}{6.1374} 
& \multicolumn{1}{c|}{{69.3907}} 
& \multicolumn{1}{c|}{74.0669}  
& \multicolumn{1}{c|}{4.2837} 
& 6.2651  \\ \hline

%\multicolumn{1}{|c|}{}                       & \multicolumn{1}{c|}{Proposed}                         & \multicolumn{1}{c|}{{\textbf{ 42.0358}}} & \multicolumn{1}{c|}{{ \textbf{41.8822}}} & \multicolumn{1}{c|}{{ \textbf{4.5346}}} & \multicolumn{1}{c|}{{\textbf{ 7.0914}}} & \multicolumn{1}{c|}{{ \textbf{44.5833}}} & \multicolumn{1}{c|}{{ 29.3739}} & \multicolumn{1}{c|}{{ \textbf{4.0198}}} & \multicolumn{1}{c|}{{ \textbf{6.7696}}} & \multicolumn{1}{c|}{{ \textbf{53.2019}}} & \multicolumn{1}{c|}{{ \textbf{29.7511}}} & \multicolumn{1}{c|}{{\textbf{3.7547}}} & {\textbf{6.9723}} \\ \hline
\multicolumn{14}{|l|}{\textbf{Based on loss Configuration}}  \\ 
\hline

\multicolumn{1}{|c|}{Case 7}
& \multicolumn{1}{c|}{$\lambda_1 < \lambda_2$}
& \multicolumn{1}{c|}{59.0010} 
& \multicolumn{1}{c|}{59.0225} 
& \multicolumn{1}{c|}{5.3748}
& \multicolumn{1}{c|}{6.2002} 
& \multicolumn{1}{c|}{65.6647} 
& \multicolumn{1}{c|}{56.7225} 
& \multicolumn{1}{c|}{4.5857} 
& \multicolumn{1}{c|}{6.2623}
& \multicolumn{1}{c|}{62.7402} 
& \multicolumn{1}{c|}{74.6706}
& \multicolumn{1}{c|}{4.2903} 
& 6.7477                     \\ \hline

\multicolumn{1}{|c|}{Case 8} 
& \multicolumn{1}{c|}{$\lambda_1 = \lambda_2$}
& \multicolumn{1}{c|}{58.2988} 
& \multicolumn{1}{c|}{54.7363}
& \multicolumn{1}{c|}{5.3024} 
& \multicolumn{1}{c|}{6.2585} 
& \multicolumn{1}{c|}{67.1243} 
& \multicolumn{1}{c|}{{57.4862}}
& \multicolumn{1}{c|}{4.4539} 
& \multicolumn{1}{c|}{6.1037} 
& \multicolumn{1}{c|}{69.6612} 
& \multicolumn{1}{c|}{{75.1094}} 
& \multicolumn{1}{c|}{4.3076} 
& 6.2798                      \\ \hline

\multicolumn{1}{|c|}{} 
& \multicolumn{1}{c|}{Proposed} 
& \multicolumn{1}{c|}{{\textbf{56.2884}}} 
& \multicolumn{1}{c|}{\textbf{53.1951}}
& \multicolumn{1}{c|}{\textbf{5.1846}} 
& \multicolumn{1}{c|}{{\textbf{6.1204}}}
& \multicolumn{1}{c|}{{\textbf{57.3572} }} 
& \multicolumn{1}{c|}{\textbf{50.3821}}
& \multicolumn{1}{c|}{{4.5140 }} 
& \multicolumn{1}{c|}{{\textbf{5.8679}}}
& \multicolumn{1}{c|}{{68.0625 }} 
& \multicolumn{1}{c|}{{ \textbf{72.3912}}}
& \multicolumn{1}{c|}{{4.1987}} 
&  {\textbf{5.9021}} \\ \hline
\end{tabular}%
}
\label{tab:ablation}
\end{table*}

\begin{table*}[!t]
\centering
\caption{Hyperparameter configuration and tuning strategy of the proposed \emph{CEM-TUDASR} framework.}
\label{tab:hyperparameter}
\resizebox{\textwidth}{!}{
\begin{tabular}{p{3.2cm} p{4.0cm} p{3.5cm} p{2.8cm} p{5.0cm}}
\hline
\textbf{Hyperparameter} & \textbf{Description} & \textbf{Values / Range Tested} & \textbf{Final Selected Value} & \textbf{Tuning Methodology} \\
\hline
% Scale Factor 
% & Super-resolution upscaling factor 
% & $\times2$, $\times4$, $\times8$ 
% & $\times4$ 
% & Selected based on clinical relevance and best reconstruction quality for WCE images \\
\hline
Batch Size 
& Number of training samples per iteration 
& 2, 4, 8 
& 4 
& Chosen based on GPU memory constraints and stable adversarial training \\
\hline
Patch Size (Downsampling Model) 
& Input patch size for degradation model training 
& 128, 256, 512 
& 256 
& Selected for balanced contextual feature learning and computational efficiency \\
\hline
Crop Size (Testing) 
& Patch size used during patch-based inference 
& 64, 130, 256 
& 130 
& Selected to minimize border artifacts and improve Gaussian patch blending \\
\hline
Learning Rate 
& Initial learning rate for optimization 
& $1\times10^{-3}$, $0.5\times10^{-5}$, $1\times10^{-4}$ 
& $0.5\times10^{-5}$ 
& Finalized based on stable convergence and validation performance \\
\hline
Optimizer 
& Optimization strategy for generator and discriminator 
& Adam, SGD, RMSProp 
& Adam 
& Selected due to stable convergence and robustness in GAN-based training \\
\hline
Number of DABs  
& Number of deep attention blocks 
& 3, 5, 7 
& 5 
& Validated through ablation study; 5 blocks achieved the best trade-off between performance and complexity \\
\hline
Number of Feature Channels 
& Number of feature maps in SR generator 
& 64, 128, 256 
& 128 
& Selected based on superior reconstruction quality and computational efficiency \\
\hline
Efficient Spatial Attention (ESA) 
& Spatial saliency enhancement module 
& With / Without 
& With ESA 
& Ablation study showed ESA improves BRISQUE, PIQE, NIQE, and EndoQM \\
\hline
Efficient Attention (EA) 
& Long-range contextual dependency modeling 
& With / Without 
& With EA 
& Selected based on improved texture preservation and structural consistency \\
\hline
Fusion Attention Block (FAB) 
& Final fusion of bicubic LR and generated SR image 
& With / Without 
& With FAB 
& Ablation confirmed improved adaptive feature fusion and perceptual quality \\
\hline
% Deep Attention Blocks (DABs) 
% & Hierarchical feature refinement 
% & Full DAB / Less DAB 
% & Full DAB 
% & Selected based on best quantitative and qualitative performance \\
% \hline
Pooling Strategy 
& Feature aggregation method 
& Average Pooling / Max Pooling 
& Max Pooling 
& Max pooling preserved high-frequency information more effectively \\
\hline
% Attention Branch 
% & Auxiliary attention branch selection 
% & With / Without Attention Block 
% & With Attention Block 
% & Included based on improved contextual learning and stability \\
% \hline
Loss Function Configuration 
& Training objective combination 
& $\lambda_1 < \lambda_2$, $\lambda_1 = \lambda_2$, $\lambda_1 > \lambda_2$ 
& $\lambda_1 > \lambda_2$ 
& Selected through loss ablation study using no-reference quality metrics \\
\hline
Adversarial Loss Weight 
& Weight of GAN loss 
& 0.0001, 0.001, 0.01 
& 0.001 
& Chosen to balance perceptual enhancement and anatomical fidelity \\
\hline
% Learning Rate Decay (SR) 
% & Learning rate decay step for SR generator 
% & 200000, 400000, 600000 
% & 400000 
% & Determined using validation stability \\
% \hline
% Learning Rate Decay (Downsampling) 
% & Learning rate decay step for degradation model 
% & 30000, 50000, 70000 
% & 50000 
% & Selected for stable degradation modeling \\
% \hline
% Validation Metrics 
% & Model selection criteria 
% & BRISQUE, PIQE, NIQE, EndoQM 
% & BRISQUE, PIQE, NIQE, EndoQM 
% & Final model selected using no-reference image quality assessment \\
% \hline
Training Strategy 
& Overall optimization pipeline 
& Separate Training / Joint Training 
& Joint Training 
& Selected for better domain adaptation and reconstruction consistency \\
\hline
\end{tabular}}
\end{table*}

To comprehensively analyze the effectiveness of the proposed model (i.e., \emph{CEM-TUDASR}), an ablation study is carried out. This study evaluates the impact of different loss functions and network configurations on the performance across multiple datasets, including the newly edited Kvasir, KID, and GIANA datasets. The evaluation is performed using different reference-less quality metrics: BRISQUE, PIQE, NIQE, and EndoQM which are summarized in Table~\ref{tab:ablation}. Lower metric scores indicate better SR quality in the results. The SR results obtained using these experiments are also depicted in Fig.~\ref{fig:ablation1}, Fig.~\ref{fig:ablation2} and Fig.~\ref{fig:ablation3}  to see their performance in visual aspects. Additionally, Table~\ref{tab:hyperparameter} systematically summarizes the major hyperparameters used for training the proposed CEM-TUDASR framework, including learning rate, batch size, patch size, adversarial loss weight, number of Deep Attention Blocks (DABs), optimizer settings, and training strategy. For each hyperparameter, its description, the range of values considered during experimentation, the final selected value used in the proposed model, and the corresponding tuning strategy adopted during model optimization is provided. These additions significantly improve the reproducibility of the proposed framework and provide clearer guidance for future research and practical implementation.

\begin{itemize}
    \item \textbf{Effectiveness of Efficient Spacial Attention block (i.e., Case 1 in Table~\ref{tab:ablation} and Figs.~\ref{fig:ablation1}(b) \&~\ref{fig:ablation2}(b))} \newline
    To assess the effectiveness of the Efficient Spatial Attention (ESA) block embedded within the Transformer module, an ablation study is conducted. This block selectively emphasizes diagnostically relevant spatial features while suppressing irrelevant background information, crucial for medical imaging tasks such as WCE and retinal image analysis. The qualitative comparisons of this study indicate that the removal of ESA block noticeably reduces visual quality, leading to blurred edges and loss of subtle anatomical details. Conversely, incorporating ESA enhances clarity and sharpness, accurately preserving critical textures and structural boundaries. Quantitatively, the presence of ESA consistently achieves lower BRISQUE, PIQE, and NIQE scores across all tested datasets, validating improved perceptual quality. Additionally, domain-specific EndoQM scores further confirm that ESA significantly aids in preserving medically relevant details, underscoring its importance in enhancing the overall clinical performance of the proposed model.

    \item \textbf{Effectiveness of Efficient Attention (EA) block (i.e., Case 2 in Table~\ref{tab:ablation} and Figs.~\ref{fig:ablation1}(c) \&~\ref{fig:ablation2}(c))} \newline
    To evaluate the impact of the Efficient Attention (EA) block integrated within our proposed framework (i.e., \emph{CEM-TUDASR}), we performed an ablation study by removing the EA module and assessing the resulting performance differences qualitatively and quantitatively. The EA block effectively models long-range spatial and channel-wise dependencies with minimal computational overhead, providing enhanced global context. Here, visual results demonstrate that exclusion of the EA block leads to noticeably inferior reconstructions, characterized by reduced structural coherence and loss of critical anatomical details such as lesion boundaries and fine vascular textures. Incorporation of the EA block significantly improves visual sharpness, structural continuity, and textural consistency. Quantitative analysis further corroborates these improvements, as evidenced by lower BRISQUE, NIQE, and PIQE scores across all evaluated datasets. Additionally, domain-specific EndoQM scores indicate enhanced preservation of medically relevant features. These findings highlight the crucial role of the Efficient Attention module in balancing effective global context modeling with computational efficiency, thereby significantly enhancing the clinical applicability and quality of SR outputs.

    \item \textbf{Effectiveness of Fusion Attention  Block (i.e., Case 3 in Table~\ref{tab:ablation} and Figs.~\ref{fig:ablation1}(d) \&~\ref{fig:ablation2}(d))} \newline
    We assessed the Fusion Attention Block (FAB), a critical module designed to adaptively integrate features from the bicubic-upsampled image and learned high-frequency representations. The FAB employs dual attention (channel and spatial) to selectively enhance diagnostically relevant anatomical structures and textures. When comparing the proposed model against variants lacking FAB—where fusion was performed via simple addition or concatenation without attention—we observed qualitatively reduced visual quality, including diminished lesion boundaries, vascular clarity, and mucosal textures. Conversely, incorporating FAB improved perceptual fidelity, structural continuity, and edge sharpness. Quantitative evaluation further validated these findings, as evidenced by lower BRISQUE, PIQE, and NIQE scores and improved EndoQM values, demonstrating enhanced preservation of medically significant details. These results underscore FAB's pivotal role in guiding feature integration and emphasizing diagnostically essential regions, significantly enhancing overall reconstruction quality and clinical utility.

     \item \textbf{Effectiveness of Average pooling in FAB (i.e., Case 4 in Table~\ref{tab:ablation} and Figs.~\ref{fig:ablation1}(e) \&~\ref{fig:ablation2}(e))} \newline
    To investigate the impact of pooling operations within the Fusion Attention Block (FAB), we conducted an ablation study by replacing max pooling with average pooling. The max pooling operation effectively highlights salient anatomical features, such as lesion boundaries and vascular structures, by focusing attention on dominant activations. In contrast, average pooling uniformly aggregates spatial information, potentially weakening localized details critical for diagnosis. Qualitative assessments show that average pooling results in visually softer reconstructions, diminished sharpness, and reduced clarity in diagnostically significant regions. Quantitative metrics further validate this degradation, revealing higher BRISQUE, PIQE, and EndoQM scores. These findings confirm the superior efficacy of max pooling over average pooling within FAB, highlighting its importance for generating sharp, clinically relevant super-resolved outputs.

     \item \textbf{Effectiveness of DAB block (i.e., Case 5 and Case 6 in Table~\ref{tab:ablation} and Figs.~\ref{fig:ablation1}(f,g) \&~\ref{fig:ablation2}(f,g)) }\newline
    To evaluate the influence of network depth on performance, an ablation study was conducted by varying the number of Deep Attention Blocks (DABs) within the generator architecture. Specifically, models with fewer DABs (under-parameterized) and more DABs (over-parameterized) were compared against the proposed architecture, which employs an empirically optimized number of blocks. Qualitative analysis reveals that reducing the number of DABs leads to underfitting, resulting in blurred textures and inadequate reconstruction of anatomical details. Conversely, an excessive number of DABs introduces overfitting issues, characterized by oversmoothed images, diminished detail preservation, increased computational complexity, and training instability. Quantitative evaluation across reference-less metrics (BRISQUE, PIQE, NIQE, EndoQM) confirms that both under- and over-parameterized models exhibit inferior performance relative to the optimally configured model. Thus, the chosen depth effectively balances model complexity and reconstruction quality, emphasizing the necessity of carefully tuning the number of attention blocks in medical image super-resolution networks.

    \item \textbf{Importance of different loss weightage (i.e., Case 7 and Case 8 in Table~\ref{tab:ablation} and Figs.~\ref{fig:ablation1}(h,i) \&~\ref{fig:ablation2}(h,i))} \newline
    To examine the effect of the relative weighting between the content loss and adversarial loss, we conducted an ablation study by varying the values of the hyperparameters $\lambda_1$ and $\lambda_2$ in the generator loss function. In the proposed configuration, a higher emphasis is placed on the content loss ($\lambda_1 > \lambda_2$) to ensure the preservation of structural fidelity and high-frequency details in the super-resolved outputs. For comparison, we evaluated two additional configurations: one with equal weighting ($\lambda_1 = \lambda_2$) and another with a dominant adversarial term ($\lambda_1 < \lambda_2$). Qualitative results indicate that setting $\lambda_1 = \lambda_2$ produces outputs with moderately preserved textures but lacks sharpness, while $\lambda_1 < \lambda_2$ leads to visually inconsistent and artifact-prone reconstructions, highlighting the instability introduced by excessive adversarial influence. Quantitative evaluation further supports these findings, with the proposed $\lambda_1 > \lambda_2$ setting achieving the lowest scores across BRISQUE, PIQE, NIQE, and EndoQM metrics. These results demonstrate that prioritizing content fidelity while maintaining a weaker adversarial signal yields perceptually and diagnostically superior SR outputs.
    
\end{itemize}

\begin{figure}[]
    \centering
    {\includegraphics[width=9 cm, height=7cm]{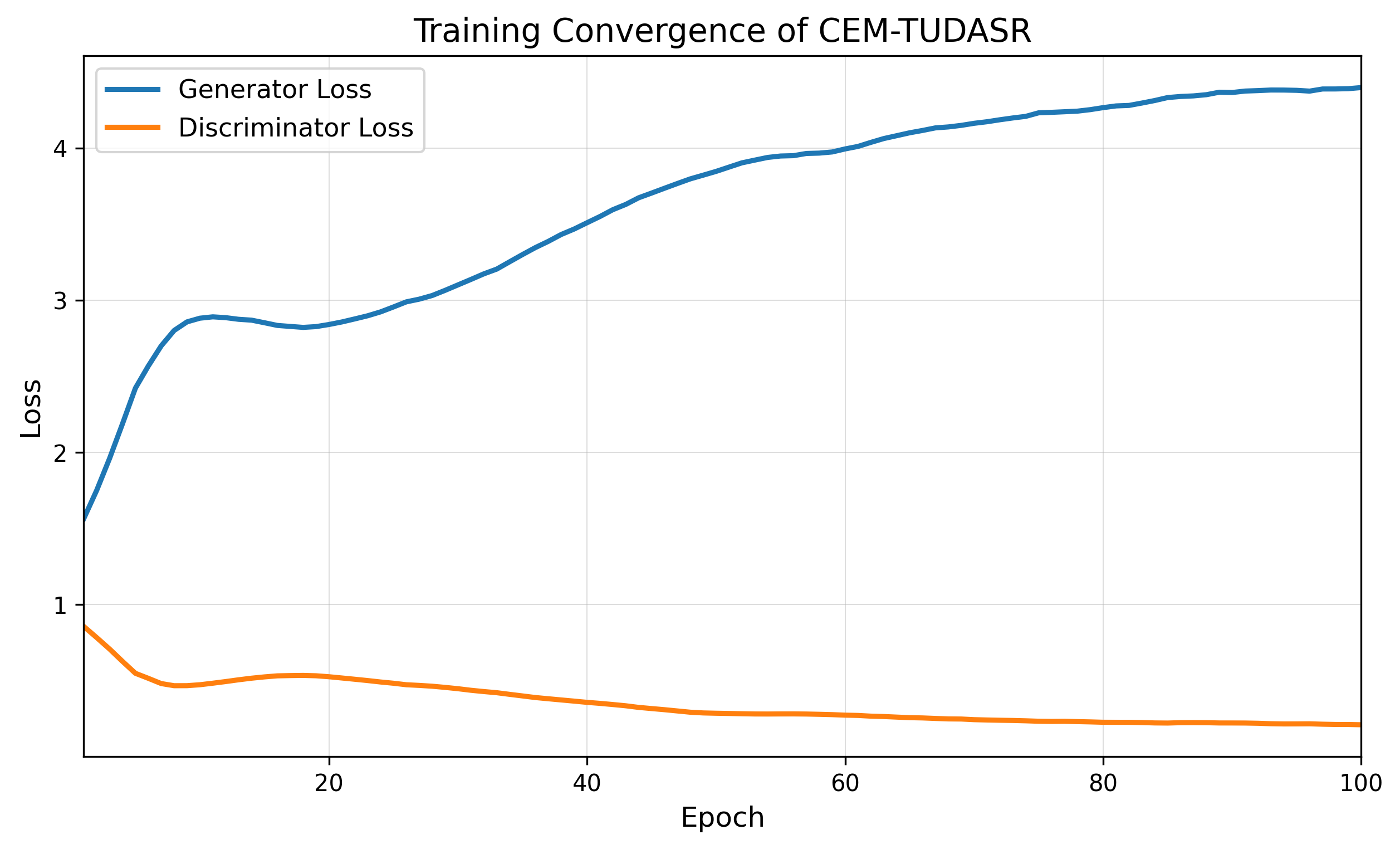}}
   \caption{Training convergence analysis of the proposed \emph{CEM-TUDASR} framework
}
    \label{fig:training_loss}
\end{figure}
To further analyze the optimization behavior and training stability of the proposed \emph{CEM-TUDASR} framework, a convergence analysis based on adversarial loss dynamics is performed. Fig.~\ref{fig:training_loss} illustrating the evolution of generator loss and discriminator loss across epochs. The results demonstrate stable adversarial optimization, where the discriminator loss decreases during initial training and gradually stabilizes, while the generator loss reaches a steady equilibrium, indicating balanced learning without mode collapse. The obtained convergence curves demonstrate stable adversarial optimization throughout the training process. Specifically, the discriminator loss initially decreases during the early training stages, indicating that the discriminator effectively learns to distinguish between generated SR images and target HR representations. As training progresses, the discriminator loss gradually stabilizes, reflecting the establishment of a balanced adversarial learning process. Simultaneously, the generator loss exhibits a gradual increase followed by stable convergence, which is characteristic of balanced GAN optimization where the generator progressively improves its capability to reconstruct perceptually realistic and structurally consistent SR images.
\subsection{Computational Complexity}

\begin{table}[!t]
\setlength{\tabcolsep}{1pt}
\centering
\caption{The number of parameters of different unsupervised real-world SR models.}
\label{tab:compu_param}
\vspace{-0.2cm}
\begin{tabular}{|c|c|c|}
\hline
\textbf{Method} & \textbf{Parameters (in million ($M$))} & \textbf{Flops (in GIGA)} \\ \hline
ZSSR \cite{zssr}           & 0.22                             &   2.31                       \\ \hline
DASR \cite{arr41}            & 11.24                            & 470.83                   \\ \hline
dSRVAE \cite{dSRVAE}         & 7.96                             & 689.95                   \\ \hline
DUSGAN \cite{dusgan}         & 5.86                             & 335.86                   \\ \hline
BSRGAN  \cite{bsrgan}        & 16.70                            & 1405.57                  \\ \hline
MDASR  \cite{MDASR}         & 6.22                             & 628.88                   \\ \hline
TUDASR \cite{tudasr}         & 193.21                           & 8499.19                  \\ \hline
Proposed        & 2.67                             & 169.94                   \\ \hline
\end{tabular}
\vspace{-0.7 cm}
\end{table}

\begin{table*}[t]
\centering
\caption{Layer-wise architectural and computational summary of the proposed \emph{CEM-TUDASR} framework.}
\label{tab:layerwise}
\resizebox{\textwidth}{!}{
\begin{tabular}{p{3.0cm} p{4.0cm} p{2.5cm} p{2.2cm} p{2.2cm} p{2.5cm}}
\hline
\textbf{Module/Layer} & \textbf{Function} & \textbf{Trainable Parameters} & \textbf{GFLOPs} & \textbf{GMACs} & \textbf{Execution Time (ms/img)} \\
\hline
Input LR WCE image 
& Input to SR generator 
& 0 
& -- 
& -- 
& -- \\
\hline
Initial Feature Extraction 
& Extracts shallow spatial features using $3\times3$ convolution and activation 
& 3,584 
& 0.45 
& 0.225 
& 3.42 \\
\hline
Deep Attention Blocks (DABs) 
& Hierarchical feature refinement using 5 EnhancedBB blocks with grouped convolutions and residual learning 
& 13,36,010 
& 92.50  
& 46.25 
& 79.18 \\
\hline
Efficient Attention (EA) 
& Captures long-range contextual dependencies using Q/K/V attention mechanism 
& 66,048  
& 28.60  
& 14.30  
& 67.32  \\
\hline
Efficient Spatial Attention (ESA) 
& Enhances spatially salient regions using pooling, convolution, and sigmoid gating 
& 46,400 
& 1.20  
& 0.60  
& 0.93  \\
\hline
Upsampling Block 
& Performs $\times4$ spatial enlargement using convolution and PixelShuffle$_2$ operations 
& 11,84,131 
& 40.80  
& 20.40  
& 11.84\\
\hline
Fusion Attention Block (FAB) 
& Fuses bicubic-upsampled LR image and generated SR image using QKVFusionBlock 
& 2,088 
& 1.39  
& 0.575  
& 4.90  \\
\hline
\textbf{CEM-TUDASR} 
& \textbf{Unsupervised WCE image super-resolution} 
& \textbf{$\approx$ 2.67M} 
& \textbf{164.94 } 
& \textbf{82.47 } 
& \textbf{167.59 } \\
\hline
\end{tabular}}
\end{table*}

In addition to achieving high reconstruction quality SR results, it is essential for any SR models to maintain computational efficiency, particularly in real-time or resource-constrained clinical applications such as WCE and point-of-care retinal screening. To this end, we compare the computational complexity of the proposed method with several state-of-the-art unsupervised SR methods in terms of two key metrics: the total number of learnable parameters (in millions) and the number of floating point operations (FLOPs, in Giga operations) required during inference. These metrics provide a holistic understanding of the model’s memory footprint and computational demand, which are critical for practical deployment.

As depicted in Table~\ref{tab:compu_param}, \emph{CEM-TUDASR} comprises only $2.67 M$ parameters and requires $169.94$ GFLOPs, making it significantly more efficient than existing SR approaches. In comparison, models such as DASR and dSRVAE require $11.24$M and $7.96M$ parameters, respectively, and incur FLOP counts of $470.83$ and $689.95$ Giga operations. BSRGAN, one of the more complex baselines, demands $16.70M$ parameters and $1405.57$ GFLOPs, reflecting high computational and memory overheads that limit its usability in real-time settings. Notably, TUDASR, a Transformer-based endoscopy-specific SR model, exhibits the highest complexity with $193.21M$ parameters and $8499.19$ GFLOPs, making it impractical for deployment in real-time embedded systems or battery-operated devices such as capsule endoscopes. 

Additionally, Table~\ref{tab:layerwise} presents the layer-wise architectural and computational summary of the proposed framework. The analysis includes major modules such as the Initial Feature Extraction layer, Deep Attention Blocks (DABs), Efficient Attention (EA), Efficient Spatial Attention (ESA), Upsampling Block, and Fusion Attention Block (FAB). The proposed model achieves this low computational footprint through the use of a carefully designed lightweight architecture that balances expressiveness with efficiency. First, the integration of a moderate number of Deep Attention Blocks (DABs) allows for deep feature extraction without excessive stacking of layers. Second, the Transformer-based components within the DAB are implemented using Efficient Attention (EA) and Efficient Spatial Attention (ESA) modules, which reduce the computational overhead typically associated with self-attention mechanisms by leveraging downsampled intermediate representations and linear attention transformations. Third, the Fusion Attention Block (FAB) is designed to perform spatial refinement using only a minimal number of convolutional layers and a lightweight self-attention path, further reducing parameter count while maintaining high-frequency detail reconstruction. Further, Execution-time profiling shows that the tracked neural network operations require approximately 167.59 ms per image, whereas the overall end-to-end inference time is measured as 4500 ms per image during patch-based super-resolution testing. The difference arises because the total execution time includes additional operations such as patch extraction, patch merging, Gaussian blending, bicubic interpolation, CPU–GPU data transfer, memory movement, and image input/output operations. These auxiliary steps are necessary for high-resolution image reconstruction in practical deployment settings. Consistent with the objective of achieving computationally efficient and perceptually robust WCE image super-resolution, the proposed framework demonstrates superior perceptual quality (BRISQUE, PIQE, NIQE) and clinical relevance (EndoQM) across all evaluated datasets. Despite its lightweight architecture and low parameter complexity, the proposed method achieves efficient reconstruction performance suitable for real-time deployment in portable and embedded medical imaging systems without compromising diagnostic structural fidelity. 

\subsection{Discussion and Limitations}

Despite the promising performance of the proposed \emph{CEM-TUDASR} framework for unsupervised super-resolution of WCE images, several limitations remain that merit further investigation. The accurate reconstruction of extremely fine-grained anatomical structures, such as subtle vascular patterns, low-contrast mucosal textures, and small lesion boundaries, remains challenging under severe degradation conditions. Since WCE images are often affected by motion blur, sensor noise, non-uniform illumination, and compression artifacts, recovering diagnostically significant high-frequency details from highly degraded inputs remains an inherently ill-posed problem. The performance of the proposed framework is strongly influenced by the effectiveness of the learned degradation model. The degradation network is designed to generate WCE-like low-resolution images from high-resolution conventional endoscopy data. However, it may not fully capture all complex real-world degradation patterns associated with different capsule devices, acquisition protocols, patient-specific variations, and illumination inconsistencies. This limitation may affect the generalization capability of the model when applied to unseen clinical datasets with significantly different imaging characteristics.
Furthermore, although the integration of transformer-based modules such as Deep Attention Blocks (DABs) significantly improves the modeling of long-range contextual dependencies, it introduces additional computational complexity compared to conventional CNN-based super-resolution methods. Similar to other GAN-based super-resolution frameworks, the proposed method may exhibit a trade-off between perceptual quality and anatomically faithful reconstruction. In certain cases, the network may generate over-enhanced textures or visually plausible details that may not perfectly correspond to the true underlying anatomical structures. In medical imaging applications, such deviations must be carefully considered, as they may influence clinical interpretation and diagnostic confidence.
\section{Conclusion}
This paper presents an unsupervised Transformer-based SR framework i.e., \emph{CEM-TUDASR} tailored for WCE images, addressing key challenges such as the absence of paired LR-HR datasets, fine-detail preservation, and domain-specific degradation. By incorporating a degradation model inspired from the TUDASR framework and introducing efficient attention mechanisms—such as EA, ESA, and FAB—the proposed architecture effectively reconstructs diagnostically critical features while maintaining computational efficiency. Through rigorous evaluation on the newly curated Kvasir SR dataset and external clinical datasets (KID and GIANA), the model consistently outperforms state-of-the-art unsupervised SR methods in both quantitative metrics and perceptual quality. Importantly, the cross-domain testing on retinal images further highlights the model's strong generalization capability across distinct medical imaging modalities. Notably, the model achieves high visual quality at a significantly reduced parameter and FLOP count, making it suitable for deployment in real-time and resource-constrained medical environments. In addition, extensive ablation studies confirm the critical role of each network component, including the importance of attention modules and the optimal configuration of DABs. In summary, the proposed method establishes a strong foundation for unsupervised, domain-adaptive SR in medical imaging and paves the way for future integration into clinical diagnostic workflows.

While the proposed \emph{CEM-TUDASR} framework demonstrates strong performance in unsupervised super-resolution of WCE images, several challenging aspects remain that warrant further investigation. In particular, accurate recovery of extremely fine-grained structures, such as subtle vascular patterns and low-contrast mucosal textures, remains difficult under severe degradations and noise conditions. Additionally, preserving structural consistency without introducing over-enhancement or artificial textures continues to be a critical challenge in GAN-based SR frameworks. 
Future work can focus on improving the robustness of reconstruction under complex, real-world degradations by designing more adaptive and physically consistent degradation models. Enhancing edge preservation and texture realism, especially in clinically significant regions, is another important direction. Furthermore, extending the current framework to leverage temporal information in WCE video sequences could improve reconstruction stability and reduce inconsistencies across frames.
One important avenue for future work is the incorporation of task-driven optimization, where the super-resolved outputs can be jointly optimized with downstream clinical tasks such as lesion detection, bleeding localization, or polyp segmentation. Such a joint learning framework would ensure that the reconstructed images are not only visually enhanced but also clinically more informative.

\vspace{-0.5cm}

\section*{Acknowledgements}
The authors are thankful to the CapsNetwork – International Network for Capsule Imaging in Endoscopy (project no. 322600) funded by the Research Council of Norway.

\bibliographystyle{elsarticle-num}
\bibliography{Reference.bib}

@ARTICLE{EndoL2h,
  author={Almalioglu, Yasin and Bengisu Ozyoruk, Kutsev and Gokce, Abdulkadir and others},
  journal={IEEE TMI}, 
  title={EndoL2H: Deep Super-Resolution for Capsule Endoscopy}, 
  year={2020},
  volume={39},
  number={12},
  pages={4297-4309},
  doi={10.1109/TMI.2020.3016744}}

@ARTICLE{2002,
  author={Karargyris, Alexandros and Bourbakis, Nikolaos},
  journal={IEEE Engineering in Medicine and Biology Magazine}, 
  title={Wireless Capsule Endoscopy and Endoscopic Imaging: A Survey on Various Methodologies Presented}, 
  year={2010},
  volume={29},
  number={1},
  pages={72-83},
  doi={10.1109/MEMB.2009.935466}}

@article{WCE2000,
 author = { G. Iddan and G. Meron and A. Glukhovsky and P. Swain},
title = {Wireless Capsule Endoscopy},
journal = {Nature},
volume = {405},
number = {6785},
pages = {417},
year = {2000},
doi = {https://doi.org/10.1038/35013140},
}

@article{brief,
   title={Deep Learning for Single Image Super-Resolution: A Brief Review},
   volume={21},
   ISSN={1941-0077},
   url={http://dx.doi.org/10.1109/TMM.2019.2919431},
   DOI={10.1109/tmm.2019.2919431},
   number={12},
   journal={IEEE Transactions on Multimedia},
   publisher={Institute of Electrical and Electronics Engineers (IEEE)},
   author={Yang, Wenming and Zhang, Xuechen and Tian, Yapeng and Wang, Wei and Xue, Jing-Hao and Liao, Qingmin},
   year={2019},
   month={Dec},
   pages={3106–3121}
}

@article{Application,
author = {Singh, Amanjot and Singh, Jagroop},
year = {2016},
month = {09},
pages = {6-8},
title = {Super Resolution Applications in Modern Digital Image Processing},
volume = {150},
journal = {International Journal of Computer Applications},
doi = {10.5120/ijca2016911458}
}

@InProceedings{SRCNN,
author="Dong, Chao
and Loy, Chen Change
and He, Kaiming
and Tang, Xiaoou",
editor="Fleet, David
and Pajdla, Tomas
and Schiele, Bernt
and Tuytelaars, Tinne",
title="Learning a Deep Convolutional Network for Image Super-Resolution",
booktitle="Computer Vision -- ECCV 2014",
year="2014",
publisher="Springer International Publishing",
address="Cham",
pages="184--199",
isbn="978-3-319-10593-2"
}

@inproceedings{VDSRN,
  title={Accurate image super-resolution using very deep convolutional networks},
  author={Kim, Jiwon and Lee, Jung Kwon and Lee, Kyoung Mu},
  booktitle={Proceedings of the IEEE conference on CVPR},
  pages={1646--1654},
  year={2016}
}

@INPROCEEDINGS{srdensenet,
  author={Tong, Tong and Li, Gen and Liu, Xiejie and Gao, Qinquan},
  booktitle={2017 IEEE ICCV}, 
  title={Image Super-Resolution Using Dense Skip Connections}, 
  year={2017},
  volume={},
  number={},
  pages={4809-4817},
  doi={10.1109/ICCV.2017.514}}

@article {Swainiv48,
	author = {Swain, P},
	title = {Wireless capsule endoscopy},
	volume = {52},
	number = {suppl 4},
	pages = {iv48--iv50},
	year = {2003},
	doi = {10.1136/gut.52.suppl_4.iv48},
	publisher = {BMJ Publishing Group},
	issn = {0017-5749},
	URL = {https://gut.bmj.com/content/52/suppl_4/iv48},
	eprint = {https://gut.bmj.com/content/52/suppl_4/iv48.full.pdf},
	journal = {Gut}
}

@inproceedings{recursive,
  title={Deeply-recursive convolutional network for image super-resolution},
  author={Kim, Jiwon and Lee, Jung Kwon and Lee, Kyoung Mu},
  booktitle={Proceedings of the IEEE conference on CVPR},
  pages={1637--1645},
  year={2016}
}

@inproceedings{SRGAN,
  title={Photo-realistic single image super-resolution using a generative adversarial network},
  author={Ledig, Christian and Theis, Lucas and Husz{\'a}r, Ferenc and others},
  booktitle={Proceedings of the IEEE conference on CVPR},
  pages={4681--4690},
  year={2017}
}

@article{GAN,
  title={Generative adversarial networks},
  author={Goodfellow, Ian and Pouget-Abadie, Jean and Mirza, Mehdi and Xu, Bing and Warde-Farley, David and Ozair, Sherjil and Courville, Aaron and Bengio, Yoshua},
  journal={Communications of the ACM},
  volume={63},
  number={11},
  pages={139--144},
  year={2020},
  publisher={ACM New York, NY, USA}
}

@inproceedings{RCAN,
  title={Image super-resolution using very deep residual channel attention networks},
  author={Zhang, Yulun and Li, Kunpeng and Li, Kai and others},
  booktitle={Proceedings of the ECCV},
  pages={286--301},
  year={2018}
}

@inproceedings{EDSR,
  title={Enhanced deep residual networks for single image super-resolution},
  author={Lim, Bee and Son, Sanghyun and Kim, Heewon and others},
  booktitle={Proceedings of the IEEE conference on CVPRW},
  pages={136--144},
  year={2017}
}

@article{kvasir,
  title = {Kvasir-Capsule, a video capsule endoscopy dataset},
  author = {Pia H. Smedsrud and Vajira Thambawita and Steven A. Hicks and others},
  journal = {Scientific Data 8},
  year = {2021},
  No={142},
  doi = {https://doi.org/10.1038/s41597-021-00920-z},
}

@article{transformer,
  title={Attention is all you need},
  author={Vaswani, Ashish and Shazeer, Noam and Parmar, Niki and others},
  journal={Advances in neural information processing systems},
  volume={30},
  year={2017}
}

@inproceedings{TTSR,
  title={Learning texture transformer network for image super-resolution},
  author={Yang, Fuzhi and Yang, Huan and Fu, Jianlong and Lu, Hongtao and Guo, Baining},
  booktitle={Proceedings of the IEEE/CVF CVPR},
  pages={5791--5800},
  year={2020}
}

@inproceedings{MDASR,
  title={MDA-SR: Multi-level Domain Adaptation Super-Resolution for Wireless Capsule Endoscopy Images},
  author={Liu, Tianbao and Chen, Zefeiyun and Li, Qingyuan  and others},
  booktitle={International Conference on MICCAI},
  pages={518--527},
  year={2023},
  organization={Springer}
}

@inproceedings{dSRVAE,
  title={Unsupervised real image super-resolution via generative variational autoencoder},
  author={Liu, Zhi-Song and Siu, Wan-Chi and Wang, Li-Wen and others},
  booktitle={Proceedings of the IEEE/CVF CVPR workshops},
  pages={442--443},
  year={2020}
}

@inproceedings{zssr,
  title={“zero-shot” super-resolution using deep internal learning},
  author={Shocher, Assaf and Cohen, Nadav and Irani, Michal},
  booktitle={Proceedings of the IEEE conference on CVPR},
  pages={3118--3126},
  year={2018}
}

@inproceedings{arr41,
  title={Unsupervised degradation representation learning for blind super-resolution},
  author={Wang, Longguang and Wang, Yingqian and Dong, Xiaoyu and others},
  booktitle={Proceedings of the IEEE/CVF Conference on CVPR},
  pages={10581--10590},
  year={2021}
}

@inproceedings{lugmayr2020ntire,
  title={Unsupervised learning for real-world super-resolution},
  author={Lugmayr, Andreas and Danelljan, Martin and Timofte, Radu},
  booktitle={2019 IEEE/CVF ICCVW},
  pages={3408--3416},
  year={2019},
  organization={IEEE}
}

@inproceedings{SRResCGAN,
  title={Deep generative adversarial residual convolutional networks for real-world super-resolution},
  author={Umer, Rao Muhammad and Foresti, Gian Luca and Micheloni, Christian},
  booktitle={Proceedings of the IEEE/CVF Conference on CVPR Workshops},
  pages={438--439},
  year={2020}
}

@inproceedings{USISResNet,
  title={Unsupervised single image super-resolution network (USISResNet) for real-world data using generative adversarial network},
  author={Prajapati, Kalpesh and Chudasama, Vishal and Patel, Heena and others},
  booktitle={Proceedings of the IEEE/CVF Conference on CVPR Workshops},
  pages={464--465},
  year={2020}
}

@article{trans1,
  title={An image is worth 16x16 words: Transformers for image recognition at scale},
  author={Dosovitskiy, Alexey and Beyer, Lucas and Kolesnikov, Alexander and others},
  journal={arXiv preprint arXiv:2010.11929},
  year={2020}
}

@inproceedings{trans2,
  title={Swin transformer: Hierarchical vision transformer using shifted windows},
  author={Liu, Ze and Lin, Yutong and Cao, Yue and others},
  booktitle={Proceedings of the IEEE/CVF ICCV},
  pages={10012--10022},
  year={2021}
}

@Article{KID,
   Author="Koulaouzidis, A.  and Iakovidis, D. K.  and Yung, D. E.  and others ",
   Title="{{K}{I}{D} {P}roject: an internet-based digital video atlas of capsule endoscopy for research purposes}",
   Journal="Endosc Int Open",
   Year="2017",
   Volume="5",
   Number="6",
   Pages="E477-E483",
   Month="Jun"
}

@article{GIANA,
  title={Gastrointestinal Image ANAlysis (GIANA) Angiodysplasia D\&L challenge},
  author={Bernal, Jorge and Aymeric, Histace},
  journal={Web-page of the 2017 Endoscopic Vision Challenge},
  year={2017}
}

@inproceedings{brisque,
  title={Blind/referenceless image spatial quality evaluator},
  author={Mittal, Anish and Moorthy, Anush K and Bovik, Alan C},
  booktitle={2011 conference record of the forty fifth asilomar conference on signals, systems and computers (ASILOMAR)},
  pages={723--727},
  year={2011},
  organization={IEEE}
}

@article{niqe,
  title={Making a “completely blind” image quality analyzer},
  author={Mittal, Anish and Soundararajan, Rajiv and Bovik, Alan C},
  journal={IEEE Signal processing letters},
  volume={20},
  number={3},
  pages={209--212},
  year={2012},
  publisher={IEEE}
}

@inproceedings{piqe,
  title={Blind image quality evaluation using perception based features},
  author={Venkatanath, Narasimhan and Praneeth, D and Bh, Maruthi Chandrasekhar and Channappayya, Sumohana S and Medasani, Swarup S},
  booktitle={2015 twenty first national conference on communications (NCC)},
  pages={1--6},
  year={2015},
  organization={IEEE}
}

@inproceedings{conventional,
  title = {KVASIR: A Multi-Class Image Dataset for Computer Aided Gastrointestinal Disease Detection},
  author = {
     Pogorelov, Konstantin and Randel, Kristin Ranheim and Griwodz, Carsten and others
  },
  booktitle = {Proceedings of the 8th ACM on Multimedia Systems Conference},
  series = {MMSys'17},
  year = {2017},
  isbn = {978-1-4503-5002-0},
  location = {Taipei, Taiwan},
  pages = {164--169},
  numpages = {6},
  doi = {10.1145/3083187.3083212},
  acmid = {3083212},
  publisher = {ACM},
  address = {New York, NY, USA},
}

@INPROCEEDINGS{dcan,
  author={Vaghela, Hiren and Sarvaiya, Anjali and Premlani, Pranav and others},
  booktitle={2023 11th European Workshop on Visual Information Processing (EUVIP)}, 
  title={DCAN:DenseNet with Channel Attention Network for Super-resolution of Wireless Capsule Endoscopy}, 
  year={2023},
  volume={},
  number={},
  pages={1-6},
  doi={10.1109/EUVIP58404.2023.10323037}}

@article{risk,
  title={Risk factors of missed colorectal lesions after colonoscopy},
  author={Lee, Jeonghun and Park, Sung Won and others},
  journal={Medicine},
  volume={96},
  number={27},
  pages={e7468},
  year={2017},
  publisher={LWW}
}

@inproceedings{conv1,
  title={Computationally simple super-resolution algorithm for video from endoscopic capsule},
  author={Duda, Krzysztof and Zielinski, Tomasz and Duplaga, Mariusz},
  booktitle={2008 International Conference on Signals and Electronic Systems},
  pages={197--200},
  year={2008},
  organization={IEEE}
}

@inproceedings{conv2,
  title={POCS-based super-resolution for HD endoscopy video frames},
  author={H{\"a}fner, Michael and Liedlgruber, Michael and Uhl, Andreas},
  booktitle={Proceedings of the 26th IEEE International Symposium on Computer-Based Medical Systems},
  pages={185--190},
  year={2013},
  organization={IEEE}
}

@inproceedings{SISRWCE,
  title={Single image super-resolution via adaptive dictionary pair learning for wireless capsule endoscopy image},
  author={Wang, Yi and Cai, Cheng and Zou, YX},
  booktitle={2015 IEEE International Conference on Digital Signal Processing (DSP)},
  pages={595--599},
  year={2015},
  organization={IEEE}
}

@inproceedings{endodeep1,
  title={Endoscopic image deblurring and super-resolution reconstruction based on deep learning},
  author={Yang, Xirui and Chen, Yue and Tao, Rui and others},
  booktitle={2020 International Conference on Artificial Intelligence and Computer Engineering (ICAICE)},
  pages={168--172},
  year={2020},
  organization={IEEE}
}

@article{endodeep2,
  title={Super-resolution enhanced medical image diagnosis with sample affinity interaction},
  author={Chen, Zhen and Guo, Xiaoqing and Woo, Peter YM and Yuan, Yixuan},
  journal={IEEE TMI},
  volume={40},
  number={5},
  pages={1377--1389},
  year={2021},
  publisher={IEEE}
}

@article{CT1,
  title={Application of super-resolution convolutional neural network for enhancing image resolution in chest CT},
  author={Umehara, Kensuke and Ota, Junko and Ishida, Takayuki},
  journal={Journal of digital imaging},
  volume={31},
  pages={441--450},
  year={2018},
  publisher={Springer}
}

@article{CT2,
  title={Computed tomography super-resolution using deep convolutional neural network},
  author={Park, Junyoung and Hwang, Donghwi and Kim, Kyeong Yun and others},
  journal={Physics in Medicine \& Biology},
  volume={63},
  number={14},
  pages={145011},
  year={2018},
  publisher={IOP Publishing}}

@article{CT6,
  title={CT super resolution via zero shot learning},
  author={Zhang, Zhicheng and Yu, Shaode and Qin, Wenjian and others},
  journal={arXiv preprint arXiv:2012.08943},
  year={2020}
}

@article{CT7,
  title={4$\times$ Super-resolution of unsupervised CT images based on GAN},
  author={Li, Yunhe and Chen, Lunqiang and Li, Bo and Zhao, Huiyan},
  journal={IET Image Processing},
  volume={17},
  number={8},
  pages={2362--2374},
  year={2023},
  publisher={Wiley Online Library}
}

@INPROCEEDINGS{MRI1,
  author={Pham, Chi-Hieu and Ducournau, Aurélien and Fablet, Ronan and Rousseau, François},
  booktitle={2017 IEEE 14th ISBI}, 
  title={Brain MRI super-resolution using deep 3D convolutional networks}, 
  year={2017},
  volume={},
  number={},
  pages={197-200},
  doi={10.1109/ISBI.2017.7950500}}

@inproceedings{MRI2,
  title={Efficient and accurate MRI super-resolution using a generative adversarial network and 3D multi-level densely connected network},
  author={Chen, Yuhua and Shi, Feng and Christodoulou, Anthony G and Xie, Yibin and Zhou, Zhengwei and Li, Debiao},
  booktitle={International conference on MICCAI},
  pages={91--99},
  year={2018},
  organization={Springer}
}

@ARTICLE{MRI5,
  author={Liu, Jianan and Li, Hao and Huang, Tao and others},
  journal={IEEE Transactions on Artificial Intelligence}, 
  title={Unsupervised Representation Learning for 3-D Magnetic Resonance Imaging Superresolution With Degradation Adaptation}, 
  year={2024},
  volume={5},
  number={9},
  pages={4660-4674},
  doi={10.1109/TAI.2024.3397292}}

@InProceedings{MRI6,
author="Liu, Yikang
and Chen, Eric Z.
and Chen, Xiao
and Chen, Terrence
and Sun, Shanhui",
title="An Unsupervised Framework for Joint MRI Super Resolution and Gibbs Artifact Removal",
booktitle="Information Processing in Medical Imaging",
year="2023",
publisher="Springer Nature Switzerland",
address="Cham",
pages="403--414",
isbn="978-3-031-34048-2"
}

@article{MRI7,
  title={Unsupervised MRI super resolution using deep external learning and guided residual dense network with multimodal image priors},
  author={Iwamoto, Yutaro and Takeda, Kyohei and Li, Yinhao and Shiino, Akihiko and Chen, Yen-Wei},
  journal={IEEE Transactions on ETCI},
  volume={7},
  number={2},
  pages={426--435},
  year={2022},
  publisher={IEEE}
}

@article{other2,
  title={Unsupervised arterial spin labeling image superresolution via multiscale generative adversarial network},
  author={Cui, Jianan and Gong, Kuang and Han, Paul and Liu, Huafeng and Li, Quanzheng},
  journal={Medical Physics},
  volume={49},
  number={4},
  pages={2373--2385},
  year={2022},
  publisher={Wiley Online Library}
}

@inproceedings{bsrgan,
  title={Designing a practical degradation model for deep blind image super-resolution},
  author={Zhang, Kai and Liang, Jingyun and Van Gool, Luc and Timofte, Radu},
  booktitle={Proceedings of the IEEE/CVF ICCV},
  pages={4791--4800},
  year={2021}
}

@ARTICLE{dusgan,
  author={Prajapati, Kalpesh and Chudasama, Vishal and Patel, Heena and others},
  journal={IEEE TIP}, 
  title={Direct Unsupervised Super-Resolution Using Generative Adversarial Network (DUS-GAN) for Real-World Data}, 
  year={2021},
  volume={30},
  number={},
  pages={8251-8264},
  doi={10.1109/TIP.2021.3113783}}

@misc{retinal,
  author={A. Hoover and V. Kouznetsova and M. Goldbaum},
  title={STARE: Structured Analysis of the Retina Dataset},
  year={2000},
  url={https://cecas.clemson.edu/~ahoover/stare/}
}

@InProceedings{tudasr,
author="Sarvaiya, Anjali
and Kadel, Jay
and Upla, Kishor
and Raja, Kiran
and Pederson, Marius",
title="TUDASR - Transformer Based Unsupervised Domain Adaptive Super-Resolution for Wireless Capsule Endoscopy",
booktitle="Pattern Recognition. ICPR 2024 International Workshops and Challenges",
year="2025",
publisher="Springer Nature Switzerland",
address="Cham",
pages="395--409",
isbn="978-3-031-88220-3"
}

@INPROCEEDINGS{9607461,
  author={Zhou, Hong-Yu and Lu, Chixiang and Yang, Sibei and Yu, Yizhou},
  booktitle={2021 IEEE/CVF ICCVW}, 
  title={ConvNets vs. Transformers: Whose Visual Representations are More Transferable?}, 
  year={2021},
  volume={},
  number={},
  pages={2230-2238},
  doi={10.1109/ICCVW54120.2021.00252}}

@article{zhang2020stereo,
  title={Stereo endoscopic image super-resolution using disparity-constrained parallel attention},
  author={Zhang, Tianyi and Gu, Yun and Huang, Xiaolin and Tu, Enmei and Yang, Jie},
  journal={arXiv preprint arXiv:2003.08539},
  year={2020}
}

@INPROCEEDINGS{ulcerdetection,
  author={Hossain, Md. Sohag and Al Mamun, Abdullah and Hasan, Md. Galib and Hossain, Md. Motaher},
  booktitle={2019 1st International Conference on Advances in Science, Engineering and Robotics Technology (ICASERT)}, 
  title={Easy Scheme for Ulcer Detection in Wireless Capsule Endoscopy Images}, 
  year={2019},
  volume={},
  number={},
  pages={1-5},
  doi={10.1109/ICASERT.2019.8934510}}

@article{bleeddetection,
  title={Small intestine bleeding detection using color threshold and morphological operation in WCE images},
  author={Al Mamun, A and Hossain, MS and Em, Poh Ping and Tahabilder, Anik and Sultana, R and Islam, MA},
  journal={International Journal of Electrical and Computer Engineering},
  volume={11},
  number={4},
  pages={3040},
  year={2021},
  publisher={IAES Institute of Advanced Engineering and Science}
}
\end{document}